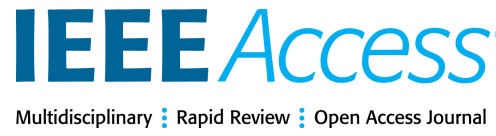

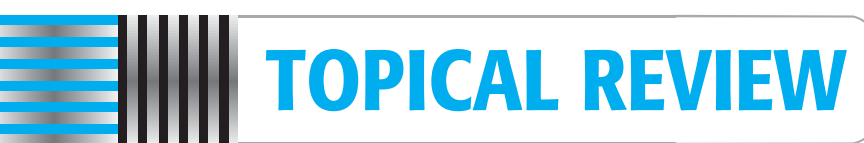

# A Decade of You Only Look Once (YOLO) for Object Detection: A Review

**LEO THOMAS RAMOS**[1,2], **(Graduate Student Member, IEEE), AND ANGEL D. SAPPA**[1,3], **(Senior Member, IEEE)**

[1]Computer Vision Center, Universitat Autónoma de Barcelona, 08193 Barcelona, Spain
[2]Kauel Inc., Menlo Park, Silicon Valley, CA 94025, USA
[3]ESPOL Polytechnic University, Guayaquil 090112, Ecuador

Corresponding authors: Leo Thomas Ramos (ltramos@cvc.uab.cat) and Angel D. Sappa (asappa@cvc.uab.cat; sappa@ieee.org)

This work was supported in part by the Air Force Office of Scientific Research Under Award FA9550-24-1-0206, in part by Grant PID2021-128945NB-I00 funded by MICIU/AEI/ 10.13039/501100011033 and by ERDF, EU and Grant PID2024-162815NB-I00 funded by MICIU/AEI/ 10.13039/501100011033 and by ERDF/EU, and in part by the ESPOL project CIDIS-003-2024. The authors acknowledge the support of the Generalitat de Catalunya CERCA Program to CVC's general activities, and the Departament de Recerca i Universitats from Generalitat de Catalunya to the SGR Research Group 2021 MACO (reference 2021 SGR 01499).

**ABSTRACT** This review marks the tenth anniversary of You Only Look Once (YOLO), one of the most influential frameworks in real-time object detection. Over the past decade, YOLO has evolved from a streamlined detector into a diverse family of architectures characterized by efficient design, modular scalability, and cross-domain adaptability. The paper presents a technical overview of the main versions (from YOLOv1 to YOLOv13), highlights key architectural trends, and surveys the principal application areas in which YOLO has been adopted. It also addresses evaluation practices, ethical considerations, and potential future directions for the framework's continued development. The analysis aims to provide a comprehensive and critical perspective on YOLO's trajectory and ongoing transformation.

**INDEX TERMS** Object detection, deep learning, computer vision, You Only Look Once (YOLO), YOLOv5, YOLOv8, YOLO-NAS, YOLOv11, YOLOv12, YOLOv13, agriculture, autonomous driving, medical image, surveillance, remote sensing, environmental monitoring.

## I. INTRODUCTION

Object detection is a key task in Computer Vision (CV) that involves identifying and localizing all relevant objects within a visual scene [1], [2]. The goal is to predict a set of bounding boxes that accurately enclose each object, along with the corresponding class labels [3]. This requires computational models to jointly address spatial localization and semantic recognition, often under real-time and resource-constrained conditions, making object detection a technically demanding and actively researched area [4].

Thanks to its capacity to extract structured information from visual data, and to the rapid advancement of deep learning techniques [4], [5], object detection has become a fundamental component in a wide range of domains, including surveillance, autonomous driving, and medical imaging [6], [7], [8]. Moreover, it serves as a foundation for more complex CV tasks such as object tracking, instance segmentation, scene understanding, and pose estimation [9], [10], where accurate localization and categorization are required as intermediate steps.

Within the field of object detection, the You Only Look Once (YOLO) family of models has established itself as one of the most influential and widely adopted approaches [11], [12]. It was developed to provide a real-time alternative to two-stage detectors (also called region-based) [13], which divide the detection process into separate stages [4], thereby introducing latency and increasing computational cost [4], [13]. YOLO, on the contrary, proposes a unified architecture that jointly predicts bounding boxes and class probabilities in a single pass over the image [14]. This design results in a significant improvement in inference speed while maintaining competitive accuracy [13], [15].

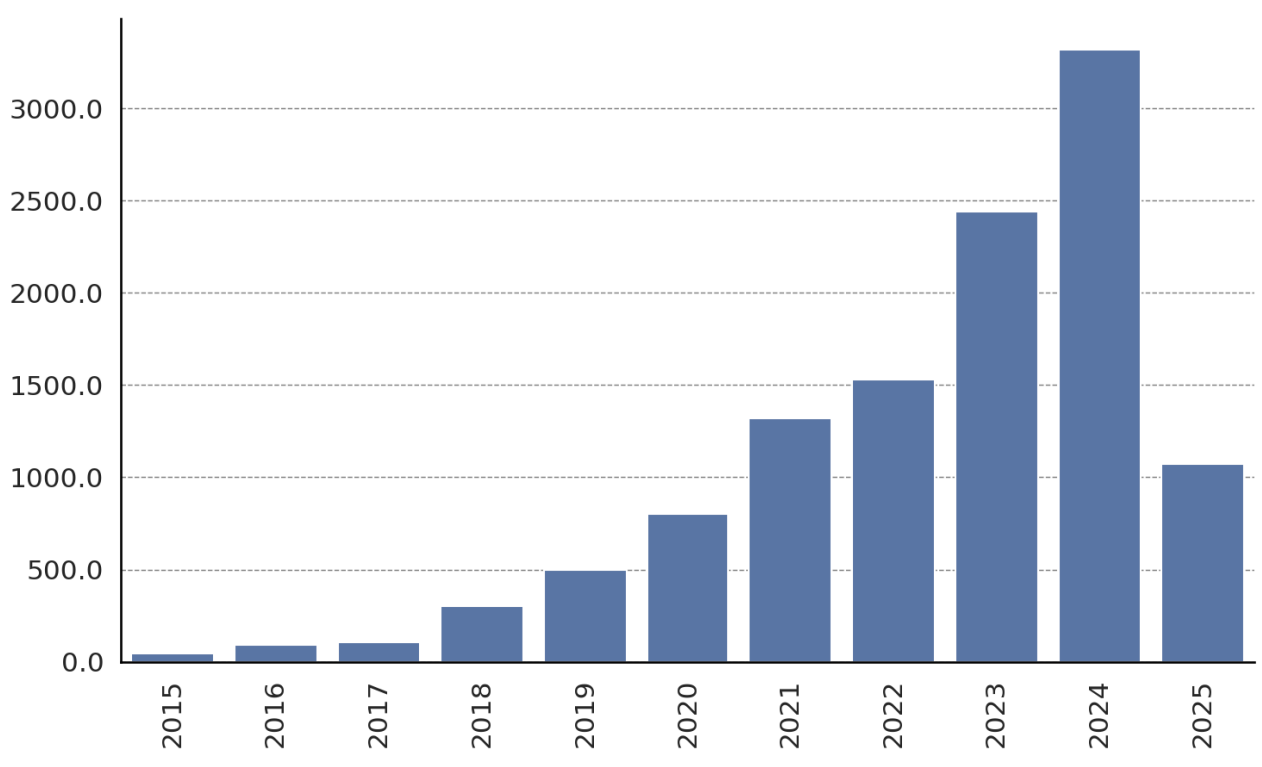

**FIGURE 1. Evolution of the number of publications related to YOLO from 2015 to 2025 (as of June). Data from Google Scholar using advanced search: allintitle: 'YOLO' OR 'You Only Look Once'.**

The introduction of YOLO marked a turning point in object detection, offering a balance between accuracy and efficiency that had not been previously achieved by existing approaches. Its design principles resonated strongly across both academic and industrial communities. As a result, YOLO has not only remained relevant over time but has also evolved through successive versions, each introducing architectural modifications aimed at improving speed, accuracy, and robustness across different deployment scenarios.

The first version of YOLO was introduced in 2015. At the time of developing this work, in 2025, ten years have passed since its initial release. During this period, YOLO has steadily consolidated its position within the object detection community. As reflected in Fig. 1, the number of YOLO-related publications has increased consistently, illustrating the sustained attention and adoption the framework has received over the past decade. Nowadays, it has become a primary choice for a wide range of applications, not only due to its efficiency but also because of its adaptability across different domains and computational settings.

This ten-year mark represents a natural point of reflection on the impact and trajectory of YOLO within the object detection landscape. Motivated by this milestone, the present work aims to provide a review of the YOLO family, examining its technical progression, core architectural ideas, and range of applications. The objective is to consolidate the knowledge accumulated over a decade of development, highlighting the elements that define the framework and the directions it has taken across different contexts. The contributions of this review can be summarized as follows:

1) A contextual introduction to object detection, including foundational methods, benchmark datasets, evaluation metrics, and the historical progression of approaches preceding the emergence of YOLO.
2) A technical overview of each major YOLO architecture developed over the past ten years.
3) An examination of the current landscape of main application areas of YOLO-based models.
4) A structured analysis and critical discussion of key architectural trends, paradigm shifts, and concerns in the evolution of YOLO.
5) A reflection on future directions and expected developments for the YOLO framework.

The remainder of this article is organized as follows. Section II summarises previous work in systematic reviews on YOLO. Section III details the methodology used to carry out this work. Section IV introduces the object detection task and provides essential background to situate the development of YOLO. Section V presents a technical overview of the main YOLO architectures. Section VI examines the current application landscape of YOLO-based models. Section VII provides an analytical outlook on the evolution of YOLO over the past decade, identifying key trends and advances. Section VIII provides insight into future directions for YOLO. Finally, Section IX concludes the paper.

## II. RELATED WORK

Over the past decade, the prominence of YOLO in the field of object detection has motivated the publication of multiple review papers. These reviews vary in scope and emphasis. Some provide broad overviews of the architectural evolution of the YOLO family, while others concentrate on specific application domains. Collectively, they highlight both the technical impact and the practical adoption of YOLO across research and industry.

Some reviews have addressed YOLO with a focus on its applications in specific domains rather than as a comprehensive technical review. For instance, [16] covers YOLOv1 through YOLOv12 with emphasis on autonomous driving, outlining its use in detecting vehicles, pedestrians, traffic lights, and lane markings. While it acknowledges architectural evolution, the treatment remains synthetic and shifts quickly toward application scenarios. Similarly, [11] provides a systematic review of YOLO in medical object detection between 2018 and 2023, considering tasks such as lesion or tumor detection and surgical instrument recognition. Although the paper briefly mentions architectural aspects, the discussion is cursory, limited to PubMed as the sole source, and largely descriptive rather than critical. In the agricultural domain, [17] presents a review of YOLO applications across versions up to YOLO-NAS, but again, the architectural overview is short and the focus rests primarily on agricultural tasks. Moreover, its reliance on Scopus alone constrains the breadth of coverage. These surveys demonstrate the relevance of YOLO across domains, but they lack the depth, breadth, and architectural analysis required for a unified technical review.

Other reviews have concentrated on the early development of the YOLO framework itself. For example, the work of [18] provides a technical comparison of five main versions (YOLOv1 through YOLOv5), analysing differences in network architecture, loss functions, and computational efficiency. Its perspective is primarily technical and explana-tory, aimed at understanding the internal mechanisms of these models rather than applications. However, the review is relatively short, omits more recent versions starting with YOLOv6. Similarly, [19] reviews one-stage detectors with special emphasis on YOLO, tracing its evolution up to YOLOv4 with attention to improvements in accuracy, inference speed, and scalability. Yet, it also excludes versions from YOLOv5 onward and limits its discussion to structural aspects, without critically engaging with applications or future research directions. These characteristics restrict both works to an early and partial view of YOLO's evolution.

More recent surveys attempt to provide broader coverage of the YOLO family across multiple versions. For example, [14] conducts a systematic review spanning YOLOv1 through YOLOv8, examining both architectural and methodological aspects. While extensive in scope, its approach remains largely descriptive, with little critical discussion of applications, broader trends, or future perspectives. Similarly, [20] offers a detailed review of versions up to YOLOv8, presenting each model's specific innovations and also addressing datasets and evaluation protocols. Despite its strengths, it does not cover developments beyond YOLOv8, leaving it partially outdated. The work of [21] extends to YOLOv8 and YOLO-NAS, providing a systematic overview of architectural changes, evaluation metrics, and training improvements. Yet, its focus remains technical, with limited discussion of applications or forward-looking analysis. Finally, [22] reviews YOLO up to YOLOv11, combining architectural coverage with applications in diverse domains, and addressing limitations, challenges, and future perspectives. Even so, its application analysis is selective and it still omits the most recent YOLO releases. Together, these surveys offer a more complete perspective than earlier efforts, but remain limited in certain respects or by their lack of coverage of the latest YOLO developments.

As seen across the surveyed literature, prior reviews of YOLO vary considerably in scope and depth. Some are primarily application-oriented, focusing on domains such as autonomous driving, medicine, or agriculture, but offering only brief or synthetic treatments of YOLO's architectural evolution. Others concentrate on the early versions of the framework, providing technical comparisons, yet omitting subsequent developments and thus remaining incomplete. More recent works cover a broader range of versions, but tend to remain limited on critical analysis of trends, ecosystem fragmentation, or long-term perspectives. This confirms the relevance of YOLO and the need for periodic synthesis, but almost none provides a decade-long, methodologically grounded, and critically oriented review that integrates architectural evolution, applications, and broader issues in a unified framework. A summary of these related work can be found in Table II.

In contrast, this review provides a comprehensive and cohesive synthesis that spans the full decade of YOLO's development, from its origins to the most recent versions. It differs from prior surveys by combining a systematic methodology, a critical thematic analysis of architectural shifts and training strategies, and a bibliometric study that grounds the mapping of applications. Through this integrated perspective, the work contributes not only a detailed account of YOLO's evolution but also a broader reflection on its challenges, future directions, and enduring role in the object detection landscape.

## III. METHODOLOGY FOR CONDUCTING THE REVIEW

This review follows a structured and transparent process inspired by the principles of the Preferred Reporting Items for Systematic Reviews and Meta-Analyses (PRISMA) [23]. In line with these guidelines, the study was designed around explicit research questions, clear eligibility criteria, and a systematic strategy for identifying and organizing the literature. First, we defined a set of research questions that guided the entire review process:

- **Q1** What are the most significant YOLO models developed over the past decade, and how do they differ in terms of architecture?
- **Q2** Which major paradigm shifts can be identified in the evolution of YOLO?
- **Q3** What are the main application domains where YOLO has been adopted?
- **Q4** What ethical, social, and environmental concerns arise from the widespread deployment of YOLO models?
- **Q5** What directions and open challenges can be identified for the future evolution of YOLO?

Next, we defined a series of criteria for the selection of works. A study was included in our review if it met the following inclusion criteria (IC):

- **IC1** Empirical research, not books, manuals, or tutorials.
- **IC2** Articles published in peer-reviewed scientific journals or presented at relevant scientific conferences.
- **IC3** Articles published between 2015 and 2025.
- **IC4** Articles describing the implementation, evaluation, or validation of a YOLO model or a model primarily based on YOLO.
- **IC5** Articles written in English.

The exclusion criteria (EC) were as follows:

- **EC1** Articles focusing on other paradigms or detector families different from YOLO.
- **EC2** Review articles, meta-analyses, editorials, commentaries, and other types of secondary research.
- **EC3** Articles not peer-reviewed or not sufficiently adopted by the scientific community.
- **EC4** Articles lacking sufficient methodological or experimental detail for meaningful interpretation.

Regarding the information sources and search strategy, the literature analysed in this review was retrieved exclusively from reputable and well-established databases to guarantee the quality and reliability of the included works. Specifically, the sources consulted were IEEE Xplore,[1] ScienceDirect,[2] SpringerLink,[3] and the ACM Digital Library.[4] In addition, Google Scholar was employed as a complementary tool to locate further relevant studies.

**TABLE 1.** Summary of existing YOLO review papers.

| YOLO versions covered | Main focus | Limitations | Year | Reference |
|---|---|---|---|---|
| v1-v5 | Technical comparison (architecture, loss, small object detection) | Omits v6 onwards, no real-world use, short paper with limited depth | 2022 | [18] |
| v1-v4 | One-stage detectors with YOLO emphasis | Covers up to v4, lacks broader discussion, limited to technical comparison | 2022 | [19] |
| v1-v8 | Medical imaging (lesions, tumors, surgical tools) | Limited architectural review, only PubMed source, descriptive not critical, misses recent versions | 2023 | [11] |
| v1-v8, YOLO-NAS | Technical evolution (architecture, metrics, training tricks) | Limited discussion/analysis, shallow future perspectives, misses recent versions | 2023 | [21] |
| v1-v8 | Systematic coverage of architecture and training strategies | Mostly descriptive, lacks critical depth on applications and trends, no recent versions | 2024 | [14] |
| v1-v8 | Detailed architectural evolution, datasets, evaluation | Covers up to v8, no v9 onwards, lacks broader perspective | 2024 | [20] |
| v1-v11 | Architecture and applications, includes challenges and future | Limited application and trend analysis, no coverage of latest versions | 2024 | [22] |
| v1-v8, YOLO-NAS | Agriculture (object detection tasks) | Focus on Scopus only, brief YOLO description, no recent versions, application-centered | 2024 | [17] |
| v1-v12 | Autonomous driving (vehicles, pedestrians, traffic signs) | General overview, limited architectural depth, focus on a specific application | 2025 | [16] |

Regarding the search strategy, targeted keywords were applied to ensure comprehensive coverage of the YOLO literature. The primary terms included variations such as *"YOLO"*, *"You Only Look Once"*, and combinations with related expressions such as *"YOLO object detection"*, *"YOLO real-time detection"*, and *"YOLO applications"*. Boolean operators were used where appropriate to refine and broaden the search results.

The selection process was carried out in successive phases. In the first stage, each author briefly examined the candidate papers by reviewing their abstracts, key results, and conclusions, which allowed for an initial screening. In the second stage, the remaining works were read in full detail and systematically categorized. At this point, the inclusion and exclusion criteria previously described were applied. In addition, preference was given to seminal contributions, highly cited studies, and the most recent works that reflect current trends. Through this process, we ensured that the review incorporates the most relevant and influential research. Finally, only those studies that fully met these criteria were included in the present review.

[1] https://ieeexplore.ieee.org
[2] https://www.sciencedirect.com
[3] https://link.springer.com
[4] https://dl.acm.org

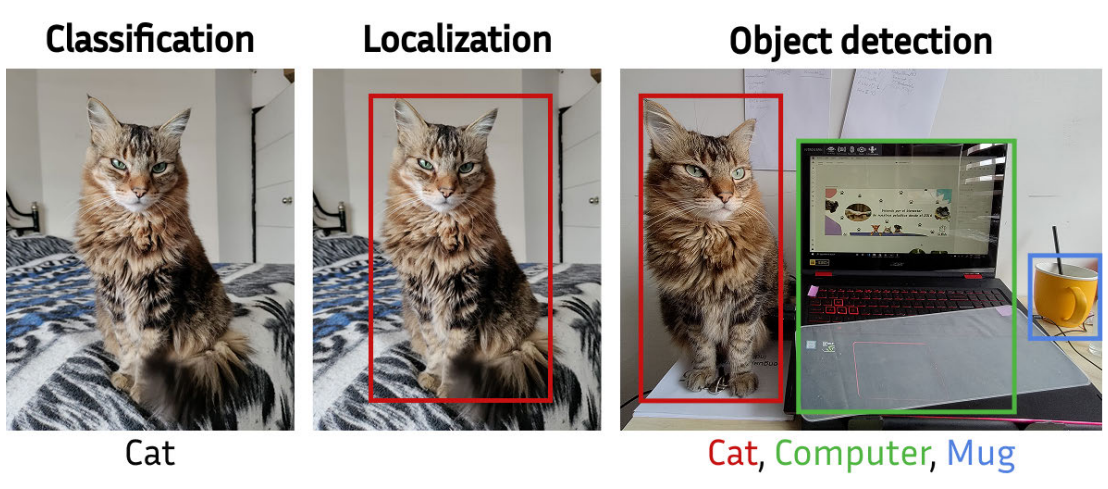

**FIGURE 2.** Comparison between classification, localization, and object detection.

## IV. BACKGROUND

### A. FUNDAMENTALS OF OBJECT DETECTION

As previously mentioned, object detection involves identifying the objects present in an image and determining where they are located. It is commonly regarded as a combination of image classification and object localization. Image classification consists of assigning a label to an image, assuming the presence of a single dominant object. Object localization extends this by predicting the position of that object using a bounding box. Object detection combines both tasks, as it locates the presence of objects through bounding boxes and assigns a class to each instance, as shown in Fig. 2. Furthermore, it generalizes to scenes with multiple objects and multiple occurrences of the same object class.

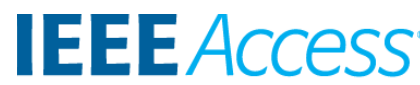

The process of object detection generally follows a structured sequence of operations. Given an input image, the first step is to extract visual features that capture relevant patterns and spatial information. These features are then used to generate a set of predictions that aim to describe the objects present in the scene. Each prediction includes both spatial information and a category label, reflecting the location and identity of an object candidate. The final step consists of selecting and refining the predictions to produce the final detection output.

The output of a detector is typically composed of three elements: a bounding box, a class label, and a confidence score. The bounding box is defined by four numerical values, usually in the format $(x, y, w, h)$ [10], representing the horizontal and vertical coordinates of the top-left corner, along with the width and height of the box. The class label corresponds to one of the predefined categories in the dataset. Finally, the confidence score is a real-valued scalar in the interval [0, 1] that quantifies the detector's certainty about the presence of an object within the predicted region [5].

Since object detection typically produces multiple overlapping predictions for the same object [24], a post-processing step is required to eliminate redundant outputs. Non-Maximum Suppression (NMS) is the standard technique for this task. It selects the prediction with the highest confidence score and removes others that exhibit significant overlap with it, based on a predefined Intersection over Union (IoU) threshold [25], [26]. This procedure is repeated until no overlapping predictions remain.

In the context of object detection, several challenges arise from the inherent complexity of real-world visual scenes. These challenges affect the accuracy, robustness, and generalization capacity of object detectors, especially under real-world conditions. Some of the most common factors that make object detection particularly challenging are:

- **Occlusion**. It refers to the partial obstruction of objects by other elements in the scene [14]. This condition reduces the amount of visible information, which can impair feature extraction and lead to incorrect or missed predictions.
- **Scale variation**. It refers to the large differences in object sizes across images or within the same image [27]. These variations challenge detectors to accurately represent and localize both small and large instances using a fixed receptive field or resolution.
- **Small objects**. Detecting objects that cover only a small region of the image is another challenge in object detection. Low spatial resolution and weak feature representations often lead to missed detections or poor localization accuracy [5].
- **Class imbalance**. This condition arises when object categories appear with highly unequal frequency in the training data [28]. This often results in degraded performance on underrepresented categories, especially when coupled with limited training samples.
- **Intra-class variation**. This challenge refers to the high diversity in appearance among instances of the same object class [5], caused by differences in pose, lighting, texture, or occlusion. Such variability increases intra-class dispersion in feature space, making it harder for detectors to consistently recognize all instances.
- **Scalability and efficiency**. Object detectors are expected to scale across a range of deployment scenarios, from high-performance servers to resource-constrained devices [29]. This requires models to be efficient in terms of memory, computation, and latency, while maintaining consistent performance across different data volumes, input resolutions, and hardware platforms.

### B. OBJECT DETECTION BEFORE YOLO: FOUNDATIONAL METHODS AND PARADIGM SHIFTS

The conception of object detection methods dates back over two decades. Early research in this area was grounded in the use of hand-crafted features and algorithmic heuristics. The process typically involved extracting local patterns using low-level features, followed by classification using traditional machine learning algorithms.

One of the earliest influential frameworks is the Viola-Jones detector [30], [31], which introduced a cascade-based architecture for rapid face detection. It relies on Haar-like features computed over integral images and applied a boosting algorithm to select and combine simple classifiers into an efficient decision process [32]. Shortly after, the Histogram of Oriented Gradients (HOG) [33] method was introduced. It works by computing histograms of oriented gradients over small spatial cells of the image, which are then normalized across blocks to increase robustness to illumination and contrast changes. When combined with a linear Support Vector Machine (SVM), this approach allows robust detection of objects with strong edge structures, such as pedestrians. Building upon these foundations, Deformable Part Models (DPM) [34] introduced a part-based representation in which an object is modeled as a root filter plus a set of parts, each with its own filter and is allowed to shift from its ideal position, with a deformation cost penalizing large displacements. Detection is performed by scoring the root and parts across the image and combining their responses according to both appearance and spatial configuration.

Although effective at the time, traditional methods were often limited in their adaptability and struggled to handle the complexity of real-world scenes. Several years later, the widespread adoption of deep learning fundamentally reshaped the CV landscape. Specifically, Convolutional Neural Networks (CNNs) became the dominant paradigm for visual recognition tasks, including object detection. This shift was enabled by the ability of CNNs to learn robust and high-level feature representations directly from data, enabling more flexible and scalable solutions to complex detection problems.

Two-stage detectors became the dominant paradigm in the early deep learning era. These methods decouple the detection process into two sequential steps: first, generating candidate regions likely to contain objects; and second, classifying and refining those regions using convolutional features [13]. A representative method in this family is Region-based Convolutional Neural Networks (R-CNN) [35], which generates region proposals using selective search and then applies a CNN independently to each region to extract features. These features are subsequently fed into class-specific linear SVMs for classification, while bounding box regression is used to refine the localization of each detection. Despite its effectiveness, R-CNN suffers from high computational and memory demands [36]. The sequential nature of the pipeline for processing each proposed region results in significant latency during inference, and limits its applicability in real-time scenarios.

Given the limitations of R-CNN, other approaches were developed aiming to achieve faster inference while maintaining competitive accuracy. Fast R-CNN [37] improves the pipeline by processing the entire image with a CNN only once to generate a shared feature map. Region proposals are then projected onto this map, and a Region of Interest (RoI) pooling layer extracts fixed-size feature vectors for each proposal. These are fed into a single network that performs both classification and bounding box regression. This design significantly reduces computational redundancy and enables end-to-end training. Shortly after, Faster R-CNN [38] was introduced to eliminate the dependency on external region proposal methods. It integrates a Region Proposal Network (RPN) directly into the detection pipeline, which shares convolutional features with the detection head and generates nearly cost-free region proposals in a fully learnable manner.

Feature Pyramid Network (FPN) [39] is another development within this family of methods, designed to enhance object detection across multiple scales by constructing a multi-level feature representation. Instead of relying solely on the deepest layers of a convolutional backbone, where semantic information is rich but spatial resolution is poor, FPN introduces a top-down pathway with lateral connections. This structure combines high-level semantic features with lower-level spatial details, generating a set of feature maps at different resolutions. By doing so, FPN enables detectors to effectively recognize both small and large objects. It was typically integrated into architectures such as R-CNN, where it contributed to substantial improvements in detection performance.

Although two-stage methods achieve high detection accuracy, and despite efforts to reduce their computational burden, their architectural complexity and inference latency remained limiting factors for practical application. In response to these constraints, a parallel line of research emerged with the development of one-stage detectors. These methods eliminate the region proposal stage entirely and perform object classification and localization in a single unified step, typically by treating detection as a dense prediction problem over a regular grid of locations. This design offers significant improvements in inference speed and architectural simplicity, making them suitable for scenarios where real-time performance is required.

A well-known example of this class of detectors is the Single Shot MultiBox Detector (SSD) [40], which performs object detection by applying a set of default bounding boxes with different aspect ratios and scales across multiple feature maps. Each location in the feature maps is used to simultaneously predict class scores and box offsets, allowing detection at multiple resolutions. SSD avoids region proposals altogether and integrates classification and localization into a single pass, offering a favorable balance between speed and accuracy. Despite these advantages, and although SSD helped popularize the one-stage paradigm, a few months earlier another architecture had been introduced; one that would redefine object detection by promoting the idea of ‘seeing only once.’

### C. BENCHMARK DATASETS

Datasets play a fundamental role in the development and evaluation of deep learning-based object detection models. Robust and diverse datasets are necessary not only for training models capable of generalizing to real-world scenarios, but also for benchmarking progress in a consistent and reproducible manner. Over the past years, a number of datasets have been developed to support research in this area, with some becoming widely adopted as standard references. Below, we briefly describe the most representative benchmarks used in the field.

#### 1) PASCAL VOC

The PASCAL Visual Object Classes (VOC) dataset [41] is one of the earliest benchmarks for object detection. It provides annotated images for 20 object categories across a variety of scenes. The most commonly used editions are VOC 2007 and VOC 2012, which contain approximately 9k and 11k images respectively, with more than 12k and 27k annotated object instances. Both versions provide predefined train, validation, and test splits.

#### 2) ILSVRC

The ImageNet Large Scale Visual Recognition Challenge (ILSVRC) [42] includes an object detection task introduced in 2013 as part of its annual competition. The detection track is based on a subset of the full ImageNet dataset and focuses on 200 object categories. Annotations are provided for both training and validation sets, while the test set is used for leaderboard evaluation.

#### 3) MS-COCO

The Microsoft Common Objects in Context (MS-COCO) [43] is a dataset designed to support object detection in complex, natural scenes. It includes 80 object categories annotated in approximately 330k images, containing more than 2.5 million object instances. One of the challenges associated with COCO is the high proportion of small objects, which increases detection difficulty and model sensitivity to spatial resolution. COCO provides annotations with bounding boxes, segmentation masks, and captions for each image, as well as keypoints for a subset of categories. The dataset includes predefined splits for training, validation, and testing, and offers multiple test sets for different evaluation tracks.

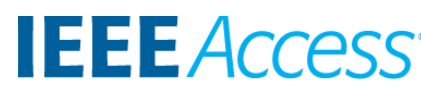

#### 4) OPENIMAGES

It is a large-scale dataset developed by Google.[5] It contains approximately 9 million images, with around 1.9 million of them annotated for object detection. It covers more than 600 object classes, with approximately 16 million bounding boxes. Annotations comprise image-level labels, object bounding boxes, visual relationships, and instance segmentations. Predefined splits are available for training, validation, and testing.

#### 5) OTHER DATASETS

In addition to the main benchmarks described above, other datasets have been developed and adapted to target specific domains or scenarios. Examples include BDD100K [44], KITTI [45], and Waymo [46], focused on urban driving/street environments; DOTA [47] and VisDrone [48], designed for object detection in aerial imagery; and CrowdHuman [49], which targets detection of people in highly crowded scenes.

### D. PERFORMANCE METRICS

Evaluating object detection models requires metrics that account for both classification accuracy and spatial localization quality. Unlike image classification, where a single label prediction suffices, object detection involves predicting multiple bounding boxes with associated class labels. Therefore, performance is typically assessed using a combination of classification and localization metrics, defined in terms of the agreement between predicted and ground-truth regions. Below, we describe the most commonly used metrics in the field.

#### 1) PRECISION

It quantifies the proportion of predicted bounding boxes that correspond to true object instances [50]. It reflects how reliable the model is when it declares the presence of an object. It is formally defined as shown in Eq. 1:

$$\text{Precision} = \frac{TP}{TP + FP}, \tag{1}$$

where true positives (TP) are correctly predicted bounding boxes, and false positives (FP) are predicted boxes that do not sufficiently match any ground-truth object. This matching is typically determined using an IoU threshold.

[5] https://storage.googleapis.com/openimages/web/index.html

#### 2) RECALL

It measures the model's ability to identify all relevant objects present in an image [50], [51]. It represents the proportion of ground-truth instances that are successfully detected by the model. It is defined as shown in Eq. 2:

$$\text{Recall} = \frac{TP}{TP + FN}, \tag{2}$$

where false negatives (FN) refer to ground-truth objects that the model failed to detect.

#### 3) F1-SCORE

F1-score provides a single value that balances precision and recall, and is often used as an alternative metric in scenarios where a unified view of detection performance is desired [5]. While less commonly used in standard object detection benchmarks, it may be relevant in application-specific evaluations. The F1-score is defined as the harmonic mean of precision and recall, as shown in Eq. 3:

$$\text{F1-score} = 2 \times \frac{\text{Precision} \times \text{Recall}}{\text{Precision} + \text{Recall}}. \tag{3}$$

#### 4) MEAN AVERAGE PRECISION

The mean Average Precision (mAP) is the most widely used evaluation metric in object detection [50]. It summarizes the performance of a model by averaging its detection precision across different recall levels and object classes. The computation begins with the Average Precision (AP), which corresponds to the area under the precision-recall curve for a single class, as shown in Eq. 4:

$$\text{AP} = \int_0^1 \text{precision}(r)\mathrm{d}r. \tag{4}$$

Detections are matched to ground truth instances using an IoU threshold. For multi-class evaluation, AP is computed independently for each class and then averaged to obtain mAP, as shown in Eq. 5:

$$\text{mAP} = \frac{1}{N}\sum_{i=1}^{N} \text{AP}_i, \tag{5}$$

where $N$ is the number of classes. Two common variants are mAP@0.5, used in the PASCAL VOC challenge, and mAP@0.5:0.95, introduced by the COCO benchmark, which averages AP over thresholds from 0.5 to 0.95 in steps of 0.05.

#### 5) INFERENCE SPEED

Inference speed is not a standard accuracy metric like F1-score or mAP, but it is a critical operational consideration in real-time object detection tasks. It reflects the practical efficiency of a model when deployed and is typically measured in terms of Frames Per Second (FPS) or latency per image. Inference speed in FPS can be defined as shown in Eq. 6:

$$\text{FPS} = \frac{N}{T}, \tag{6}$$

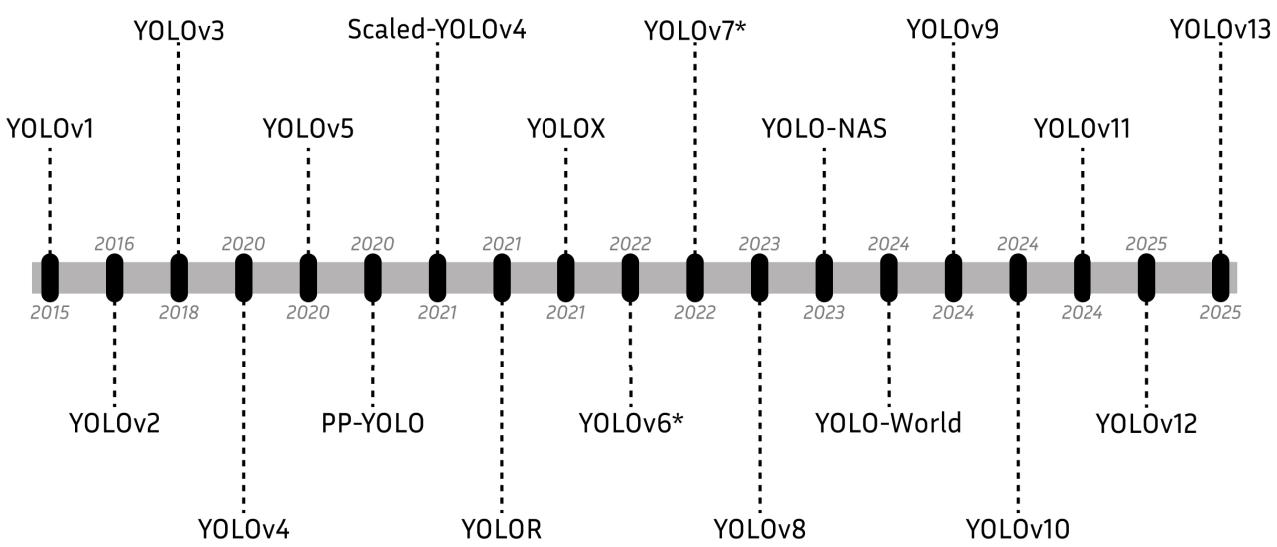

**FIGURE 3. Timeline of major developments in the YOLO framework over the past decade.**

where $N$ is the number of images processed and $T$ is the total time taken (in seconds) to process them [52]. Meanwhile, latency corresponds to the time taken to process a single image, often reported in milliseconds [53], [54].

Unlike precision or recall-based metrics, inference speed depends not only on the model architecture but also on the hardware, software optimizations, batch size, and input resolution. For this reason, inference speed is context-dependent and varies significantly across deployment setups.

## V. OVERVIEW OF YOLO ARCHITECTURES

The YOLO framework has evolved into a family of models released over the past decade. Fig. 3 illustrates the chronological order of major versions and relevant variants. In the following subsections, we provide a technical overview of each principal YOLO model, outlining their key components and contributions.

### A. YOLOv1

The YOLO family of object detectors begins with the release of its first version, YOLOv1 [55], in 2015. YOLOv1 is a pioneer in object detection, redefining the field that had previously relied heavily on region proposals or sliding window mechanisms, as exemplified by methods like R-CNN. YOLOv1 departs from this paradigm by formulating object detection as a single regression problem, predicting bounding boxes and class probabilities directly from the input image using a unified convolutional neural network.

At the core of YOLOv1 lies a design that processes an input image of fixed resolution. The image is divided into a grid of size $S \times S$, where each cell is responsible for detecting objects whose centers fall within its boundaries. Each cell predicts $B$ bounding boxes, each defined by five components: the normalized coordinates of the box center relative to the cell, the width and height relative to the image dimensions, and a confidence score that estimates both the presence of an object and the accuracy of the box. Additionally, each cell outputs a set of conditional class probabilities, assuming an object is present. The complete prediction is represented as a tensor of shape $S \times S \times (B \times 5 + C)$, where $C$ is the number of target classes. This formulation enables simultaneous localization and classification, allowing the detection process to be executed in a single forward pass through the network.

The architecture of YOLOv1 consists of a convolutional neural network with 24 convolutional layers followed by 2 fully connected layers, as shown in Fig. 4. The backbone extracts hierarchical spatial features through alternating $1 \times 1$ convolutions, used for dimensionality reduction, and $3 \times 3$ convolutions, used for spatial feature extraction. This design is inspired by GoogLeNet [56] and its network-in-network approach, but omits the use of more complex inception modules. Downsampling is performed through max-pooling operations, which progressively reduce spatial resolution while increasing the depth of the feature maps. The final convolutional block outputs a fixed-size spatial grid over which predictions are made.

Following the convolutional layers, two fully connected layers process the extracted features and produce the final output. The network outputs a multi-dimensional tensor encoding bounding box coordinates, objectness confidence scores, and conditional class probabilities for each spatial cell. Internally, all hidden layers utilize a leaky ReLU activation function, while the output layer is linear to allow for unrestricted real-valued predictions.

Although YOLOv1 marked a departure from previous object detection paradigms, it presents notable limitations. Its single-pass design enables the use of global contextual information but also imposes spatial constraints, as each grid cell predicts a fixed number of bounding boxes and a single class. This limits the detection of multiple small or overlapping objects within the same region, leading to poor performance in crowded scenes or when objects are spatially close. Furthermore, the coarse resolution of the final feature map, caused by successive downsampling, hinders the precise localization of small objects and biases the model toward larger, well-separated instances. Despite these constraints, YOLOv1 established the foundation for a new generation of real-time object detectors.

### B. YOLOv2

One year after the introduction of YOLOv1, the second version, YOLOv2 [57], was released in 2016. It builds on the core principles of YOLOv1 and introduces architectural refinements and methodological improvements aimed at increasing detection accuracy while preserving high inference speed.

At the core of YOLOv2 is a newly introduced feature extractor, Darknet-19, composed of 19 convolutional layers and 5 max-pooling layers, as shown in Table 2. It primarily employs $3 \times 3$ convolutional filters, doubling the number of channels after each pooling operation, similar to the VGG architecture. To reduce dimensionality, $1 \times 1$ convolutions are inserted between the $3 \times 3$ layers, following the network-in-network design. For object detection, the final classification layers of Darknet-19 are removed and replaced with additional convolutional layers to refine the output predictions.

Another important architectural change is the shift to a fully convolutional design. The fully connected layers

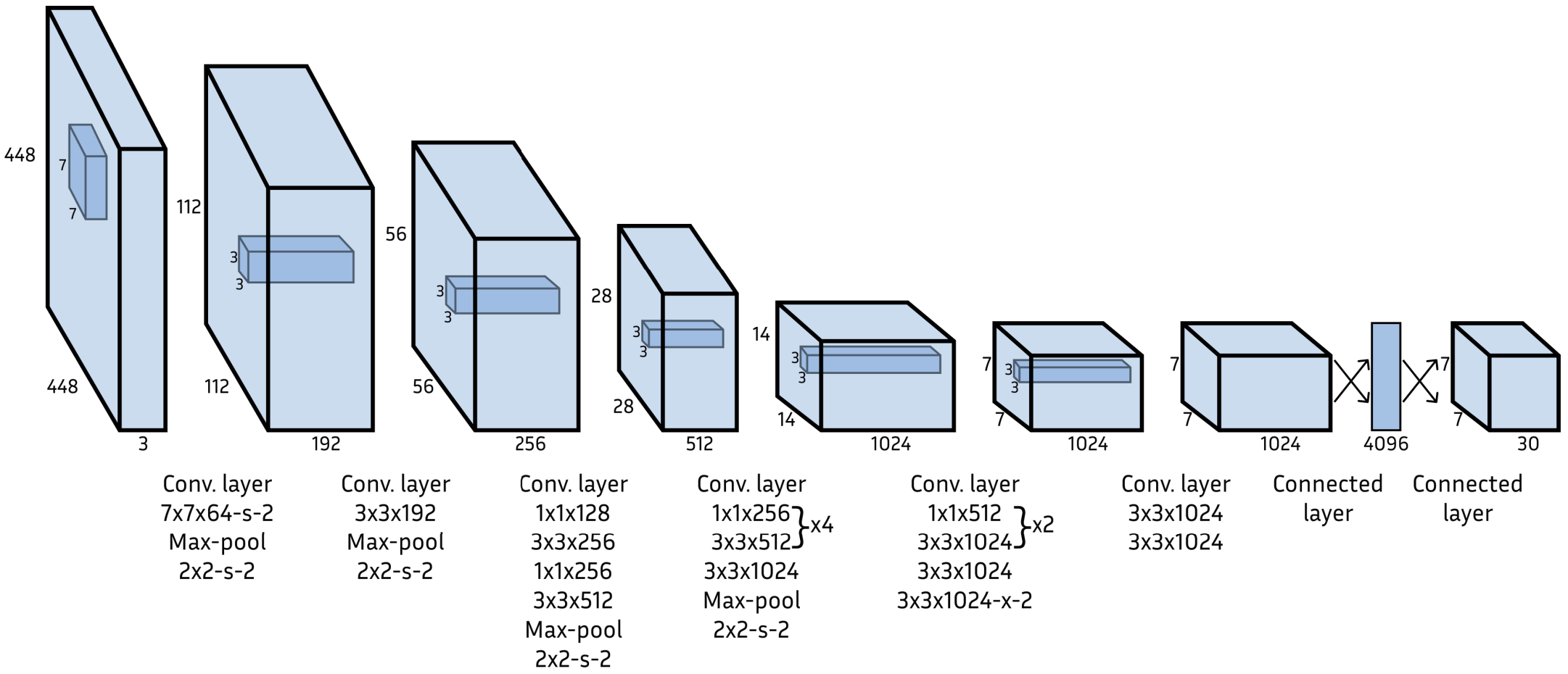

**FIGURE 4.** YOLOv1 architecture.

**TABLE 2.** Structure of Darknet-19 backbone network introduced in YOLOv2.

| Type | Filters | Size/Stride | Output |
| --- | --- | --- | --- |
| Convolutional | 32 | $3 \times 3$ | $224 \times 224$ |
| Maxpool | | $2 \times 2 / 2$ | $112 \times 112$ |
| Convolutional | 64 | $3 \times 3$ | $112 \times 112$ |
| Maxpool | | $2 \times 2 / 2$ | $56 \times 56$ |
| Convolutional | 128 | $3 \times 3$ | $56 \times 56$ |
| Convolutional | 64 | $1 \times 1$ | $56 \times 56$ |
| Convolutional | 128 | $3 \times 3$ | $56 \times 56$ |
| Maxpool | | $2 \times 2 / 2$ | $28 \times 28$ |
| Convolutional | 256 | $3 \times 3$ | $28 \times 28$ |
| Convolutional | 128 | $1 \times 1$ | $28 \times 28$ |
| Convolutional | 256 | $3 \times 3$ | $28 \times 28$ |
| Maxpool | | $2 \times 2 / 2$ | $14 \times 14$ |
| Convolutional | 512 | $3 \times 3$ | $14 \times 14$ |
| Convolutional | 256 | $1 \times 1$ | $14 \times 14$ |
| Convolutional | 512 | $3 \times 3$ | $14 \times 14$ |
| Convolutional | 256 | $1 \times 1$ | $14 \times 14$ |
| Convolutional | 512 | $3 \times 3$ | $14 \times 14$ |
| Maxpool | | $2 \times 2 / 2$ | $7 \times 7$ |
| Convolutional | 1024 | $3 \times 3$ | $7 \times 7$ |
| Convolutional | 512 | $1 \times 1$ | $7 \times 7$ |
| Convolutional | 1024 | $3 \times 3$ | $7 \times 7$ |
| Convolutional | 512 | $1 \times 1$ | $7 \times 7$ |
| Convolutional | 1024 | $3 \times 3$ | $7 \times 7$ |
| Convolutional | 1000 | $1 \times 1$ | $7 \times 7$ |
| Avgpool | | Global | 1000 |
| Softmax | | | |

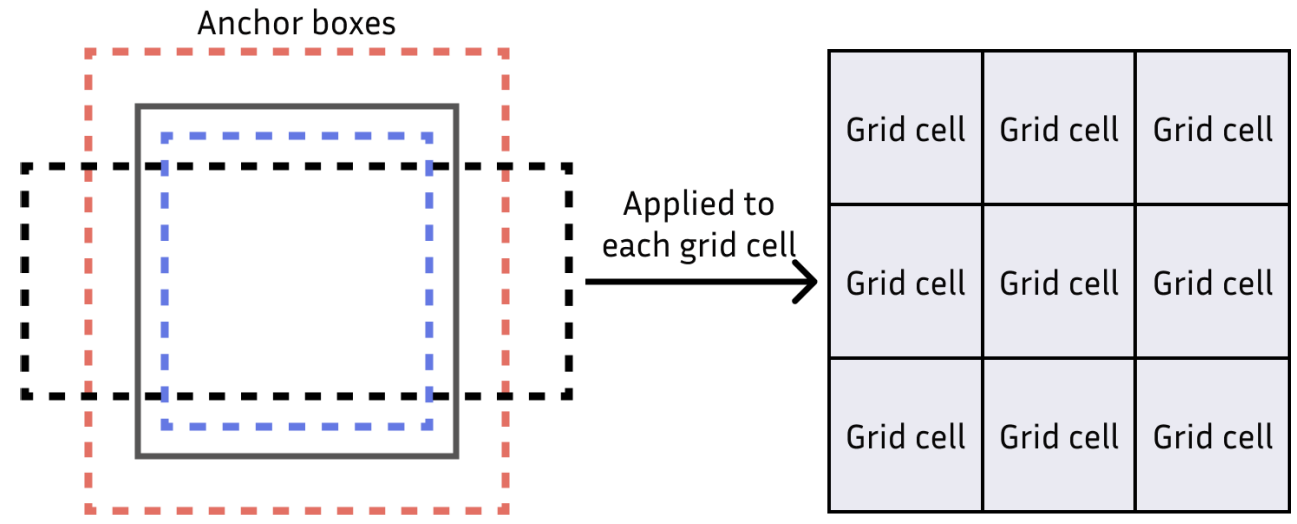

(a) Assignment of anchor boxes to each grid cell.

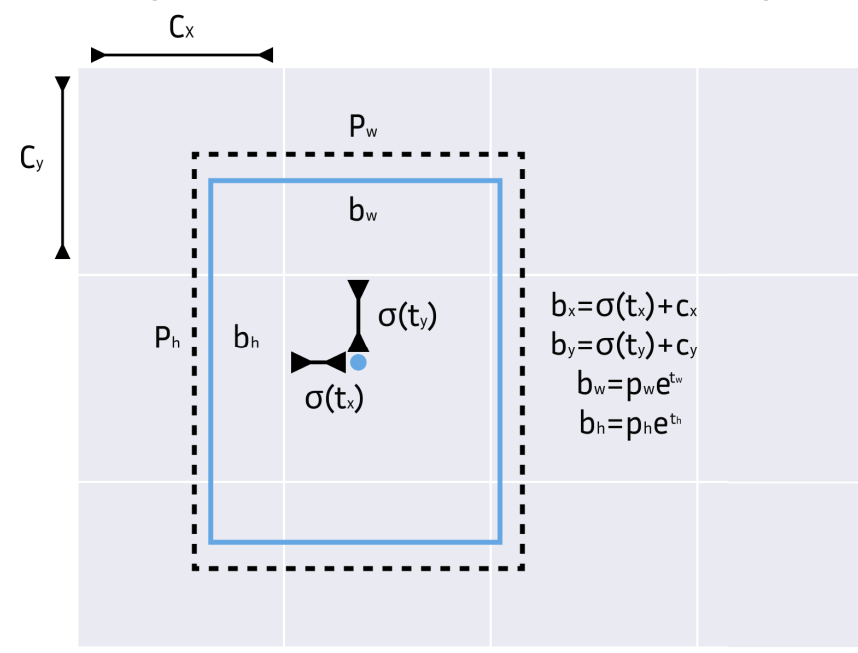

(b) Bounding box prediction mechanism.

**FIGURE 5.** Use of anchor boxes for bounding box prediction in YOLOv2.

used in YOLOv1 are removed, and predictions are now made directly on convolutional feature maps using anchor boxes, following the strategy introduced in region proposal networks. Rather than regressing bounding box coordinates directly, the network predicts offsets relative to a set of predefined anchor shapes. These anchors are uniformly applied across the final feature map, and for each one, the model predicts bounding box offsets, an objectness score, and class probabilities. This design significantly increases the number of predicted boxes per image and enhances recall.

To improve the quality of anchor boxes, YOLOv2 replaces hand-crafted dimensions with data-driven priors obtained through k-means clustering on bounding box annotations. The clustering uses Intersection over Union (IoU) as the distance metric, rather than Euclidean distance, to ensure better alignment with the detection objective. This yields anchor boxes that more closely match the dataset's distribution, facilitating the learning process. Additionally, batch normalization is applied to all convolutional layers, improving convergence and providing regularization. As a result, dropout is no longer required to prevent overfitting.

Additionally, before fine-tuning for detection, YOLOv2 performs classification pretraining at a higher resolution. The base classifier, initially trained at $224 \times 224$, is fine-tuned at $448 \times 448$ for ten epochs to help the network adapt its filters to higher-resolution inputs, which improves its detection performance. To enhance training stability and bounding box precision, YOLOv2 also introduces a direct location prediction mechanism, illustrated in Fig. 5. Bounding box centers are predicted relative to the grid cell using a logistic activation, ensuring they remain within the cell's spatial bounds. Width and height are predicted as log-space offsets from the anchor dimensions. This formulation stabilizes training and improves localization accuracy.

YOLOv2 also addresses a key limitation of YOLOv1 by improving its performance on small objects. To enhance the capture of fine-grained spatial features, it introduces a passthrough layer that merges high-resolution activations from earlier layers with deeper, lower-resolution features from the final detection layers. Concretely, a $26 \times 26$ feature map from an earlier layer is reshaped and concatenated with the final $13 \times 13$ detection layer. This mechanism is similar to Residual Networks (ResNet) [58] skip connections and enhances the detector's access to fine-grained spatial information.

A final architectural innovation in YOLOv2 is the use of multi-scale training. Since the network is composed entirely of convolutional and pooling layers, it can accept inputs of varying resolution. During training, the input size is randomly changed every few iterations, selected from a predefined set of multiples of 32 ranging from $320 \times 320$ to $608 \times 608$. This strategy forces the model to learn scale-invariant representations, enabling deployment with different input sizes depending on the desired balance between speed and accuracy.

#### 1) YOLO9000

The authors of YOLOv2 further extend their model by introducing YOLO9000, a system designed to perform real-time detection across more than 9000 object categories. It retains the same underlying detection framework as YOLOv2, including the use of anchor boxes, passthrough layers, and multi-scale training, but introduces a novel mechanism for incorporating large-scale classification data into the detection pipeline.

At the core of YOLO9000 is a hierarchical classification system known as WordTree. Based on WordNet [59], it organizes object categories into a tree-like taxonomy, where each node represents a concept and each path from the root to a leaf defines a valid class label. Rather than predicting a flat set of class probabilities, the model estimates conditional probabilities along the paths of this hierarchy, enabling consistent predictions across multiple levels of abstraction.

To enable joint learning across detection and classification tasks, YOLO9000 adopts a unified training strategy. The model is trained simultaneously on detection datasets with bounding box annotations, such as COCO, and on classification datasets without localization information, such as ImageNet. For images lacking bounding boxes, only the classification loss is applied, enabling the model to benefit from the semantic richness of large-scale classification data. This joint optimization allows the model to generalize to object categories for which only image-level labels are available during training.

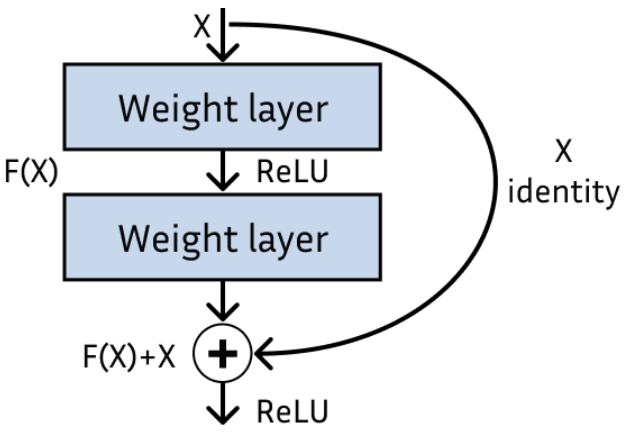

**FIGURE 6. Skip connections employed in ResNet, also adopted in the Darknet-53 backbone of YOLOv3.**

**TABLE 3. Structure of Darknet-53 backbone network used in YOLOv3.**

| Layer | Filters | Size | Repeat | Output |
|---|---|---|---|---|
| Image | - | - | - | $416 \times 416$ |
| Conv | 32 | $3 \times 3$ / 1 | 1 | $416 \times 416$ |
| Conv | 64 | $3 \times 3$ / 2 | 1 | $208 \times 208$ |
| Conv | 32 | $1 \times 1$ / 1 | Conv $\times$ 1 | $208 \times 208$ |
| Conv | 64 | $3 \times 3$ / 1 | Conv $\times$ 1 | $208 \times 208$ |
| Residual | - | - | Residual $\times$ 1 | $208 \times 208$ |
| Conv | 128 | $3 \times 3$ / 2 | 1 | $104 \times 104$ |
| Conv | 64 | $1 \times 1$ / 1 | Conv $\times$ 2 | $104 \times 104$ |
| Conv | 128 | $3 \times 3$ / 1 | Conv $\times$ 2 | $104 \times 104$ |
| Residual | - | - | Residual $\times$ 2 | $104 \times 104$ |
| Conv | 256 | $3 \times 3$ / 2 | 1 | $52 \times 52$ |
| Conv | 128 | $1 \times 1$ / 1 | Conv $\times$ 8 | $52 \times 52$ |
| Conv | 256 | $3 \times 3$ / 1 | Conv $\times$ 8 | $52 \times 52$ |
| Residual | - | - | Residual $\times$ 8 | $52 \times 52$ |
| Conv | 512 | $3 \times 3$ / 2 | 1 | $26 \times 26$ |
| Conv | 256 | $1 \times 1$ / 1 | Conv $\times$ 8 | $26 \times 26$ |
| Conv | 512 | $3 \times 3$ / 1 | Conv $\times$ 8 | $26 \times 26$ |
| Residual | - | - | Residual $\times$ 8 | $26 \times 26$ |
| Conv | 1024 | $3 \times 3$ / 2 | 1 | $13 \times 13$ |
| Conv | 512 | $1 \times 1$ / 1 | Conv $\times$ 4 | $13 \times 13$ |
| Conv | 1024 | $3 \times 3$ / 1 | Conv $\times$ 4 | $13 \times 13$ |
| Residual | - | - | Residual $\times$ 4 | $13 \times 13$ |

### C. YOLOv3

YOLOv3 [60], introduced in 2018, continues the single-stage detection approach and incorporates architectural changes to improve performance, especially on small objects. It addresses limitations of YOLOv2, such as reduced ability to detect small targets due to the loss of fine-grained features and weak gradient flow to earlier layers. YOLOv3 adopts a deeper, more structured architecture that supports multi-scale feature integration, while preserving fully convolutional predictions and anchor-based bounding box regression.

At the core of YOLOv3 is Darknet-53, a backbone that builds on the structure of Darknet-19 from YOLOv2. It integrates residual connections similar to those in ResNet, shown in Fig. 6, and consists of 53 convolutional layers with a combination of $3 \times 3$ and $1 \times 1$ filters. Shortcut connections improve gradient flow and support deeper network training. Each layer uses batch normalization and Leaky ReLU activation to ensure stability. Compared to ResNet-101 and ResNet-152, Darknet-53 offers similar accuracy with greater

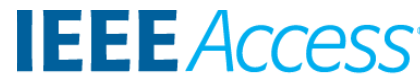

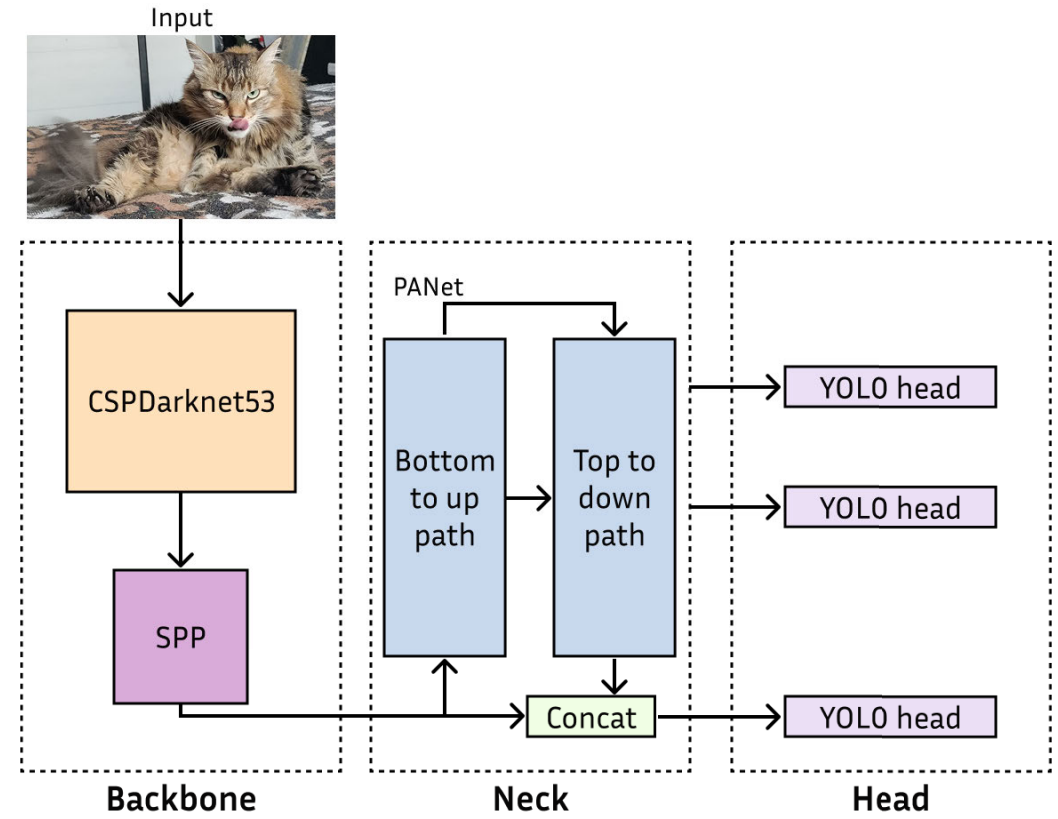

**FIGURE 7.** High-level overview of the YOLOv4 architecture.

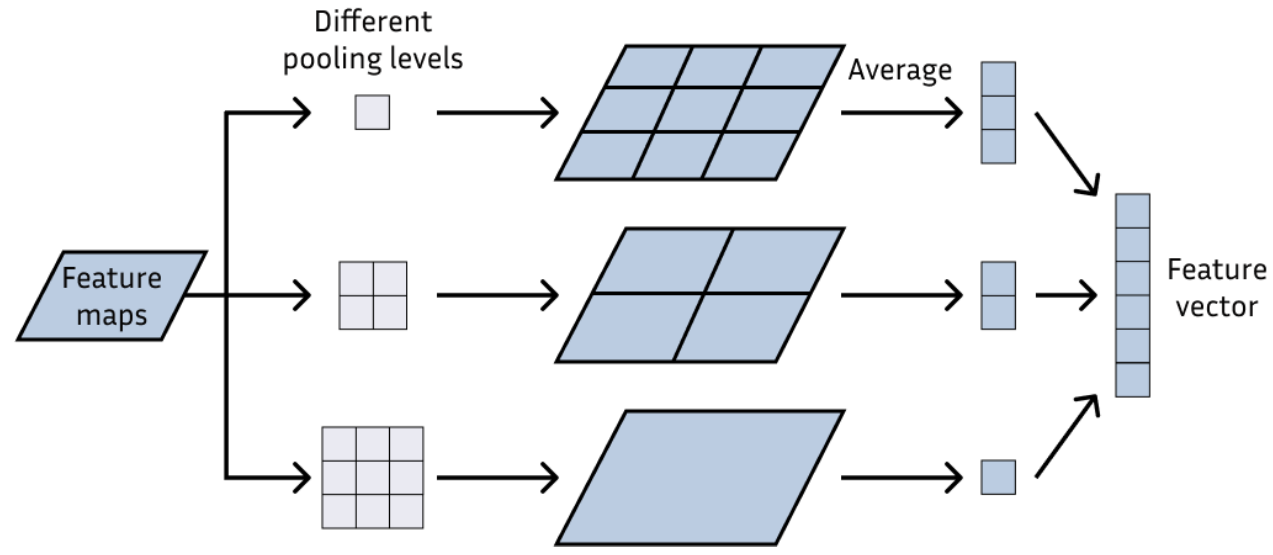

**FIGURE 8.** Spatial pyramid pooling operation.

speed and efficiency. Its complete architecture is shown in Table 3.

Another architectural advancement in YOLOv3 is the use of multi-scale predictions. The network predicts bounding boxes at three resolutions, $13 \times 13$, $26 \times 26$, and $52 \times 52$, by extracting features from different layers, following a design inspired by FPN. At each scale, a set of convolutional layers processes a specific feature map to produce bounding box coordinates, objectness scores, and class probabilities. To enrich the semantic content of the higher-resolution predictions, YOLOv3 upsamples coarser features and concatenates them with earlier feature maps before making predictions. This strategy enhances detection across object sizes, particularly for small targets.

Finally, class prediction in YOLOv3 replaces the softmax-based formulation of earlier versions with independent logistic classifiers for each class, treating the task as multi-label classification. This change is especially beneficial in datasets with overlapping or hierarchical labels, as it avoids enforcing mutual exclusivity. Additionally, binary cross-entropy loss is used during training, which improves performance in scenarios with ambiguous class boundaries.

### D. YOLOv4

YOLOv4 [61], introduced in 2020, advances real-time object detection within the YOLO framework by targeting practical deployment on conventional hardware. Unlike earlier versions that define architectures from scratch, YOLOv4 integrates state-of-the-art components and training strategies available at the time to achieve a balance between speed and accuracy. Its goal is to provide high-performance detection while remaining accessible for training and inference on a single GPU, without requiring large-scale infrastructure.

YOLOv4 organizes its improvements into two categories: the Bag of Freebies, which includes techniques that enhance accuracy without affecting inference speed, and the Bag of Specials, which introduces modules that improve performance with minor computational cost. This distinction guides the architectural and training choices throughout the model. The architecture consists of three modular components: a backbone for feature extraction, a neck for feature aggregation, and a head for dense prediction, as illustrated in Fig. 7. This design enables the integration of diverse optimizations while preserving the unified, single-shot detection paradigm of the YOLO family.

YOLOv4 adopts CSPDarknet-53 as its backbone, a variant of Darknet-53 enhanced with Cross Stage Partial (CSP) [62] connections. These connections split the input feature map, processing one part through a series of residual blocks while allowing the other to bypass them, merging the outputs later. The architecture comprises 29 convolutional layers with $3 \times 3$ filters and five CSP blocks. Batch normalization follows each convolutional layer, and the Mish activation [63] function is used throughout. The network processes the input through successive downsampling stages, generating feature maps at multiple spatial resolutions. This structure extracts hierarchical features while reducing computational cost. Compared to the standard Darknet-53, CSPDarknet-53 improves accuracy and reduces inference time by optimizing parameter utilization and gradient flow.

Following the backbone, YOLOv4 uses a neck that combines a Path Aggregation Network (PANet) [64] structure with Spatial Pyramid Pooling (SPP) [65]. PANet introduces additional bottom-up paths that complement the top-down flow, improving the fusion of semantic and localization features. This enhances the transmission of spatial detail to deeper layers. The SPP block increases the receptive field by applying parallel max-pooling operations with multiple kernel sizes, as shown in Fig. 8, enabling the network to capture context at various scales without changing the spatial dimensions of the feature maps. In YOLOv4, the SPP module is placed between the final convolutional layers and the detection head. The modified PANet implementation concatenates features instead of adding them, improving information retention compared to the original formulation.

At the head of the network, YOLOv4 retains the anchor-based dense prediction approach introduced used in YOLOv3. It continues to use predefined anchor boxes to regress bounding box coordinates, objectness scores, and class probabilities at multiple spatial resolutions. This strategy allows the model to effectively detect objects of varying sizes using the outputs from different levels of the neck.

Beyond its architectural design, YOLOv4 incorporates several training strategies and optimizations that contribute to its performance. Mosaic data augmentation combines four images into one during training, increasing object diversity and positional variance without enlarging the batch size. CutMix [66] and MixUp [67] create synthetic samples by merging pixel content and labels from different images, exposing the model to occlusions and label ambiguity. DropBlock is applied as structured regularization in place of standard dropout, better suited to convolutional layers. For bounding box regression, the model uses the Complete IoU (CIoU) loss [25], which accounts for overlap, center distance, and aspect ratio. Hyperparameter tuning is conducted via an automated genetic algorithm, enabling efficient search over the configuration space.

#### 1) SCALED YOLOv4

Scaled-YOLOv4 [68], introduced in 2021 by the authors of YOLOv4, extends the original architecture with a scalable design. Its goal is to support deployment across a range of model sizes and input resolutions without requiring manual redesign. The model retains key components from YOLOv4, including the CSPDarknet-53 backbone, the PANet-based neck, the YOLOv3-style detection head, and the modular structure. Instead of incorporating entirely new elements, Scaled-YOLOv4 scales the architecture along three dimensions: depth ($d$), width ($w$), and input resolution ($r$), allowing it to adapt to varying computational and application constraints.

The scaling method used in Scaled-YOLOv4 generalizes the compound scaling strategy from EfficientNet [69] to architectures with residual and cross-stage connections. It modifies three components of the network: the number of residual blocks ($d$), the number of channels per block ($w$), and the input image resolution ($r$). Each dimension is controlled by an independent coefficient that determines its growth as the model scales. Unlike architectures with a single computation path, YOLOv4 incorporates CSP connections that split the feature flow into residual and transition paths. Since these paths contribute differently to learning and computation, the scaling process adjusts them separately. One coefficient controls the number of residual blocks, while another governs the scaling of transition layers. This separation preserves computational efficiency and avoids unbalanced growth. As a result, the model can be expanded or reduced predictably, maintaining consistent behaviour across a range of deployment scenarios.

Scaled-YOLOv4 is instantiated in three main variants: P5, P6, and P7, corresponding to input resolutions of 896, 1280, and 1536, respectively. Each version preserves the same architectural layout but varies in depth and width according to the scaling coefficients.

### E. YOLOv5

YOLOv5 [70], released in 2020, introduces a significant shift in the evolution of the YOLO family, as it is the first version developed and maintained by the company Ultralytics. Unlike previous versions implemented in the Darknet framework with C and CUDA, YOLOv5 is entirely written in PyTorch, improving accessibility and extensibility for the deep learning community. The model preserves the modular architecture of its predecessors, structured around a backbone for feature extraction, a neck for feature aggregation, and a head for dense object prediction, as illustrated in Fig. 9.

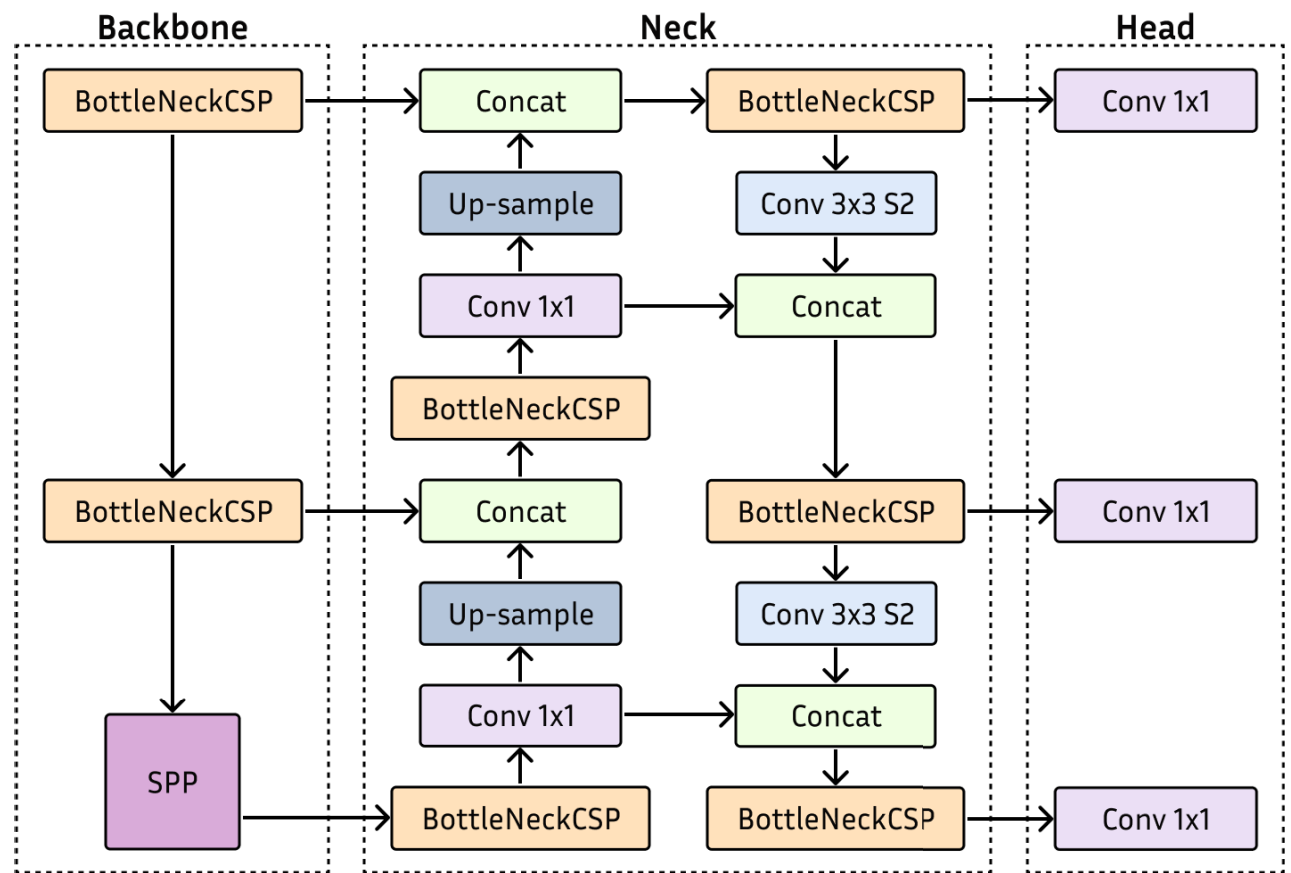

**FIGURE 9.** **YOLOv5 architecture.**

YOLOv5 retains several architectural components introduced in YOLOv4. The backbone remains CSPDarknet-53, which combines the Darknet-53 architecture with CSP connections to improve gradient flow and reduce computational redundancy. As previously described, this backbone includes residual blocks, batch normalization, and the Mish activation function across the feature extraction layers. It processes the input image through successive downsampling stages to generate multi-scale feature maps used by the neck and head of the network.

The neck of YOLOv5 introduces modifications that differentiate it from earlier versions. It uses a customized variant of PANet, where feature maps are concatenated instead of summed, as in the original design. Several CSP layers are added within the PANet structure to improve gradient flow and feature propagation without significant computational overhead. This CSP-PANet configuration enhances the balance between semantic richness and spatial precision during feature aggregation. YOLOv5 also replaces the SPP module used in YOLOv4 with a more efficient version called Spatial Pyramid Pooling Fast (SPPF). Instead of applying multiple max-pooling operations in parallel, SPPF applies them sequentially using the same kernel size, as shown in Fig. 10. This strategy reduces computation while preserving the benefits of a large receptive field.

The head of YOLOv5 reuses the detection head introduced in YOLOv3. It uses anchor-based prediction layers applied at three spatial resolutions and performs object localization and classification in a fully convolutional manner. YOLOv5 employs the CIoU loss function for bounding box regression, applies an Exponential Moving Average (EMA) strategy to stabilize training, and includes automatic anchor box learning. The training pipeline incorporates data augmentation techniques such as Mosaic augmentation and HSV-space image transformations, which increase sample diversity and improve generalization. YOLOv5 is released in five model variants, with architectural details summarized in Table 4.

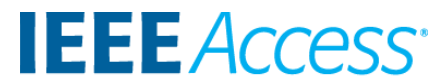

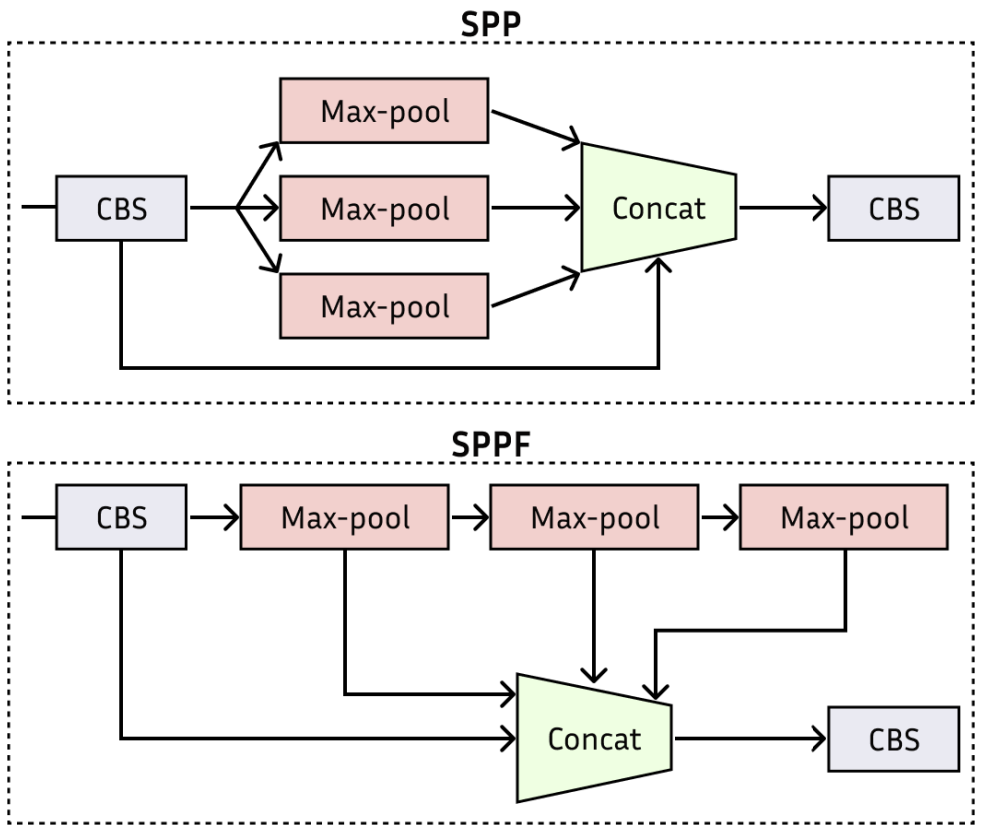

**FIGURE 10.** Comparison between the SPP module used in YOLOv4 and the SPPF module employed in YOLOv5. CBS stands for Convolution + Banch Normalization + Sigmoid Linear Unit (SiLU) activation function.

**TABLE 4.** Details of YOLOv5 variants. Speed is measured using a V100 GPU on the COCO dataset.

| Model | Size (pixels) | mAP (50-95) | mAP (50) | Speed (ms) | Params (M) | FLOPs (B) |
|---|---|---|---|---|---|---|
| YOLOv5n | 640 | 28.0 | 45.7 | 6.3 | 1.9 | 4.5 |
| YOLOv5s | 640 | 37.4 | 56.8 | 6.4 | 7.2 | 16.5 |
| YOLOv5m | 640 | 45.4 | 64.1 | 8.2 | 21.2 | 49 |
| YOLOv5l | 640 | 49.0 | 67.3 | 10.1 | 46.5 | 109.1 |
| YOLOv5x | 640 | 50.7 | 68.9 | 12.1 | 86.7 | 205.7 |

### F. PP-YOLO

PP-YOLO [71] is introduced in 2020 by Baidu as a derivative of YOLOv3, developed within the PaddlePaddle[6] deep learning framework, from which it takes its name. It retains the structural foundations of YOLOv3 and focuses on practical enhancements. Its objective is to improve the trade-off between accuracy and efficiency by integrating a set of established techniques that enhance detection performance with minimal impact on model complexity. Although it is not an official successor in the YOLO series, its alignment with the core design principles of YOLO justifies its inclusion within the broader landscape of YOLO-based approaches.

Regarding the backbone, PP-YOLO replaces the original Darknet-53 used in YOLOv3 with a variant of ResNet50 known as ResNet50-vd. This is motivated by the widespread use of ResNet architectures across deep learning frameworks, which facilitates deployment and potentially improves runtime efficiency. To compensate for the performance degradation caused by this substitution, the model introduces deformable convolutional layers in the final stage of the network. Specifically, the 3 × 3 convolutions in the last residual block are replaced with deformable convolutions, resulting in the backbone referred to as ResNet50-vd-dcn. This change enhances the network's capacity to capture geometric variation without significantly increasing the number of parameters or computational cost.

Beyond the change in backbone, PP-YOLO preserves the structural layout of YOLOv3. It maintains the original detection neck based on FPN, as well as the multi-scale detection heads and the anchor-based output format. The detection head architecture remains unchanged, consisting of a pair of convolutional layers at each scale that produce class probabilities, bounding box regressions, and objectness scores.

The core contribution of PP-YOLO lies in its systematic integration of established techniques to improve detection accuracy without compromising inference efficiency. The model incorporates training enhancements such as increased batch size, EMA for parameter stabilization, and DropBlock for regularization. It also introduces loss-related improvements, including an auxiliary IoU loss branch and IoU-aware confidence scoring. At the inference stage, refinements like grid sensitivity calibration and Matrix NMS enhance output quality. Lightweight modules such as CoordConv and SPP are added selectively to preserve computational efficiency. These combined modifications result in a significant accuracy improvement over YOLOv3 while maintaining faster inference than models like YOLOv4.

Subsequent iterations of the model led to the release of PP-YOLOv2 [72] in 2021 and PP-YOLOE [73] in 2022, both of which introduced further refinements to the original design, although their impact remained largely confined to the PaddlePaddle ecosystem.

### G. YOLOR

You Only Learn One Representation (YOLOR) [74], introduced in 2021 by the authors of YOLOv4 and Scaled-YOLOv4, proposes a unified model capable of supporting multiple computer vision tasks. Built upon Scaled-YOLOv4, YOLOR addresses a core limitation of conventional convolutional neural networks: features learned for one task, such as object detection, often fail to generalize to others like segmentation or classification. To overcome this, YOLOR introduces a mechanism that fuses explicit knowledge, derived directly from the input, with implicit knowledge acquired in a latent, task-independent manner. This design draws inspiration from human cognition, where reasoning is informed by both conscious and subconscious processing. The resulting unified representation supports dynamic adaptation to multiple vision tasks, enabling a shared feature space optimized for general-purpose visual understanding. The overall architecture is illustrated in Fig. 11.

YOLOR incorporates three core mechanisms: kernel space alignment, prediction refinement, and multi-task learning. Kernel space alignment mitigates inconsistencies between feature distributions across different tasks or prediction heads by transforming outputs through additive or multiplicative interactions with an implicit representation. This allows the feature space to be adjusted via translation, scaling, or rotation for improved alignment. Prediction refinement injects implicit representations into the output layers of the detection head, enabling the model to adjust predictions such as object coordinates or anchor dimensions based on latent patterns. For multi-task learning, implicit representations are introduced into each task-specific branch, enhancing the shared feature space with task-adaptive signals. These mechanisms operate within the same backbone and can be executed concurrently, allowing the network to support detection, classification, and embedding tasks in a single forward pass.

[6] https://github.com/PaddlePaddle/Paddle

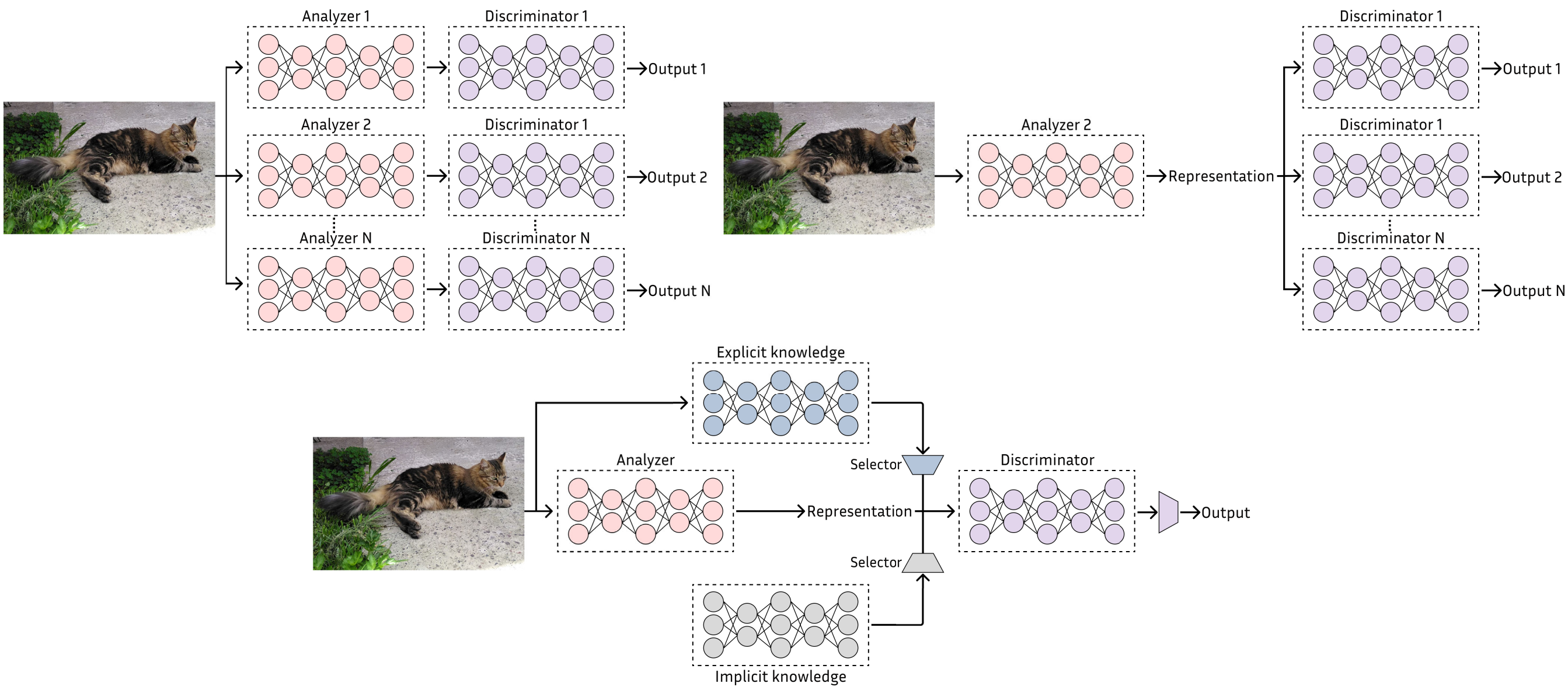

**FIGURE 11.** Comparison between a framework using distinct models for distinct tasks (top-left), a framework employing a shared backbone with different heads for different tasks (top-right), and the unified framework proposed in YOLOR that integrates explicit and implicit knowledge (bottom).

**TABLE 5.** Details of YOLOvR variants. Speed is measured using a V100 GPU on the COCO dataset.

| Model | Size (pixels) | mAP (50) | FPS | Params (M) | FLOPs (G) |
|---|---|---|---|---|---|
| YOLOR-P6 | 1280 | 71.8 | 76 | 37 | 326 |
| YOLOR-W6 | 1280 | 73.2 | 66 | 80 | 454 |
| YOLOR-E6 | 1280 | 74.1 | 45 | 116 | 684 |
| YOLOR-D6 | 1280 | 75.0 | 34 | 152 | 937 |

At the core of YOLOR is its method for modeling implicit knowledge. Unlike explicit knowledge, which is directly derived from input data, implicit knowledge is treated as latent information that is independent of observation and shared across tasks. This latent representation is defined as a constant tensor $Z = \{z_1, z_2, \cdots, z_k, \}$ and is injected into the network through learnable transformations. The model supports three ways to encode this information: (1) as a vector with independent dimensions; (2) as a neural network, where each dimension is dependent and the representation is generated via a learned mapping $Wz$, with $z$ treated as a latent prior and $W$ as a mapping matrix that transforms it into the implicit representation; and (3) through matrix factorization, which constructs a composite representation using latent bases and coefficients. These options provide flexibility in adapting implicit priors to different tasks. During training, both the implicit representation and its transformation parameters are learned via backpropagation. At inference time, the implicit component is fixed, adding negligible computational overhead.

The training objective in YOLOR extends the conventional neural network formulation by incorporating both explicit and implicit knowledge into the modeling of error. Rather than minimizing a single error term, the model defines a composite objective composed of explicit error $\epsilon_{ex}(x)$ and implicit error $\epsilon_{im}(z)$, combined through a function $g_\phi$. This structure enables the network to learn general-purpose representations applicable across tasks. To support this integration, YOLOR introduces three types of operators: addition, multiplication, and concatenation, used to merge explicit and implicit representations. The operator depends on the functional role of the target layer. Addition is applied in feature alignment modules to introduce implicit offsets that shift feature responses. Multiplication is used in prediction refinement to scale anchor dimensions and adjust bounding box sizes. Concatenation is employed in multi-task branches to enrich the feature space with task-specific latent information. These operations are selected based on the computational context of each module and allow the integration of implicit knowledge without increasing inference cost.

YOLOR is implemented in four principal variants, each corresponding to a different model size and computational target. These configurations differ in depth, width, and input resolution, and are summarized in Table 5. While the initial release included these four versions, additional variants were later introduced in the official repository to support broader deployment scenarios.

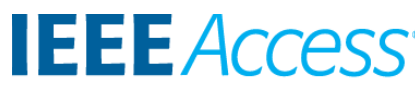

### H. YOLOX

YOLOX [75] is introduced in 2021. Unlike previous versions such as YOLOv4 and YOLOv5, which follow an anchor-based design, YOLOX adopts an anchor-free approach. This change simplifies model design, reduces computational complexity, and improves flexibility during training and deployment. The model builds upon YOLOv3, specifically using the Darknet-53 backbone with an SPP layer. From this starting point, YOLOX incorporates a series of modifications that reflect recent advancements in object detection.

To address the limitations associated with anchor-based detection, YOLOX adopts an anchor-free approach. Instead of generating multiple predictions per location using predefined anchor boxes, the model predicts a single bounding box at each spatial position. Each prediction encodes two offsets from the top-left corner of the grid cell and the width and height of the bounding box. Positive samples are assigned to the center of each ground-truth object, and a fixed scale range is used to associate objects with specific levels of the FPN. This design reduces heuristic parameters, simplifies training and decoding, and lowers memory usage, making the model more suitable for deployment on resource-constrained devices.

YOLOX also revises the label assignment strategy to allow multiple positive samples per object, rather than restricting supervision to a single location. In the anchor-free setting, the default approach assigns the object's center point as the sole positive sample. However, this ignores nearby high-quality predictions that could provide useful gradients. To mitigate this, YOLOX defines a $3 \times 3$ region around the object center and marks all grid cells within it as positives. This increases the number of positive samples per object and helps reduce the class imbalance between positive and negative examples during training.

Following the multi-positive assignment strategy, YOLOX introduces SimOTA, a simplified variant of Optimal Transport Assignment (OTA), to enable adaptive and global label matching. SimOTA computes a pairwise cost for each prediction–ground-truth pair by combining classification and regression losses via a weighted sum. For each ground-truth, the top-k predictions with the lowest cost values within a predefined center region are selected as positive samples. The value of k varies dynamically according to object characteristics, allowing flexible and accurate assignment. Unlike the original OTA, SimOTA avoids the computational overhead of solving the full transport problem with the Sinkhorn-Knopp algorithm, making it efficient for training while retaining the benefits of global and adaptive matching.

Additionally, YOLOX adopts a decoupled detection head to mitigate conflicts between classification and regression tasks. Previous YOLO versions use a single shared branch (coupled head) for both objectives, which can lead to suboptimal learning due to task interference. In contrast, YOLOX applies a $1 \times 1$ convolution to reduce channel dimensions, followed by two separate branches of two $3 \times 3$ convolutional layers each: one for classification and one for regression, as illustrated in Fig. 12. An extra IoU prediction branch is attached to the regression path. This structural separation enables each task to learn specialized features, improving optimization and convergence. It also plays a crucial role in supporting the end-to-end variant of YOLOX, where the coupled head has been shown to limit performance.

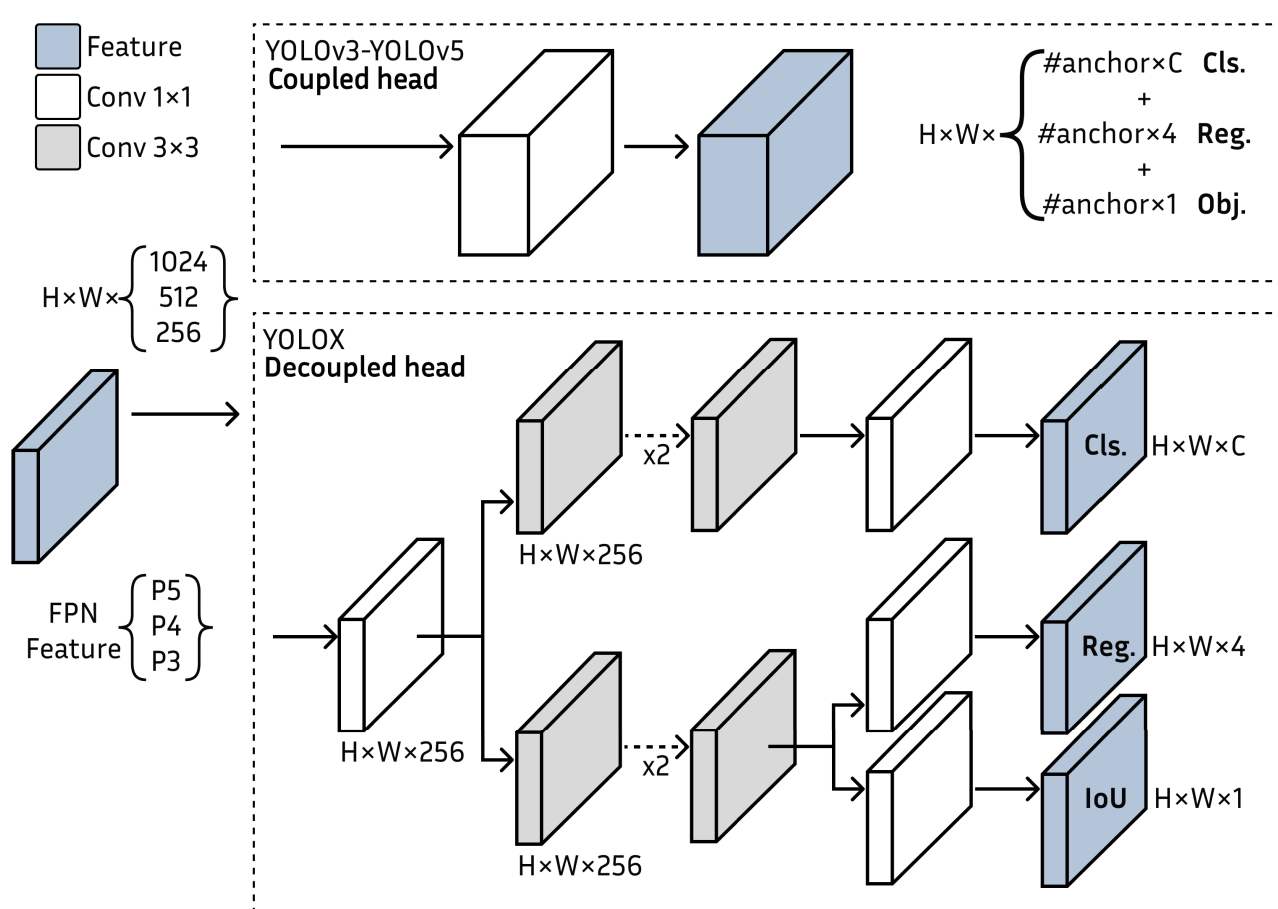

**FIGURE 12. Comparison between the coupled head used in YOLOv3-v5 versions and the decoupled head introduced in YOLOX.**

To improve training effectiveness, YOLOX uses Mosaic and MixUp augmentations, allowing the model to learn from more varied contexts and object scales. These augmentations are applied during training and disabled in the final epochs to aid convergence. The implementation modifies standard MixUp by incorporating random scale jittering before combining images, drawing inspiration from the Copy-Paste method without requiring instance masks. The augmentation strategy is adapted according to model size: larger models use stronger augmentation, while in smaller variants such as YOLOX-Nano, MixUp is omitted and the Mosaic scale range is reduced.

YOLOX also includes an optional end-to-end configuration that eliminates the need for NMS. This is achieved by incorporating one-to-one label assignment and stop-gradient operations, allowing the model to be trained directly for final predictions without relying on post-processing. Although this setup simplifies the inference pipeline, it introduces a slight performance trade-off and is not used in the default configurations. The YOLOX family includes multiple model variants tailored to different deployment requirements. Details are summarized in Table 6.

### I. YOLOv6

YOLOv6 [76], released in 2022 by Meituan, introduces a series of architectural and training-level modifications aimed at improving performance and deployment efficiency in real-time object detection. Unlike previous versions that focus on general-purpose detection, YOLOv6 emphasizes practical applicability in industrial scenarios. The model follows a modular structure composed of a backbone, neck, and head, as illustrated in Fig. 13.

**TABLE 6.** Details of YOLOvX variants. Speed is measured using a V100 GPU on the COCO dataset.

| Model | Size (pixels) | mAP (50-95) | Speed (ms) | Params (M) | FLOPs (B) |
|---|---|---|---|---|---|
| YOLOX-s | 640 | 40.5 | 9.8 | 9.0 | 26.8 |
| YOLOX-m | 640 | 46.9 | 12.3 | 25.3 | 73.8 |
| YOLOX-l | 640 | 49.7 | 14.5 | 54.2 | 155.6 |
| YOLOX-x | 640 | 51.1 | 17.3 | 99.1 | 281.9 |

**TABLE 7.** Details of YOLOv6 variants. Speed is measured using a T4 GPU on the COCO dataset.

| Model | Size (pixels) | mAP (50-95) | FPS (fp16) | Params (M) | FLOPs (G) |
|---|---|---|---|---|---|
| YOLOv6n | 640 | 37.5 | 779 | 4.7 | 11.4 |
| YOLOv6s | 640 | 45.0 | 339 | 18.5 | 45.3 |
| YOLOv6m | 640 | 50.0 | 175 | 34.9 | 85.8 |
| YOLOv6l | 640 | 52.8 | 98 | 59.6 | 150.7 |

The backbone in YOLOv6 varies depending on the model scale. For lightweight configurations, it adopts the EfficientRep backbone, inspired by RepVGG, which integrates reparameterizable convolutional blocks (RepConv) to balance training flexibility and inference efficiency. During training, RepConv uses a multi-branch structure with identity and $1 \times 1$ convolutions, later fused into a single $3 \times 3$ convolution at inference to reduce latency, as shown in Fig. 14. In larger variants, the backbone switches to CSPStackRep, which builds on the CSP mechanism combined with stacked RepConv blocks. This structure enhances representational capacity while preserving the benefits of reparameterization. In all cases, the backbone generates multi-scale feature maps that feed into the neck.

The neck of YOLOv6 builds on the PANet structure used in previous versions, with adaptations that improve inference speed and feature representation. Its core component, Rep-PAN, replaces standard convolutional blocks with reparameterized RepBlocks. In larger variants, these are extended into CSPStackRep blocks to match the backbone design. The architecture preserves the top-down and bottom-up fusion strategy of PANet while leveraging reparameterization to accelerate inference without degrading multi-scale feature aggregation. The neck produces three output feature maps at different resolutions, which are passed to the head for prediction.

The head of YOLOv6 adopts an anchor-free design and employs the Efficient Decoupled Head structure. Unlike the coupled heads in earlier YOLO versions, it separates the classification and regression branches, assigning each its own convolutional layers. The design uses a hybrid channel strategy in which initial layers are shared to reduce redundancy, while final layers are decoupled to focus on their respective tasks. This separation improves convergence stability and supports better optimization of the distinct objectives in object detection.

YOLOv6 introduces a series of training and deployment-level enhancements aimed at improving both learning performance and hardware efficiency. For classification, it adopts Varifocal Loss (VFL) [77], which assigns dynamic weights to positive and negative samples based on prediction confidence to address class imbalance. For bounding box regression, it uses Distribution Focal Loss (DFL) [78] in larger variants, predicting discrete probability distributions to improve localization precision. The model also applies self-distillation jointly to classification and regression branches, using a pre-trained teacher to guide the student during training. For deployment on constrained devices, YOLOv6 incorporates structural reparameterization at inference time, supports Quantization-aware Training (QAT), and applies channel-wise knowledge distillation along with selective float fallback for sensitive layers. These strategies allow efficient model conversion and execution on target platforms without significant loss in accuracy. YOLOv6 is released in four main variants, detailed in Table 7.

### *J. YOLOv7*

YOLOv7 [79], released in 2022 by the authors of YOLOv4, introduces a series of architectural and training-level refinements aimed at improving both accuracy and efficiency. It was released only a few months before YOLOv6, making it a parallel development rather than a direct successor. The model retains the modular structure of backbone, neck, and head, as shown in Fig. 15, and includes optimizations that unify the training framework for both auxiliary and lead heads, enabling a more efficient and scalable design.

The backbone of YOLOv7 is based on E-ELAN, an extended version of the Efficient Layer Aggregation Network (ELAN) [80]. It is designed to enhance feature learning while preserving gradient flow and computational efficiency through a scalable composition of deeper layers. E-ELAN draws from two prior architectures: CSPNet, used in earlier YOLO models to improve gradient propagation and feature reuse, and VoVNet [81], which replaces dense connections with One-Shot Aggregation (OSA) modules. These modules aggregate outputs from several convolutional layers in a single step, improving efficiency and GPU utilization. E-ELAN also incorporates expansion, shuffling, and merging operations to support flexible hierarchical feature aggregation. Unlike earlier YOLO versions, where deeper networks risked degradation or vanishing gradients, E-ELAN enables stable training even in architectures with extensive depth.

The neck in YOLOv7 is constructed using a PANet combined with FPN (PAN-FPN) design, which facilitates bidirectional feature flow across different scales. This structure enhances the fusion of semantic features from deeper layers with localization-rich information from shallower layers. YOLOv7 also integrates E-ELAN modules within the neck to maintain consistency with the backbone's architectural pattern and to further reinforce feature aggregation at each scale. The neck generates multi-resolution feature maps that are optimized for the subsequent detection head.

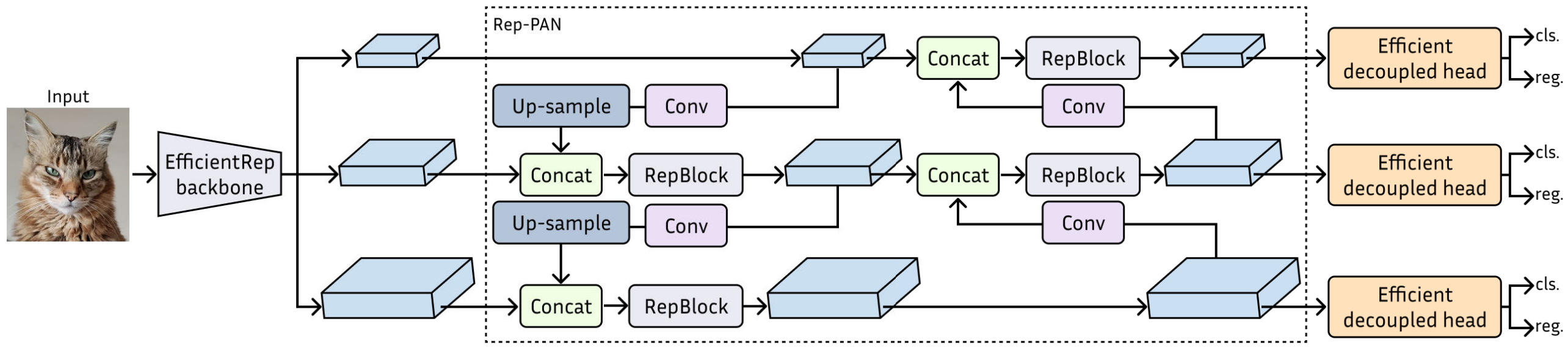

**FIGURE 13.** YOLOv6 architecture. This diagram corresponds to the nano and small variants of the architecture. For the complex variants, RepBlocks are replaced with CSPStackRep.

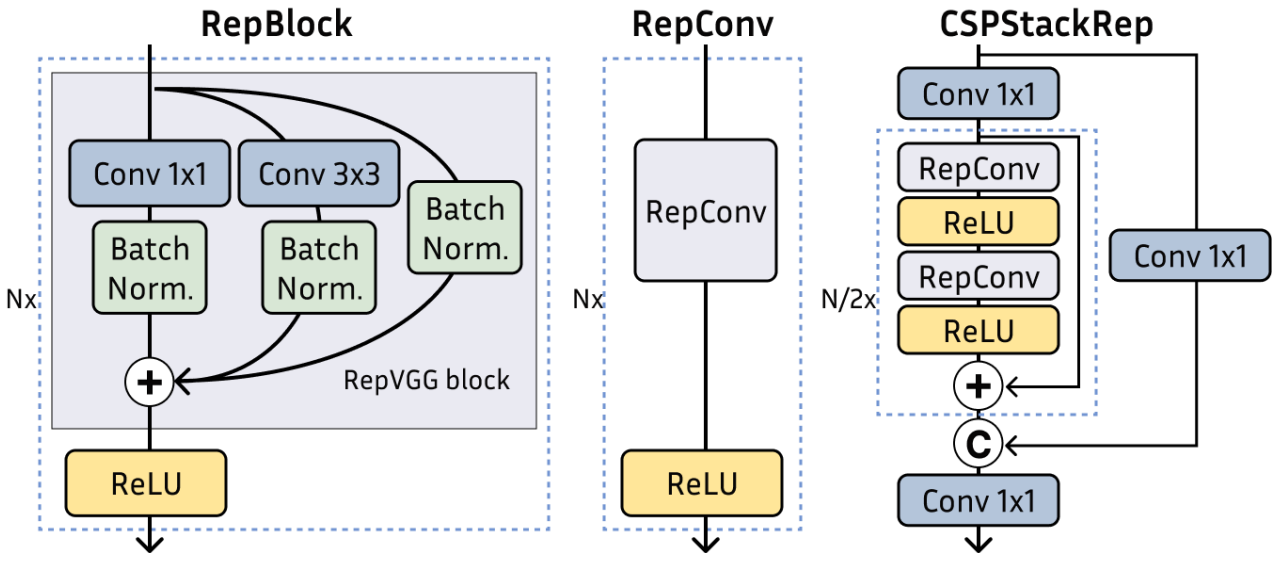

**FIGURE 14.** Modules used in YOLOv6. RepBlock (left), used during training; RepConv (center), used during inference; CSPStackRep block (right), composed of stacked RepBlocks within a CSP structure, used in larger models.

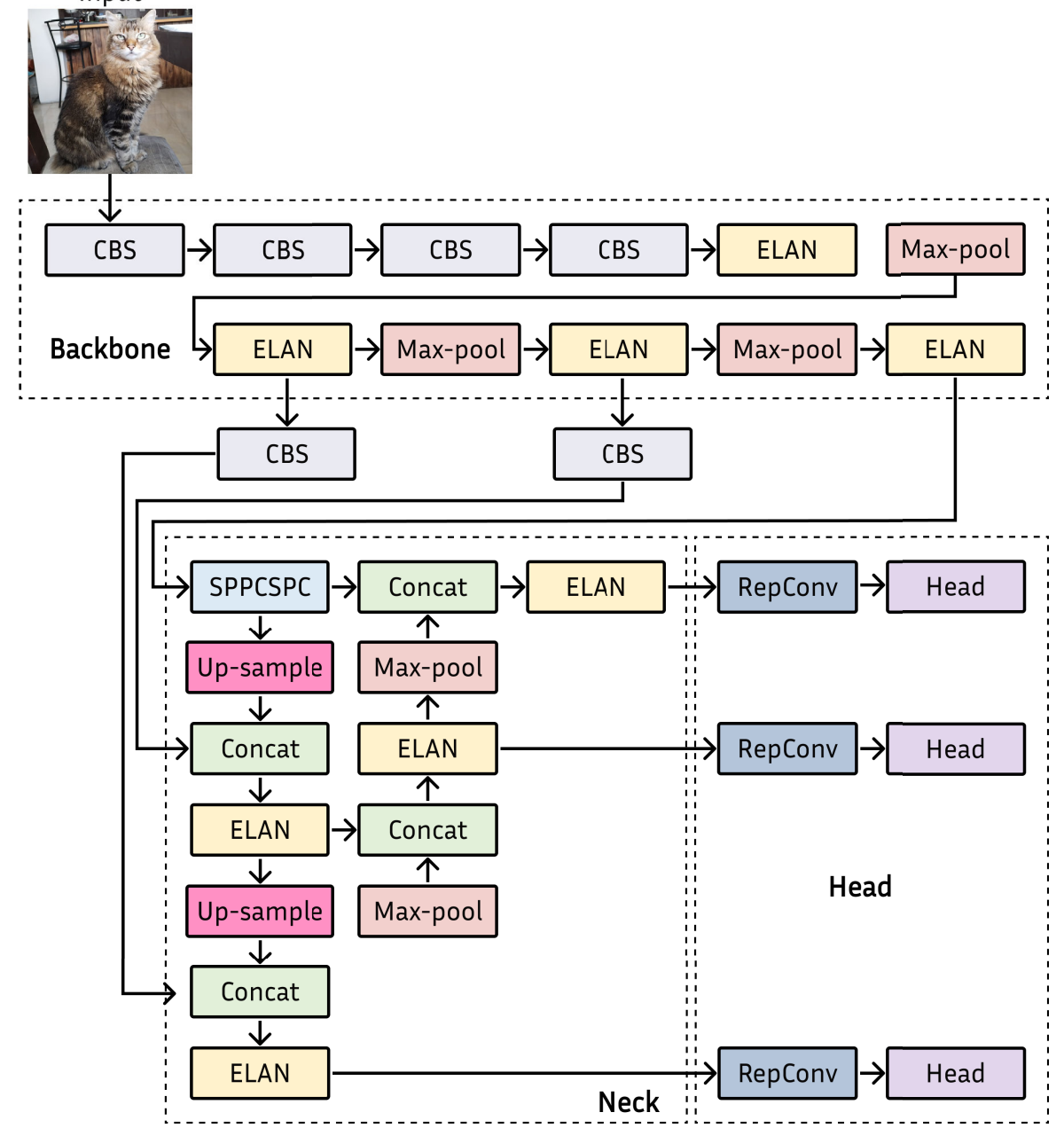

**FIGURE 15.** YOLOv7 architecture.

Another key inclusion in YOLOv7 is the planned reparameterization strategy, which adapts RepConv for use in architectures that include residual and concatenation structures. The original RepConv module, composed of a $3 \times 3$ convolution, a $1 \times 1$ convolution, and an identity mapping, interferes with residual and concatenation pathways by limiting gradient diversity. To address this, the authors introduce RepConvN, a variant that removes the identity connection to ensure compatibility with architectures such as ResNet and DenseNet. The framework also integrates model weight averaging, where parameters from multiple training instances, obtained from different data subsets or training stages, are averaged to produce a more stable inference model. Gradient flow propagation paths guide the selection of modules where these strategies are applied.

Finally, the head in YOLOv7 incorporates an Auxiliary Head Coarse-to-Fine mechanism, which extends the network with an auxiliary head to support the main lead head and enable deep supervision during training. Instead of assigning labels independently, the system generates soft labels by combining the predictions of the lead head with ground-truth annotations. These labels are then used to supervise both heads. The Coarse-to-Fine mechanism, shown in Fig. 16, defines two types of labels: fine labels, used by the lead head, are derived under strict matching conditions, while coarse labels, assigned to the auxiliary head, are generated with relaxed constraints to increase positive sample coverage. This arrangement improves recall in early layers without compromising final precision. Decoder-level constraints limit the influence of coarse predictions, ensuring that the lead head retains control over the final outputs.

As a notable detail, YOLOv7 models are trained entirely from scratch using only the MS-COCO dataset, without any pretraining on ImageNet. The architecture includes two main variants, YOLOv7 and YOLOv7-X, both operating at $640 \times 640$ input resolution, along with additional scaled versions that are designed for higher input sizes and specific hardware constraints. Details of these are provided in Table 8.

### *K. YOLOv8*

YOLOv8 [82], released in 2023 by Ultralytics, retains the modular structure established in YOLOv5, consisting of a backbone, neck, and detection head, as shown in

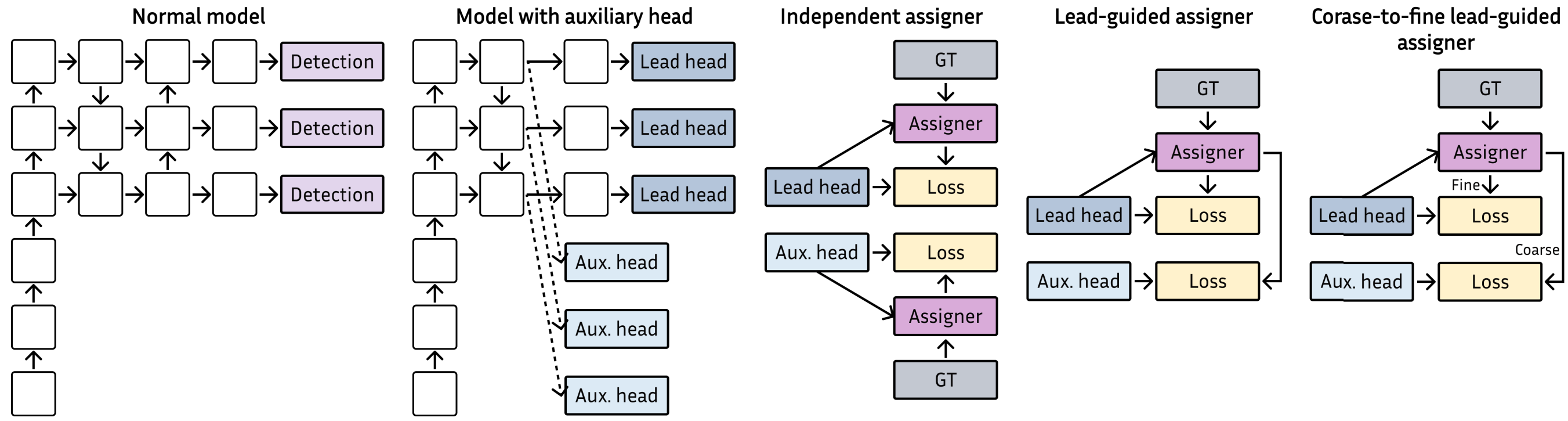

**FIGURE 16. Coarse-to-fine mechanism for auxiliary and fine for lead head label assigner utilized in YOLOv7.**

**TABLE 8. Details of YOLOv7 variants. Speed is measured on a V100 GPU on the COCO dataset.**

| Model | Size (pixels) | mAP (50) | FPS | Params (M) | FLOPs (G) |
|---|---|---|---|---|---|
| YOLOv7 | 640 | 69.7 | 161 | 36.9 | 104.7 |
| YOLOv7x | 640 | 71.2 | 114 | 71.3 | 189.9 |
| YOLOv7-W6 | 1280 | 72.6 | 84 | 70.04 | 360.0 |
| YOLOv7-E6 | 1280 | 73.5 | 56 | 97.2 | 515.2 |
| YOLOv7-D6 | 1280 | 74.0 | 44 | 154.7 | 806.8 |
| YOLOv7-E6E | 1280 | 74.4 | 36 | 151.7 | 843.2 |

**TABLE 9. Details of YOLOv8 variants. Speed is measured on a A100 GPU on the COCO dataset.**

| Model | Size (pixels) | mAP (50-95) | Speed (ms) | Params (M) | FLOPs (B) |
|---|---|---|---|---|---|
| YOLOv8n | 640 | 37.3 | 0.99 | 3.2 | 8.7 |
| YOLOv8s | 640 | 44.9 | 1.20 | 11.2 | 28.6 |
| YOLOv8m | 640 | 50.2 | 1.83 | 25.9 | 78.9 |
| YOLOv8l | 640 | 52.9 | 2.39 | 43.7 | 165.2 |
| YOLOv8x | 640 | 53.9 | 3.53 | 68.2 | 257.8 |

Fig. 17. The backbone continues to use CSPDarknet-53, which incorporates CSP connections to improve gradient flow and computational efficiency. In addition, YOLOv8 supports alternative backbones such as EfficientDet, offering flexibility to meet different performance and efficiency requirements. The neck combines PANet with SPPF for multi-scale feature aggregation, enabling the fusion of spatial information across different resolutions. The detection head maintains the dense configuration with decoupled branches for classification and regression.

Despite maintaining the same architectural foundation as YOLOv5, YOLOv8 introduces several innovations and design changes. A major shift is the adoption of an anchor-free detection paradigm. Rather than relying on predefined anchor boxes to guide bounding box regression, YOLOv8 predicts object centers directly. This modification simplifies training and enhances generalization, particularly on custom datasets where anchor priors may not align well with the data distribution. Additionally, by reducing the number of candidate boxes per image, the anchor-free design decreases computational load during NMS, improving inference speed without compromising detection accuracy.

Another architectural refinement in YOLOv8 is the replacement of the C3 module in the backbone with a new structure called C2f. Whereas the C3 module outputs only the final bottleneck block, C2f concatenates the outputs of all bottleneck blocks within the module, as illustrated in Fig. 18. This design increases the effective receptive field and enables the integration of richer contextual information at each stage. By altering how intermediate features are aggregated, C2f improves the backbone's representational capacity while preserving computational efficiency.

In addition to structural changes, YOLOv8 introduces training-level enhancements. It employs Mosaic data augmentation, increasing variability and promoting generalization across diverse object contexts. CutMix and MixUp are also utilized to generate interpolations between training samples, enhancing robustness. For optimization, the model uses CIoU and DFL for bounding box regression, combined with binary cross-entropy for classification. These strategies improve convergence and overall detection performance across a variety of scenarios. YOLOv8 is released in five detection variants, each targeting different trade-offs between computational cost and accuracy; their details are summarized in Table 9.

### *L. YOLO-NAS*

YOLO-NAS [83], introduced in 2023 by Deci.ai, is a real-time object detection model designed to improve the accuracy-latency trade-off and enable efficient deployment on edge devices. Its name refers to Neural Architecture Search (NAS), a technique that automates network design by optimizing over a defined search space rather than relying on fixed heuristics. YOLO-NAS focuses on maintaining fast inference while improving performance, particularly on small object detection.

The architecture of YOLO-NAS is produced using AutoNAC, Deci.ai's proprietary NAS engine. This method replaces manual design with a hardware- and data-aware optimization process that explores architectural configurations under task-specific constraints. AutoNAC evaluates candidates across a defined search space, varying block types, the number of blocks per stage, and channel dimensions. The search is guided toward the efficiency frontier, which represents optimal trade-offs between accuracy and latency. Unlike generic NAS systems, AutoNAC considers the full inference stack, including quantization, compilation, and hardware compatibility, ensuring that the resulting architectures are both accurate and deployment-ready.

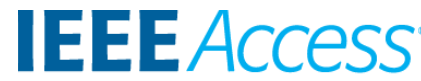

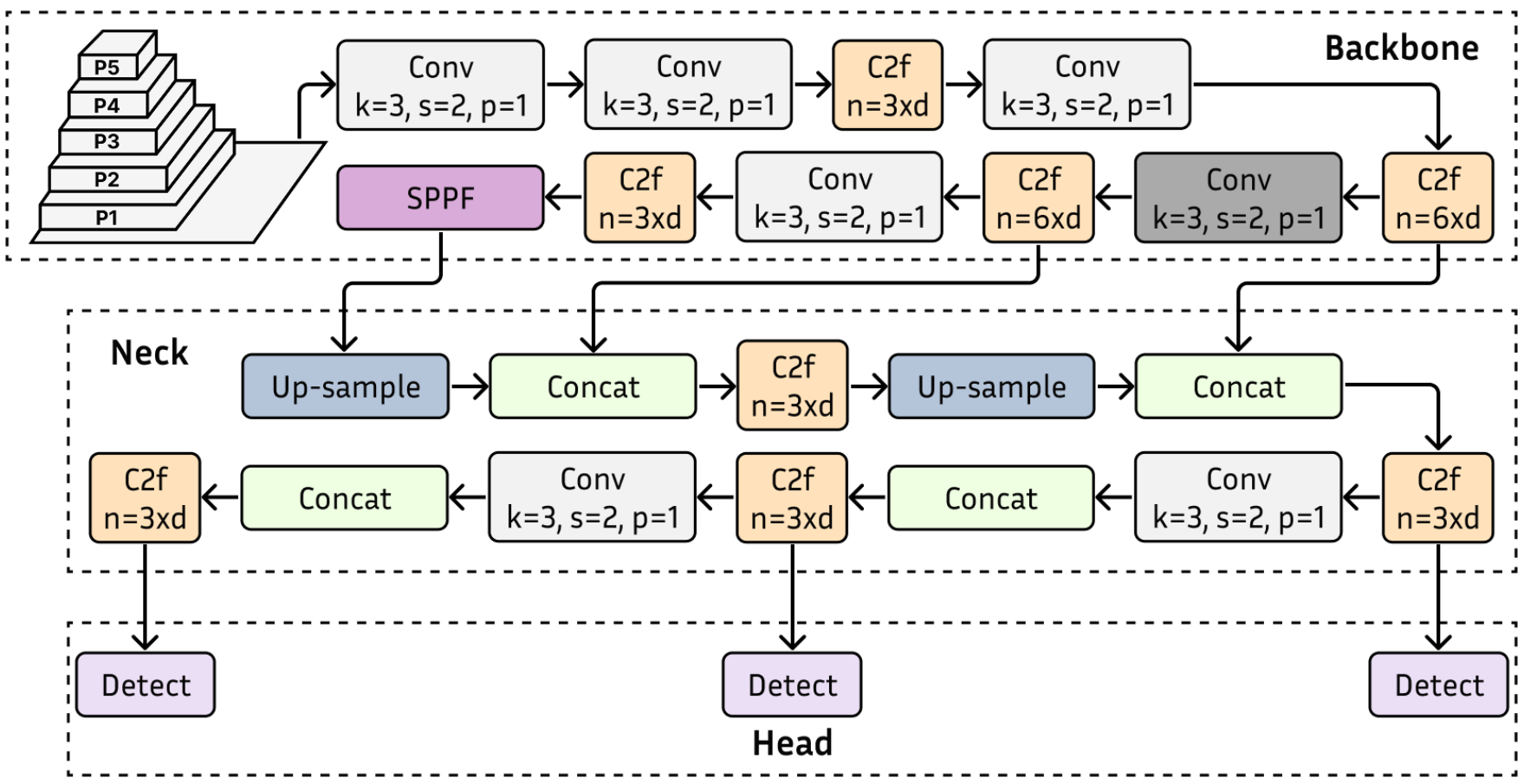

**FIGURE 17.** YOLOv8 architecture.

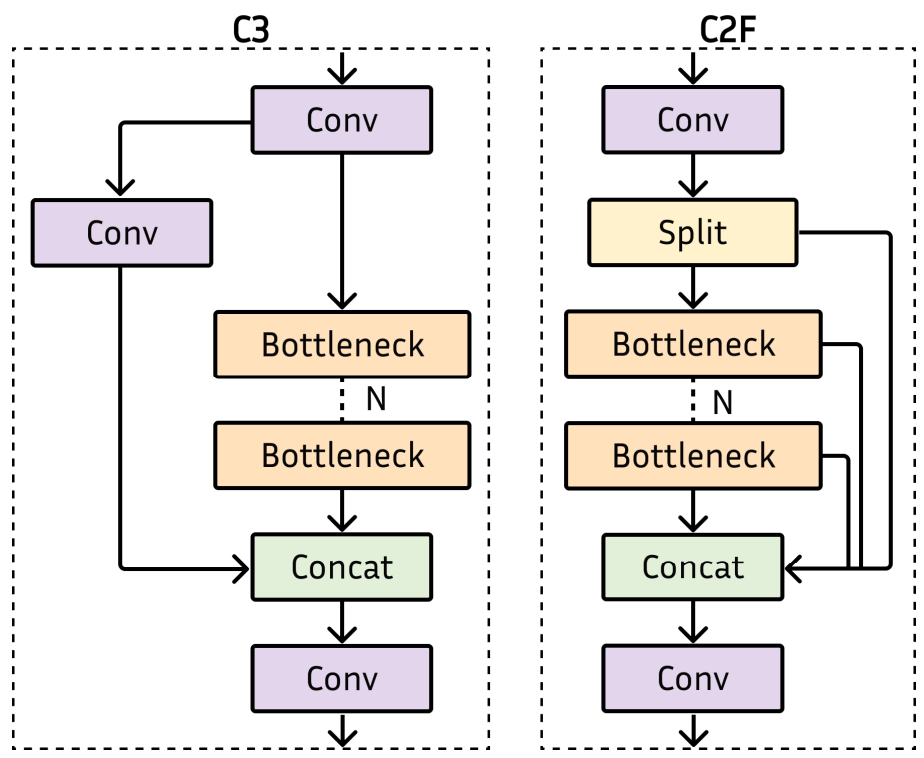

**FIGURE 18.** Comparison between the C3 module and the C2f module used in YOLOv8.

As a result of the NAS-based optimization process, YOLO-NAS integrates components specifically designed for quantization and efficient inference. Its core is built using Quantization-Aware RepVGG (QA-RepVGG) blocks, which support 8-bit quantization. These blocks are expanded during training to enhance learning capacity and collapsed at inference to reduce latency through reparameterization. Two specialized modules, Quantization Shortcut Path (QSP) and Quantization-Compatible Inference (QCI), are constructed from these blocks to facilitate post-training quantization while preserving accuracy. Additionally, the architecture incorporates attention mechanisms and reparameterization operations at inference time to improve object localization and reduce computational cost. All of these elements are outcomes of the NAS process and contribute to a design optimized for deployment.

**TABLE 10.** Details of YOLO-NAS variants. Speed is measured using a T4 GPU on the COCO dataset.

| Model | Size (pixels) | mAP (50) | Speed (ms) | Params (M) |
|---|---|---|---|---|
| YOLO-NASs | 640 | 47.5 | 3.21 | 19 |
| YOLO-NASm | 640 | 51.55 | 5.85 | 51.1 |
| YOLO-NASl | 640 | 52.22 | 7.87 | 66.9 |

To further enhance performance, YOLO-NAS employs a hybrid quantization strategy. Instead of quantizing all layers uniformly, this approach selectively applies quantization to specific parts of the network based on their sensitivity to precision loss. Layers critical for maintaining accuracy are kept in higher precision (FP32 and FP16), while less sensitive layers are quantized to INT8. This selective process improves the accuracy-latency trade-off and enables minimal degradation in detection performance after quantization.

To improve learning and calibration, YOLO-NAS incorporates Knowledge Distillation and DFL during training. Knowledge Distillation transfers semantic and structural information from a larger teacher model to the student, enhancing generalization. DFL replaces standard classification losses by modeling label distributions more precisely, which improves localization and stabilizes convergence. The NAS process yields three model variants, all sharing a modular structure composed of a backbone, neck, and head, as shown in Fig. 19, in line with previous YOLO architectures. Their specifications are summarized in Table10.

### *M. YOLO-WORLD*

YOLO-World [84] is introduced in 2024 as an extension of the YOLO family for open-vocabulary object detection. Unlike traditional YOLO models trained on a fixed set of categories, YOLO-World detects arbitrary object classes based on textual input at inference. The model builds upon the YOLOv8 architecture, maintaining its modular structure of backbone, neck, and detection head, while integrating components for combining visual and linguistic representations, as shown in Fig. 20. Its objective is to support zero-shot and few-shot detection by using semantic embeddings, extending the applicability of YOLO to open-set scenarios.

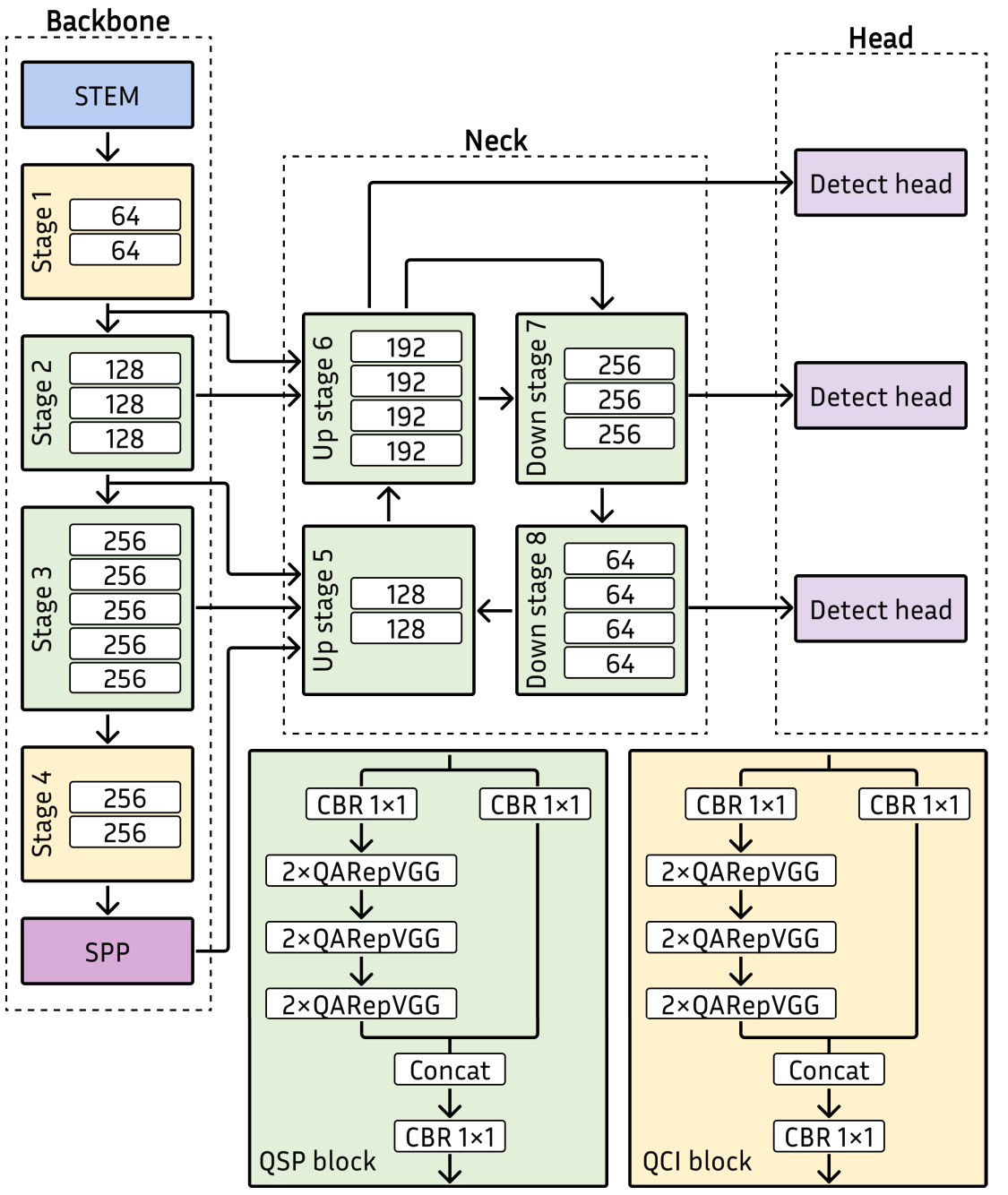

**FIGURE 19.** **YOLO-NAS architecture. CBR stands for Convolution + Banch Normalization + Rectified Linear Unit (ReLU) activation function.**

Although its focus diverges from the conventional YOLO paradigm by shifting from closed-set classification to open-vocabulary recognition, YOLO-World maintains architectural and methodological continuity with the core YOLO design. Given its growing relevance and its reliance on the YOLOv8 framework as foundation, it is included in this review as a development emerging from the YOLO ecosystem.

The architecture of YOLO-World comprises two main encoders, one for images and one for text, followed by a multimodal fusion module that enables open-vocabulary object detection. The image encoder adopts the YOLOv8 design and extracts multi-scale features from the input image using a convolutional backbone and a FPN. The text encoder is a Transformer-based module pre-trained under the CLIP framework. It maps input phrases into a dense embedding space that captures semantic relationships between object categories.

At the core of the cross-modal fusion mechanism is the Reparameterizable Vision-Language Path Aggregation Network (RepVL-PAN), which replaces the standard neck of YOLO. RepVL-PAN operates on the multi-scale visual features extracted by the backbone and integrates textual embeddings to construct semantically enriched feature maps. This process involves two internal components. First, the Text-guided Cross Stage Partial Layer (T-CSPLayer) modulates visual features through attention signals derived from the text embeddings, allowing the detector to condition intermediate representations on linguistic input. Second, the Image-Pooling Attention mechanism updates the text embeddings using compact visual summaries, enabling bidirectional information exchange between the modalities.

A distinctive aspect of RepVL-PAN is that it supports a reparameterization strategy for efficient deployment. During inference, the text encoder can be removed entirely by precomputing the embeddings associated with a given vocabulary and folding them into the model weights. This design enables YOLO-World to operate in real-time without sacrificing its capacity for open-vocabulary recognition.

The final stage of YOLO-World's architecture replaces the conventional classification layer with a Text Contrastive Head that aligns detected object regions with precomputed text embeddings. For each region proposal, the model generates a visual embedding and compares it with textual embeddings using a scaled dot product between L2-normalized vectors, followed by an affine transformation. The resulting similarity score determines the alignment between visual and linguistic features, enabling classification based on semantic proximity rather than fixed labels. During training, an online vocabulary is constructed for each mini-batch by combining the relevant ground-truth categories with randomly sampled distractors. A contrastive loss encourages each object to align with its corresponding text label while remaining distant from unrelated ones. This formulation enables YOLO-World to support zero-shot detection by generalizing beyond closed category vocabularies.

Finally, YOLO-World adopts an inference scheme based on offline vocabulary encoding. Instead of computing text embeddings during runtime, the model supports precomputing vocabulary embeddings from user-defined prompts and integrating them into the model weights through reparameterization. This eliminates the need for a text encoder at inference time, reducing computational overhead while preserving open-vocabulary capabilities. The encoded vocabulary can be adjusted to meet task-specific requirements, allowing flexible and efficient deployment. YOLO-World is available in four variants: small, medium, large, and extra large.

### *N. YOLOv9*

YOLOv9 [85] was proposed to address limitations related to information degradation in deep neural networks. It builds upon the YOLOv7 framework, using its architecture as a foundation for further refinement. Instead of introducing a radically new structure, YOLOv9 focuses on improving information preservation and utilization, particularly in compressed regions of lightweight models. The design emphasizes stable training dynamics and enhanced semantic representation, with the goal of improving detection accuracy across model scales while maintaining inference efficiency.

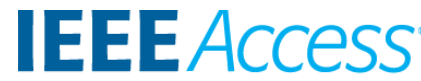

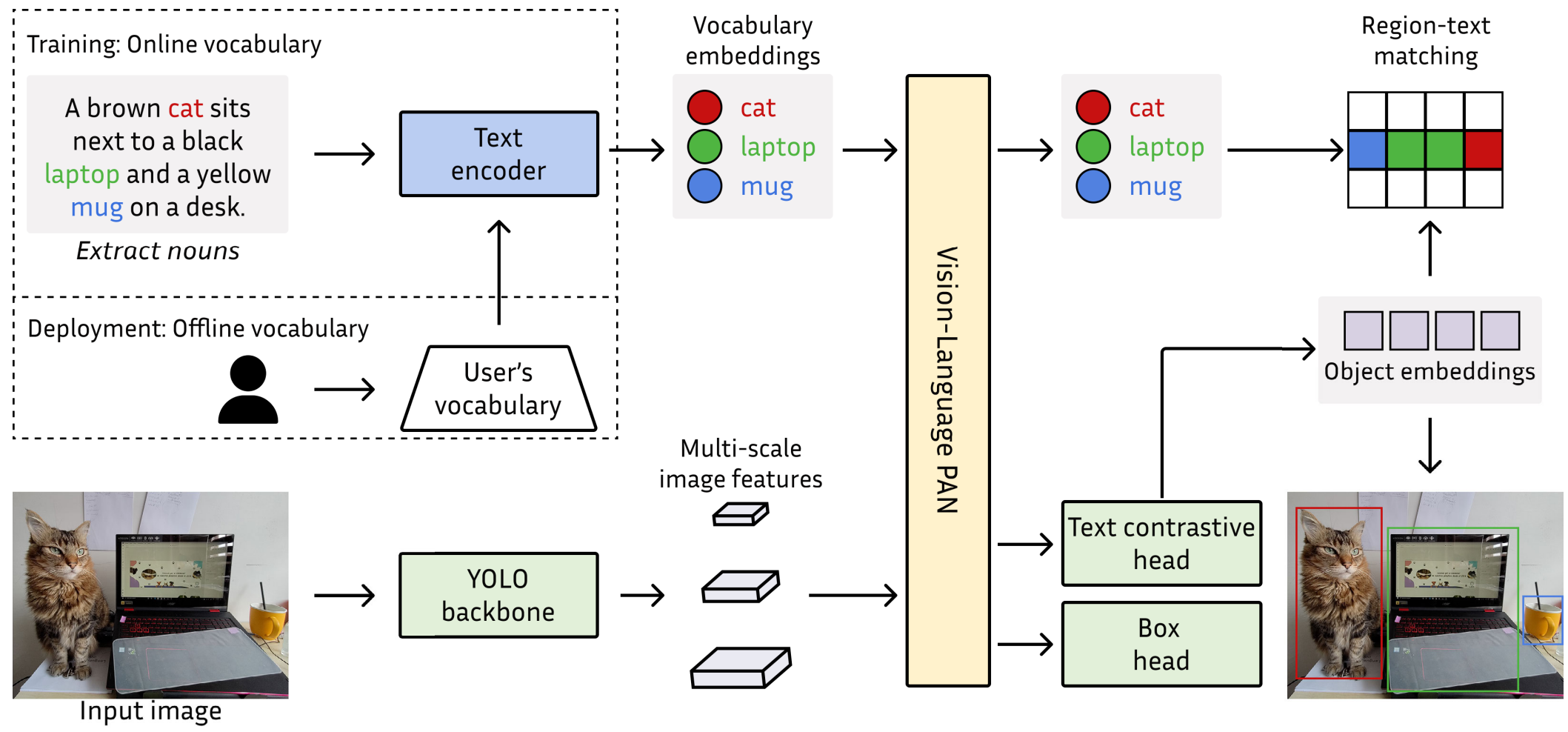

**FIGURE 20.** Overview of the YOLO-World architecture.

**TABLE 11.** Details of YOLOv9 variants. Speed is measured using a 2080 Ti GPU on the COCO dataset.

| Model | Size (pixels) | mAP (50) | Params (M) | FLOPs (G) |
|---|---|---|---|---|
| YOLOv9t | 640 | 53.1 | 2.0 | 7.7 |
| YOLOv9s | 640 | 63.4 | 7.1 | 26.4 |
| YOLOv9m | 640 | 68.1 | 20.0 | 76.3 |
| YOLOv9c | 640 | 70.2 | 25.3 | 102.1 |
| YOLOv9e | 640 | 72.8 | 57.3 | 189.0 |

A central idea in YOLOv9 is the incorporation of reversible functions to mitigate information loss during training. These functions are defined by their capacity to retain the full content of the input through transformations that can be exactly inverted. By integrating them into the network design, YOLOv9 preserves semantic and structural information as it propagates through the layers. This retention supports more accurate gradient calculations during backpropagation, leading to more stable and effective learning dynamics, especially in lightweight models where under-parameterization is a common limitation.

Additionally, YOLOv9 introduces the Programmable Gradient Information (PGI) module to improve training dynamics by enhancing gradient reliability without affecting inference cost. PGI consists of three components: a main branch used only during inference, an auxiliary reversible branch, and a multi-level auxiliary information pathway, as shown in Fig. 21. The auxiliary reversible branch provides reliable gradients to the main path during training by preserving essential features without altering the forward inference path. The auxiliary information pathway aggregates gradients from multiple semantic levels, ensuring all prediction branches contribute to shared representation learning. This structure makes PGI applicable across both deep and lightweight models, improving convergence and detection accuracy.

To support the goals of YOLOv9, the Generalized Efficient Layer Aggregation Network (GELAN) is introduced. GELAN extends the design principles of ELAN and CSPNet, combining efficient feature aggregation with optimized gradient propagation. Unlike architectures based on depthwise convolutions, GELAN relies on standard convolutions to improve parameter utilization while preserving a lightweight structure. Its modular design supports flexible stacking of computational blocks, allowing adaptation to different deployment scenarios and resource constraints. GELAN is fully compatible with the PGI module, enhancing gradient flow and training stability without sacrificing inference speed or accuracy. The structure of GELAN is shown in Fig. 22.

With the integration of PGI and GELAN, YOLOv9 refines and extends the architecture of YOLOv7, addressing limitations in information retention and gradient reliability during training. These enhancements improve optimization stability and detection performance across different model scales, while preserving the real-time efficiency that defines the YOLO family. YOLOv9 is released in five model variants, whose specifications are listed in Table 11.

### *O. YOLOv10*

YOLOv10 [86], released in 2024, builds upon YOLOv8 as its architectural baseline, preserving the modular structure of backbone, neck, and head. However, it shifts the design focus from structural refinement to addressing two long-standing limitations in real-time object detection: enabling fully end-to-end training and eliminating the reliance on post-processing with NMS. While previous YOLO models depend on NMS to resolve redundant predictions, this step introduces additional overhead and breaks the end-to-end nature of the pipeline. YOLOv10 tackles these challenges by redesigning both the training strategy and detection head, aiming to reduce computational redundancy and latency without compromising detection accuracy.

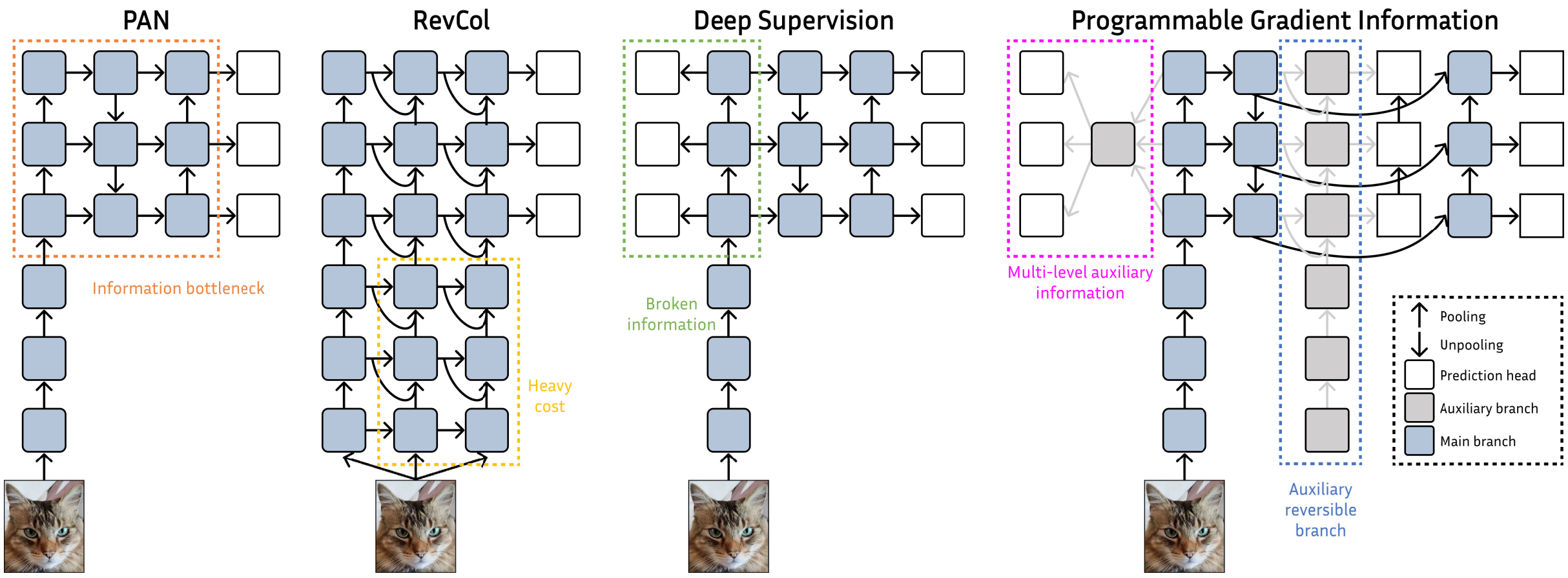

**FIGURE 21.** Comparison between the PGI mechanism in YOLOv9 and alternative gradient guidance approaches.

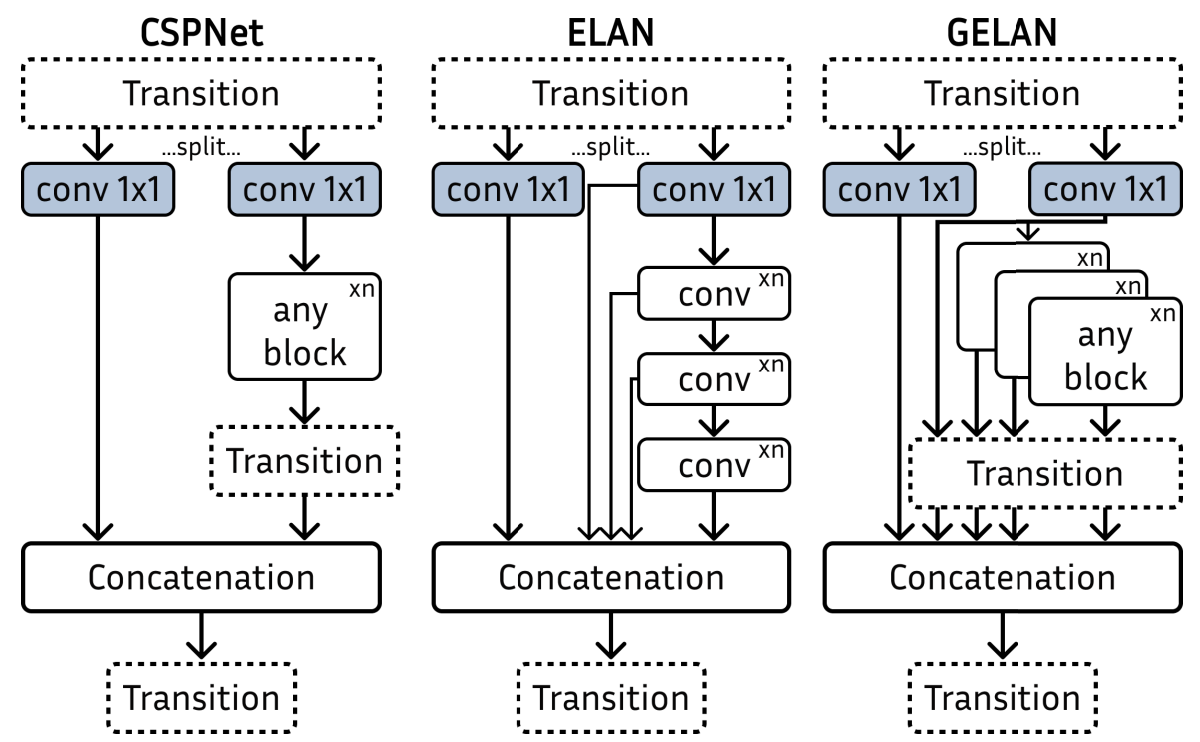

**FIGURE 22.** Comparison between the GELAN block utilized in YOLOv9 and other related methods.

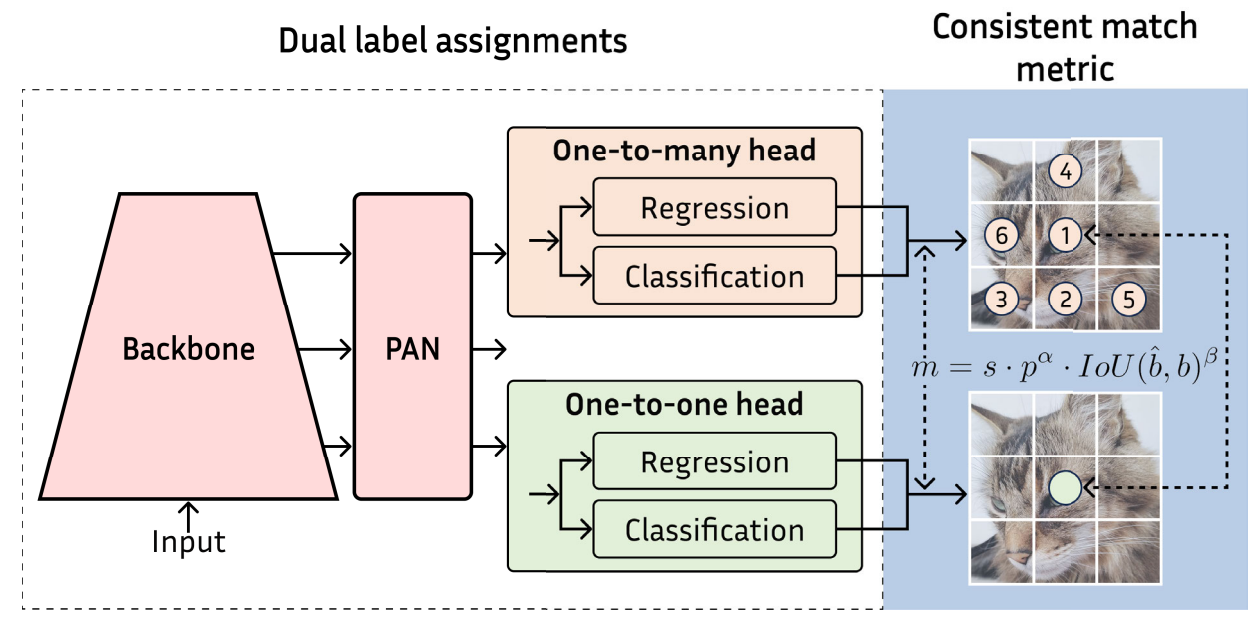

**FIGURE 23.** Illustration of the consistent dual assignment process used in YOLOv10 for NMS-free training.

First, YOLOv10 introduces the Consistent Dual Assignments (CDA) strategy, which integrates a one-to-one head alongside the traditional one-to-many head used in earlier YOLO models. Both heads share the same architecture and optimization objectives but differ in their label assignment mechanisms. During training, they are jointly optimized, allowing the backbone and neck to benefit from the dense supervision provided by the one-to-many assignment, which enhances feature learning. At inference, only the one-to-one head is used, enabling the model to produce a single, high-quality prediction per object without relying on NMS. A consistent matching metric synchronizes the optimization of both heads by ensuring that the most informative positives from the one-to-many head also serve as optimal matches for the one-to-one head. This alignment minimizes the gap between training and inference, improving prediction accuracy and reducing latency. The complete mechanism is illustrated in Fig. 23.

YOLOv10 also introduces architectural modifications to improve computational efficiency. It replaces the conventional stacked $3 \times 3$ convolutional layers in the classification head with two depth-wise separable convolutions followed by a $1 \times 1$ point-wise convolution, reducing complexity while preserving accuracy. Additionally, it implements Spatial-Channel Decoupled Downsampling. Unlike earlier models that perform spatial downsampling and channel transformation jointly, YOLOv10 decouples these steps by first applying a $1 \times 1$ convolution for channel adjustment, followed by a depth-wise convolution for spatial reduction. This separation reduces redundancy, lowers computational cost, and allows more precise control over information flow in the early stages of the network.

To further reduce redundancy and improve architectural efficiency, YOLOv10 adopts a rank-guided block design strategy. This method analyzes the intrinsic rank of each stage, defined by the final convolutional layer in its last basic block, to identify regions with excessive parameter redundancy. Based on this analysis, lower-rank stages are progressively replaced with a more efficient structure called the Compact Inverted Block (CIB). As illustrated in Fig. 24, the CIB module combines depth-wise convolutions for spatial feature mixing with point-wise convolutions for channel interaction, providing a lightweight alternative to conventional convolutional blocks. The CIB is integrated into the layer aggregation structures used in ELAN and GELAN, enabling YOLOv10 to improve efficiency without sacrificing detection performance across model scales.

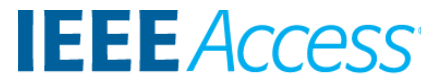

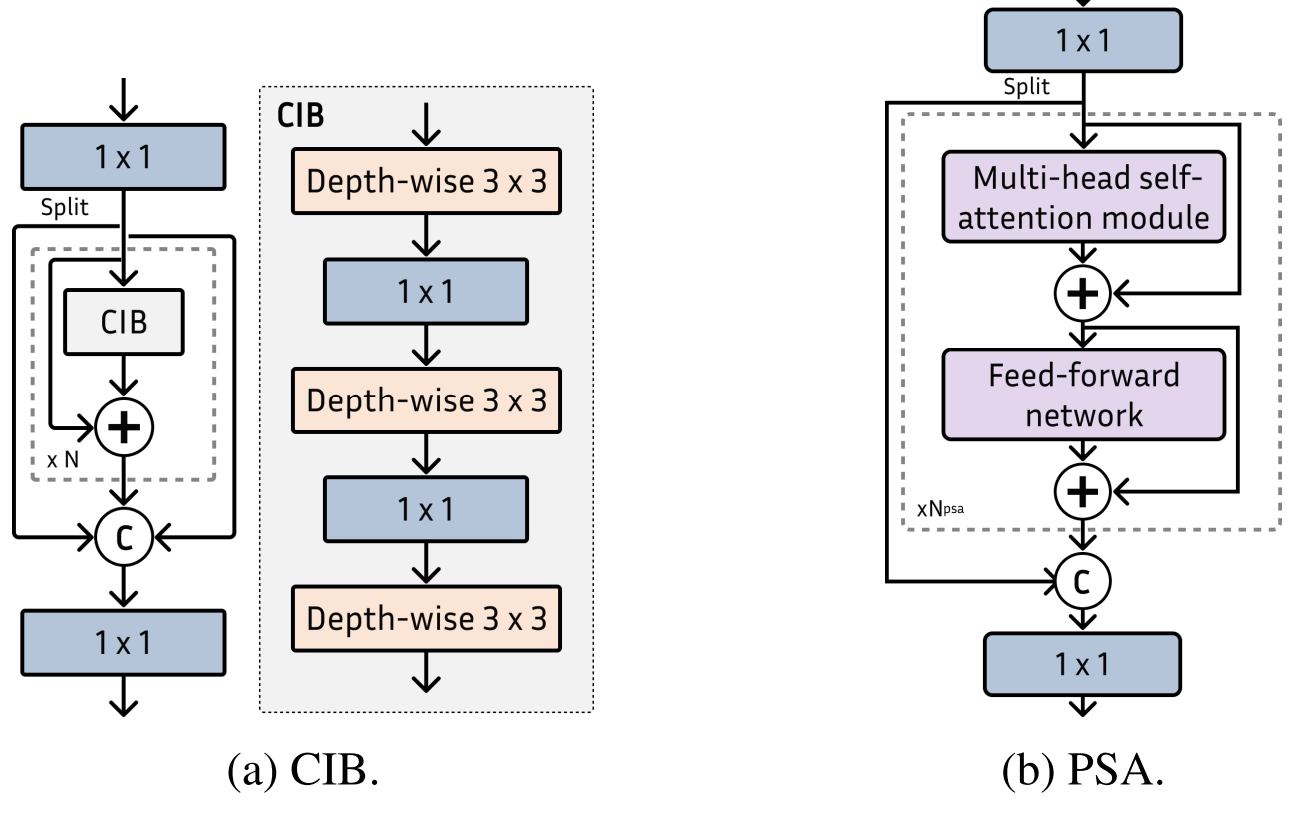

(a) CIB. (b) PSA.

**FIGURE 24.** CIB and PSA modules utilized in YOLOv10.

**TABLE 12.** Details of the YOLOv10 variants. Speed is measured using a T4 GPU on the COCO dataset.

| Model | Size (pixels) | mAP (50) | Speed (ms) | Params (M) | FLOPs (G) |
|---|---|---|---|---|---|
| YOLOv10-N | 640 | 38.5 | 1.84 | 2.3 | 6.7 |
| YOLOv10-S | 640 | 46.3 | 2.49 | 7.2 | 21.6 |
| YOLOv10-M | 640 | 51.1 | 4.74 | 15.4 | 59.1 |
| YOLOv10-B | 640 | 52.5 | 5.74 | 19.1 | 92.0 |
| YOLOv10-L | 640 | 53.2 | 7.28 | 24.4 | 120.3 |
| YOLOv10-X | 640 | 54.4 | 10.70 | 29.5 | 160.4 |

Finally, YOLOv10 incorporates two architectural components in its deeper stages. First, it introduces large-kernel depthwise convolutions by expanding the kernel size of the second 3 × 3 depthwise layer in the CIB module to 7 × 7. This increases the receptive field and enhances contextual understanding, particularly for larger objects. To mitigate optimization instability and inference overhead, this design is restricted to smaller model variants and employs structural reparameterization. Second, YOLOv10 integrates a Partial Self-Attention (PSA) module in the later stages of the network, as shown in Fig. 24b. PSA begins with a 1 × 1 convolution to split features across channels, after which self-attention is applied only to a subset using a sequence of Multi-Head Self-Attention (MHSA) and feed-forward layers. The attended and unattended branches are then fused via another 1 × 1 convolution. By limiting PSA to lower-resolution stages, the model captures global dependencies while minimizing the quadratic cost of full self-attention.

All the modifications described above make YOLOv10 a potentially suitable architecture for end-to-end deployment and applications where inference speed is critical, without compromising robust performance. YOLOv10 is available in six model variants, one more than YOLOv8, with specifications provided in Table 12.

### P. YOLOv11

YOLOv11 [87], released in 2024 by Ultralytics, builds upon the advances introduced in YOLOv8, YOLOv9, and YOLOv10. It maintains the modular architecture composed of a backbone, neck, and detection head, as illustrated in Fig. 25. This version focuses on lightweight deployment, aiming to reduce parameter count while delivering performance comparable to contemporary state-of-the-art object detection models.

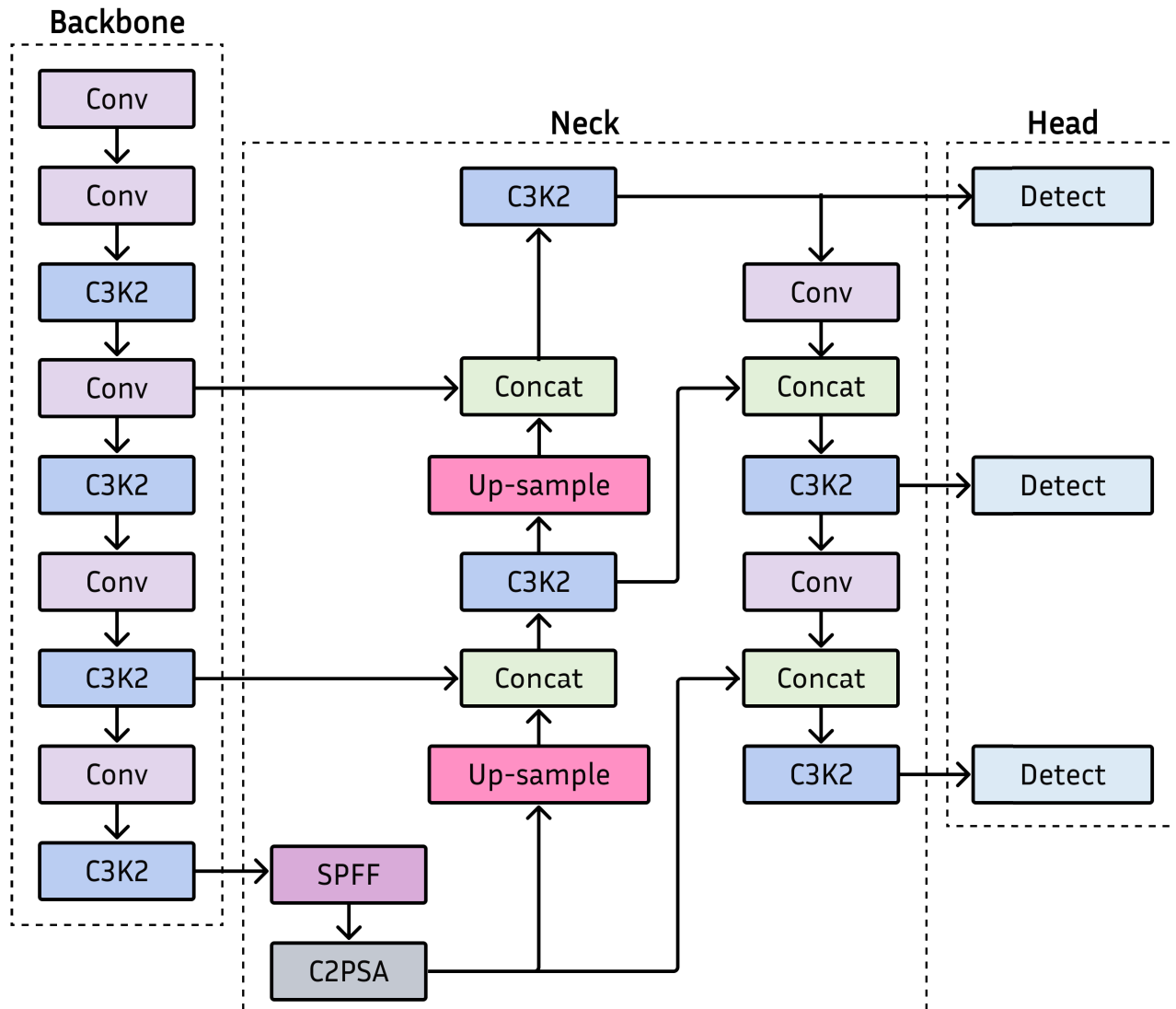

**FIGURE 25.** High-level diagram of the YOLOv11 architecture.

The backbone of YOLOv11 remains consistent with previous versions, using convolutional blocks composed of convolutional layers, batch normalization, and the SiLU activation function. These blocks downsample the input and extract hierarchical features. A key change in YOLOv11 is the replacement of the C2f structure with the C3k2 block, an evolution of the CSP bottleneck design aimed at improving feature extraction and information flow. C3k2 blocks split the feature maps and apply multiple 3 × 3 convolutions, offering a good trade-off between accuracy and speed. Each block contains a submodule called C3K, which resembles C2f but omits the splitting operation. Instead, it processes the input through an initial convolutional layer, a sequence of $n$ bottleneck layers with concatenations, and a final convolutional layer. This structure, shown in Fig. 26, is designed to enhance efficiency during feature extraction.

The neck of YOLOv11 preserves the core structure of YOLOv8, including the SPPF module for multi-scale context aggregation. It introduces the C3k2 blocks, already used in the backbone, and a new component called Convolutional Block with Partial Spatial Attention (C2PSA). The C2PSA block integrates attention mechanisms to emphasize spatially relevant regions, improving detection of small or partially occluded objects. Structurally, it includes two Partial Spatial Attention (PSA) modules applied to separate branches of the input feature map. These branches are concatenated following the same pattern as in the C2f block. Each PSA module applies attention to the input, concatenates the original and weighted features, and passes the result through feed-forward and convolutional layers. A final concatenation completes the block. The structure of C2PSA and PSA is shown in Fig. 27. This design improves spatial focus while maintaining computational efficiency.

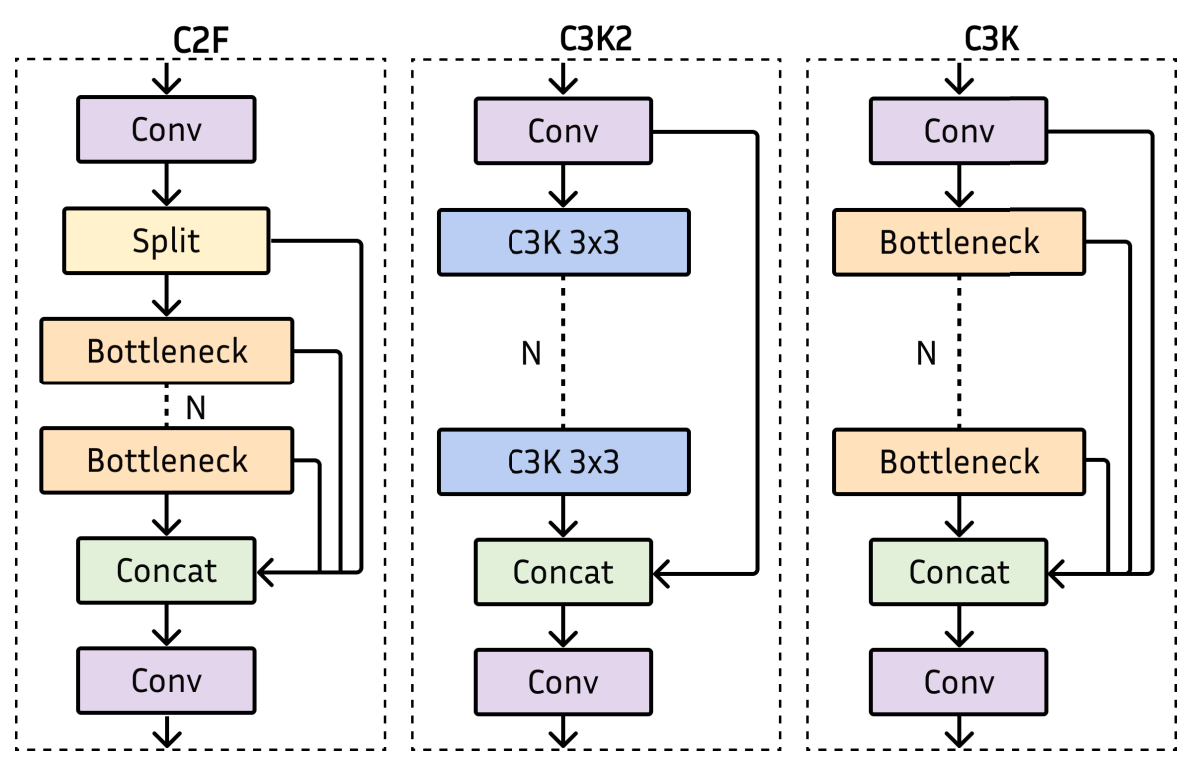

**FIGURE 26.** Structure of the C3k2 block utilized in YOLOv11.

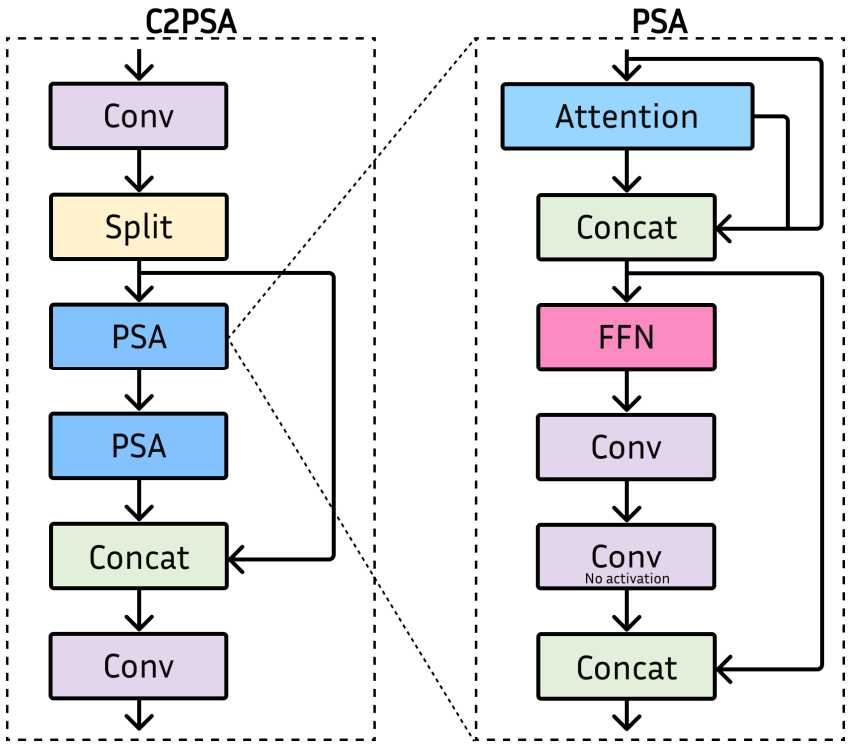

**FIGURE 27.** Structure of C2PSA and PSA blocks utilized in YOLOv11.

Finally, the detection head in YOLOv11 follows the same multi-scale prediction strategy used in previous versions of the YOLO series. It produces bounding box predictions at three spatial resolutions, corresponding to low, medium, and high-level feature maps, allowing the model to detect objects of different sizes. These predictions are based on the feature representations extracted by the backbone and refined by the neck. YOLOv11 is released in five variants, whose details are summarized in Table 13.

**TABLE 13.** Details of YOLOv11 variants. Speed is measured using a T4 GPU on the COCO dataset.

| Model | Size (pixels) | mAP (50-95) | Speed (ms) | Params (M) | FLOPs (B) |
|---|---|---|---|---|---|
| YOLOv11n | 640 | 56.1 | 1.5 | 2.6 | 6.5 |
| YOLOv11s | 640 | 90.0 | 2.5 | 9.4 | 21.5 |
| YOLOv11m | 640 | 183.2 | 4.7 | 20.1 | 68.0 |
| YOLOv11l | 640 | 238.6 | 6.2 | 25.3 | 86.9 |
| YOLOv11x | 640 | 462.8 | 11.3 | 56.9 | 194.9 |

### *Q. YOLOv12*

YOLOv12 [88], introduced in 2025, is the one of the most recent release in the YOLO series at the time of writing. It builds on previous versions such as YOLOv11 and maintains the modular structure composed of a backbone, neck, and detection head. In contrast to earlier iterations that rely primarily on convolutional networks, YOLOv12 adopts an attention-focused design to leverage the representational capacity of attention mechanisms. These modules have typically been avoided in YOLO architectures due to concerns related to computational cost and memory access patterns. The main goal of YOLOv12 is to incorporate attention modules into the detection pipeline without compromising real-time inference speed.

The backbone of YOLOv12 maintains a hierarchical architecture consistent with previous versions, allowing progressive multi-scale feature extraction. The initial stages are inherited from YOLOv11 and preserve a convolutional structure optimized for efficiency. A central modification is the integration of the Residual Efficient Layer Aggregation Network (R-ELAN), which replaces the ELAN blocks used in earlier YOLO models. R-ELAN introduces a residual shortcut connection from the block input to its output, modulated by a learnable scaling factor, as shown in Fig. 28. This design helps prevent gradient degradation and improves convergence, particularly in deeper configurations.

Unlike the original ELAN, which splits input features across multiple parallel paths, R-ELAN adjusts the channel dimensions through a transition layer before processing the features sequentially and merging them via concatenation. This redesign produces a bottleneck-style structure that improves stability and computational efficiency. To enhance spatial modeling with minimal parameter overhead, YOLOv12 employs $7 \times 7$ depthwise separable convolutions, which expand the receptive field without the memory cost of standard large-kernel convolutions. This strategy maintains spatial sensitivity without requiring explicit positional encoding. Additionally, YOLOv12 removes the triple-block stacking used in the final stage of earlier versions, replacing it with a single R-ELAN block to reduce redundancy and improve training efficiency while preserving representational capacity.

The neck of YOLOv12 retains the modular fusion strategy used in earlier versions, aggregating multi-scale features to support detection across object sizes. It introduces modifications to support attention-based processing, most notably the integration of Area Attention (A2), a lightweight mechanism designed for real-time applications. Instead of using complex partitioning schemes, A2 segments feature maps into uniformly spaced vertical or horizontal areas via simple reshape operations, as shown in Fig. 29. This approach maintains a wide receptive field while limiting computational cost, making it more suitable for fast inference than global or axial attention. To further improve efficiency, YOLOv12 includes FlashAttention, which restructures attention as optimized memory I/O operations, reducing bandwidth usage and latency. Combined with A2, it enables high-speed localized attention, enhancing the model's ability to identify informative regions in cluttered or high-resolution inputs.

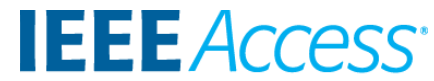

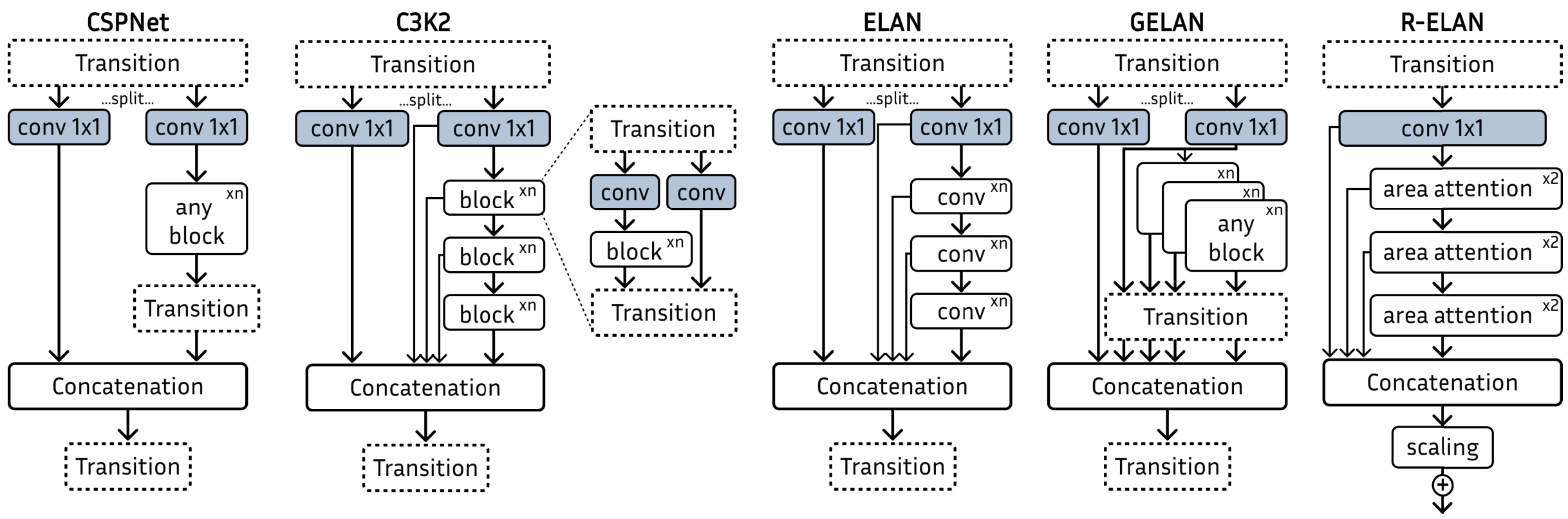

**FIGURE 28.** Comparison of the R-ELAN block used in YOLOv12 with prior modules, including ELAN, GELAN, C3K2, and CSPNet.

**TABLE 14.** Details of the YOLOv12 variants. Speed is measured using a T4 GPU on the COCO dataset.

| Model | Size (pixels) | mAP (50-95) | Speed (ms) | Params (M) | FLOPs (G) |
|---|---|---|---|---|---|
| YOLOv12n | 640 | 40.4 | 1.60 | 2.5 | 6.0 |
| YOLOv12s | 640 | 47.6 | 2.42 | 9.1 | 19.4 |
| YOLOv12m | 640 | 52.5 | 4.27 | 19.6 | 59.8 |
| YOLOv12l | 640 | 53.8 | 5.83 | 26.5 | 82.4 |
| YOLOv12x | 640 | 55.4 | 10.38 | 59.3 | 184.6 |

Additionally, YOLOv12 adjusts the MLP expansion ratio used in its attention blocks. While traditional vision transformers adopt a 4:1 ratio between the feed-forward layer and input dimension, YOLOv12 reduces this ratio to 1.2:1 in smaller variants (n, s, m) and to 2:1 in larger ones (l, x). This rebalancing lowers parameter count and memory consumption while maintaining effective representational capacity, contributing to overall improvements in inference efficiency.

The prediction head in YOLOv12 retains the core functional design of previous YOLO versions, generating class scores, bounding box coordinates, and objectness values through convolutional layers. While the overall structure remains consistent, YOLOv12 introduces refinements to improve both accuracy and computational efficiency. The detection paths are optimized for multiple spatial predictions, enabling consistent performance across varied object sizes. Additionally, the integration of Area Attention into the head enhances the model's focus on spatially informative features, contributing to more precise and responsive predictions. YOLOv12 is released in five variants, with their specifications detailed in Table 14.

### *R. YOLOv13*

YOLOv13 [89] is the latest model in the family released so far. Despite being an early release, it is already positioning itself as the thirteenth official version of the series. YOLOv13 builds on the limitations observed in earlier versions, such as YOLOv11 and YOLOv12, which were restricted to modeling local and pairwise (pixel-to-pixel) correlations. These previous architectures, whether based on convolutions or local attention mechanisms, failed to capture global semantic correlations or higher-order relationships across multiple regions of an image, which limited performance in complex scenarios. The model retains the three-part structure of backbone, neck, and head, as shown in Fig. 30, but incorporates new additions.

First, YOLOv13 incorporates the Hypergraph-based Adaptive Correlation Enhancement (HyperACE) mechanism, shown in Fig. 31, one of its main innovations. HyperACE enables efficient modeling of higher-order correlations through the construction of adaptive hypergraphs, rather than relying on conventional binary correlations. In a hypergraph, each hyperedge connects multiple nodes (pixels or regions), allowing the capture of complex many-to-many interactions. Unlike previous works that manually constructed these hyperedges with fixed thresholds, HyperACE dynamically learns the participation degree of each node, making the modeling more robust and accurate.

Another distinctive component of YOLOv13 is the Full-Pipeline Aggregation-and-Distribution (FullPAD) paradigm. This system distributes the features enriched by HyperACE throughout the entire architecture, from the backbone to the detection head. It employs specialized tunnels to ensure a refined and uninterrupted flow of information, enhancing the synergy among intermediate representations. This strategy also facilitates better gradient propagation during training, leading to improvements in learning.

In terms of efficiency, YOLOv13 enhances its architecture by replacing conventional heavy convolutions with lightweight blocks built on depthwise separable convolutions. Two new modules, DS-C3k and DS-C3k2, embody this design philosophy as shown in Fig. 32. The DS-C3k block inherits the CSP-C3 structure while integrating depthwise-separable bottlenecks. The input features are first compressed with a $1 \times 1$ convolution, then processed through cascaded DS-Bottleneck units, combined with a parallel $1 \times 1$ convolutional shortcut, and finally fused back with another $1 \times 1$ layer. This design preserves the cross-channel branching of CSP while significantly reducing computational cost.

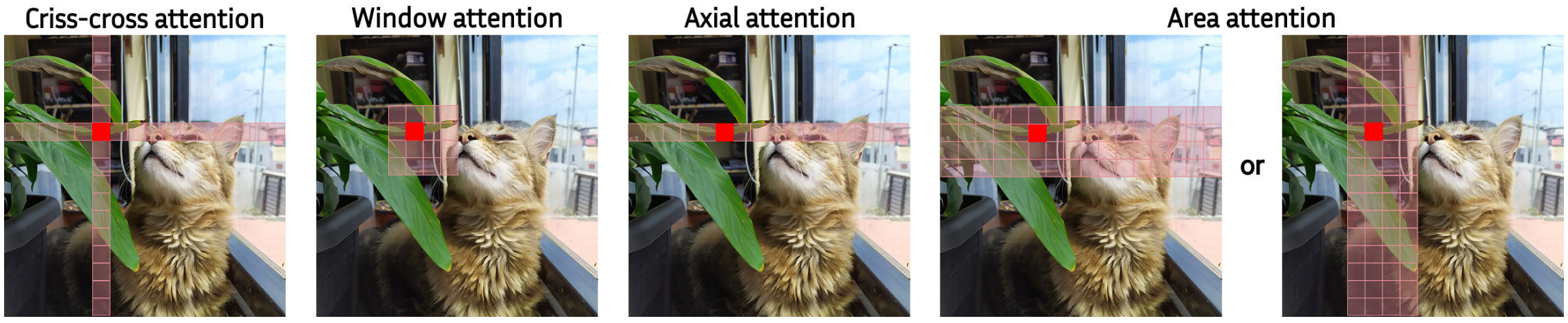

**FIGURE 29.** Illustration of the Area Attention approach in YOLOv12 in contrast with alternative attention strategies.

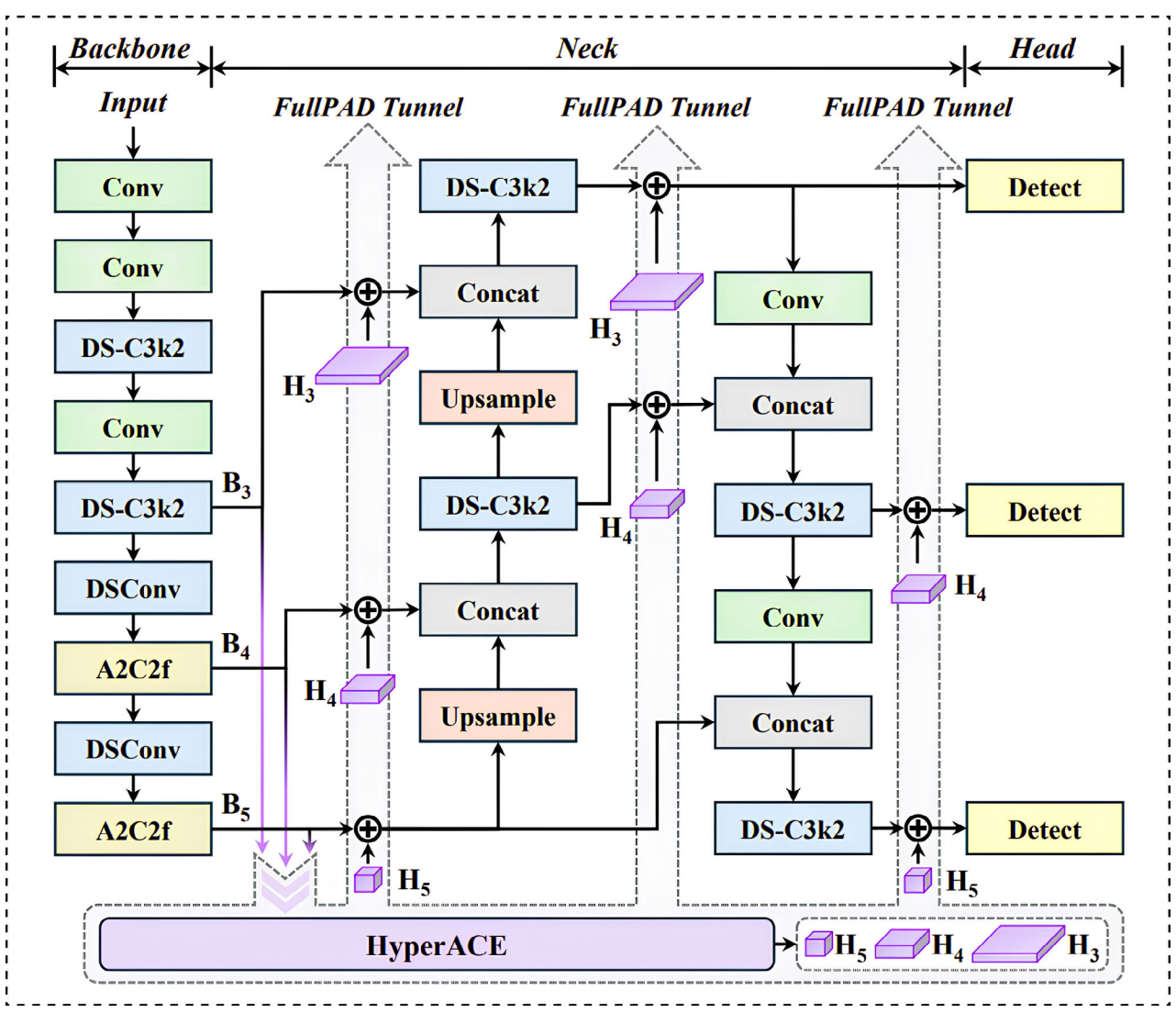

**FIGURE 30.** YOLOv13 architecture.

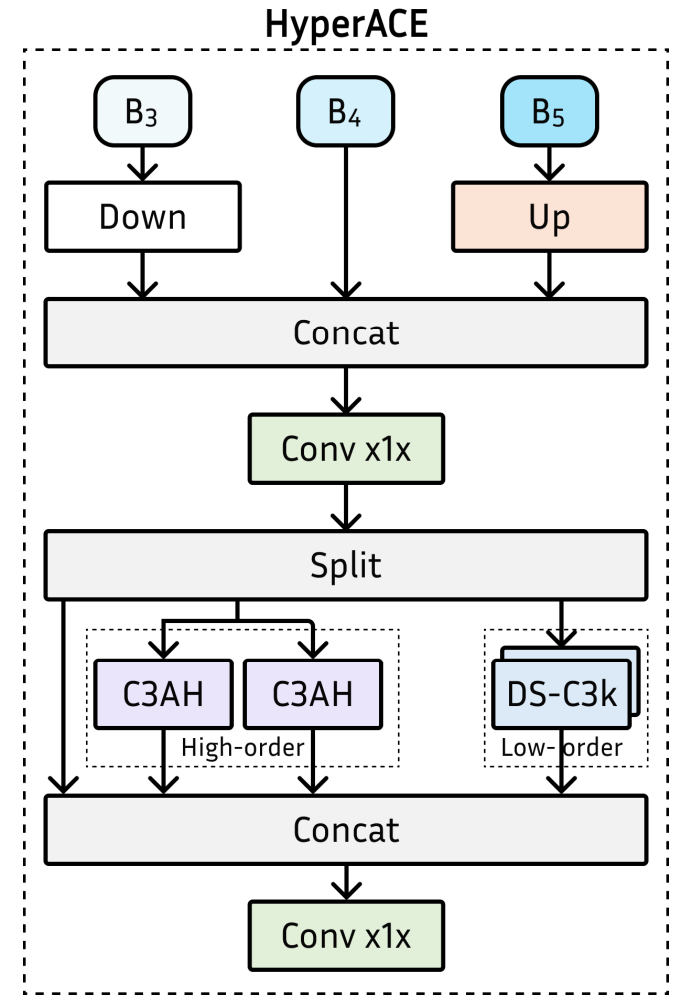

**FIGURE 31.** HyperACE mechanism introduced in YOLO13.

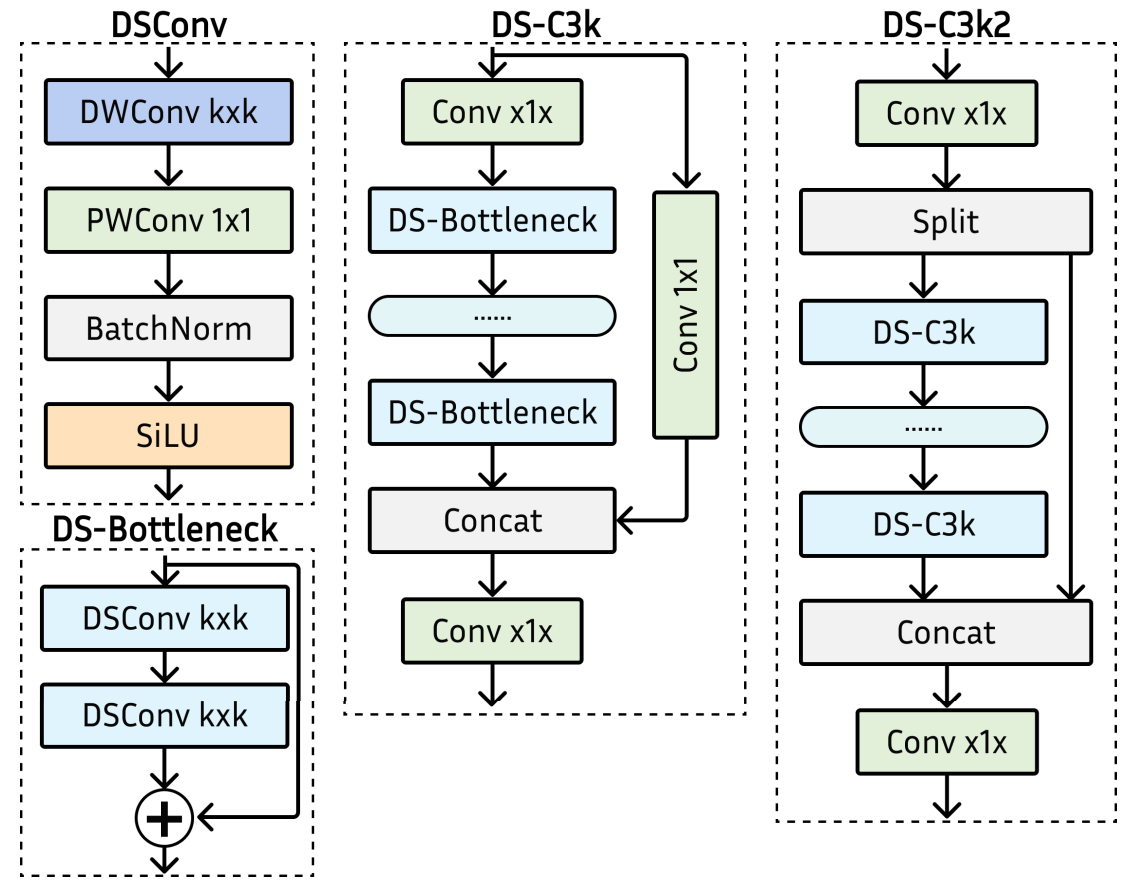

**FIGURE 32.** DS-C3k and DS-C3k2 modules introduced in YOLO13.

The DS-C3k2 block builds on the C3k2 structure by unifying channels with a $1 \times 1$ convolution, splitting the features into two paths, and passing one through multiple DS-C3k modules while the other follows a shortcut connection. The outputs are then concatenated and refined with a $1 \times 1$ convolution. DS-C3k2 serves as the fundamental building block across both the backbone and neck of YOLOv13, leading to substantial reductions in parameters without compromising accuracy. These refinements make YOLOv13 particularly suitable for real-time detection and deployment on hardware with limited computational resources.

Finally, the design of the C3AH, shown in Fig. 33, block is another key contribution. This block combines the traditional CSP structure with an adaptive hypergraph module, enabling high-order global semantic aggregation across diverse spatial positions. The block is integrated within the HyperACE module to further enrich visual representations at multiple scales. At present, YOLOv13 is available in four versions, whose details are provided in Table 15.

## VI. APPLICATION LANDSCAPE OF YOLO

YOLO's versatility, efficiency, and real-time performance have enabled its integration into a wide array of application domains. Over the years, it has been adopted in both academic and industrial settings to address diverse visual recognition tasks. As illustrated in Fig. 34, the range of YOLO-based applications spans multiple levels and contexts, from general-purpose monitoring to highly specialized and safety-critical fields. This variety reflects the model's adaptability to different data modalities, operational requirements, and deployment environments. What follows is a selection of some of the most prominent areas where YOLO has been applied.

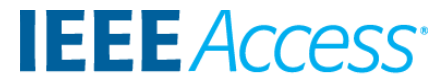

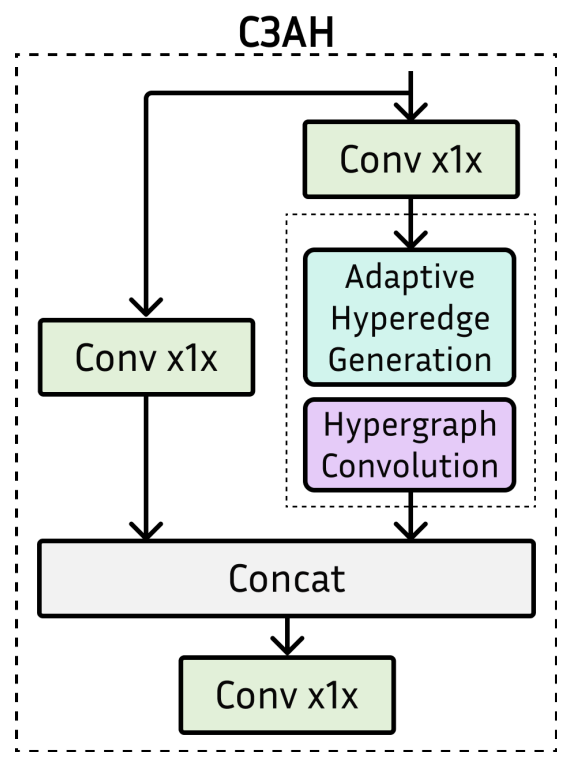

**FIGURE 33.** C3AH block introduced in YOLO13.

**TABLE 15.** Details of the YOLOv13 variants. Speed is measured using a T4 GPU on the COCO dataset.

| Model | Size (pixels) | mAP (50-95) | Speed (ms) | Params (M) | FLOPs (G) |
|---|---|---|---|---|---|
| YOLOv13n | 640 | 41.6 | 1.97 | 2.5 | 6.4 |
| YOLOv13s | 640 | 48.0 | 2.98 | 9.0 | 20.8 |
| YOLOv13l | 640 | 53.4 | 8.63 | 27.6 | 88.4 |
| YOLOv13x | 640 | 54.8 | 14.67 | 64.0 | 199.2 |

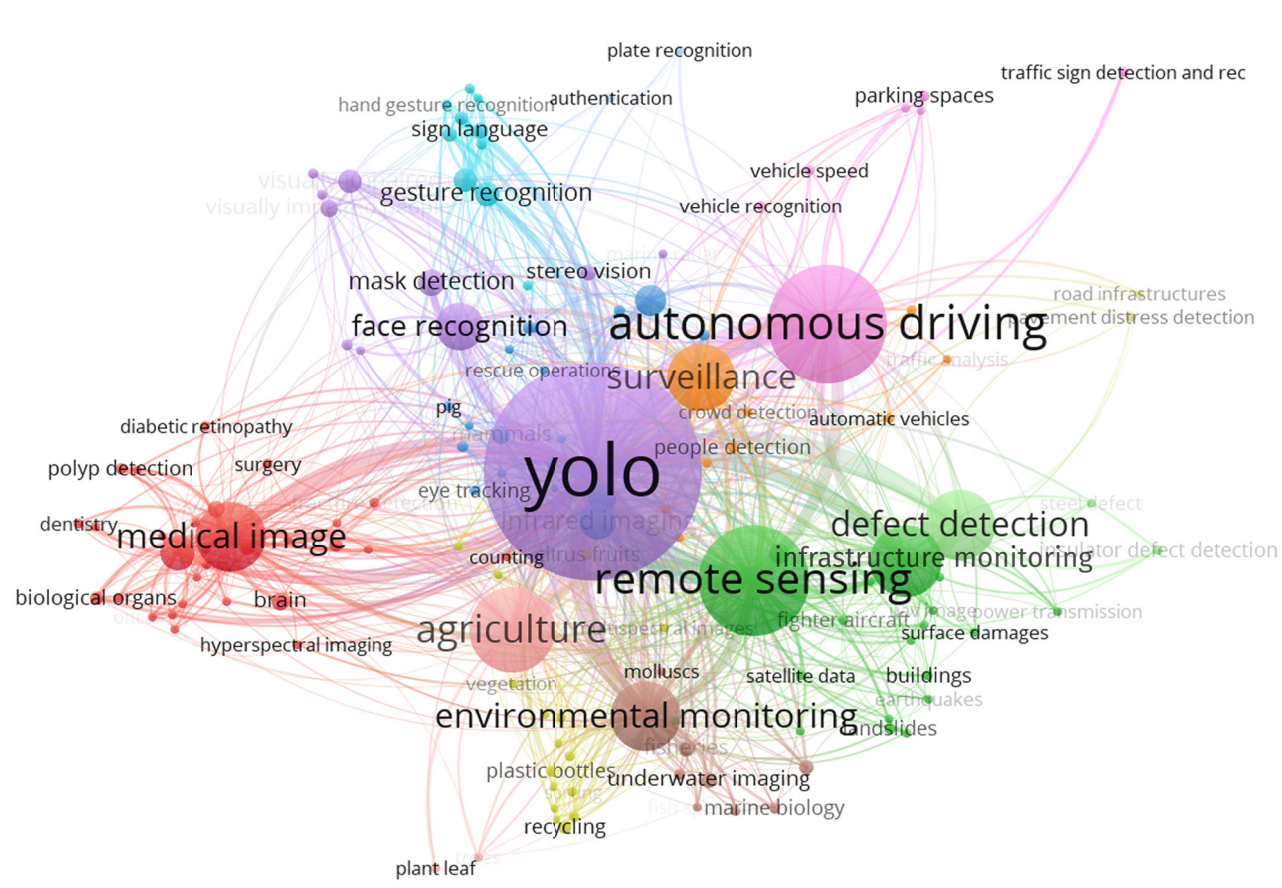

**FIGURE 34.** Bibliometric network visualization of YOLO-related application domains. Created with VOSviewer software using Scopus data (20,707 articles) retrieved with the query: 'YOLO' OR 'You Only Look Once.'

## *A. AUTONOMOUS DRIVING*

Autonomous driving represents one of the most active and demanding application domains for real-time object detection. This field requires continuous perception and decision-making under dynamic and potentially hazardous conditions. In this context, YOLO-based models are widely adopted due to their ability to deliver fast and accurate predictions, making them suitable for perception systems that must operate reliably in complex road environments. Some visual examples of detection in this domain can be seen in Fig. 35.

One of the most critical tasks in this domain is pedestrian detection. YOLO has been applied in multiple studies [90], [91], [92], [93], [94], [95], [96], [97] to identify pedestrians for safety and collision avoidance purposes. Models must recognize the silhouette and posture of individuals under varying lighting conditions, occlusions, and crowd densities, which makes this task particularly challenging due to its direct implications for real-time decision-making. Vehicle detection is another central application where YOLO has been adopted [98], [99], [100], [101], [102], [103], [104], [105]. The goal is to reliably identify different vehicle types such as cars, buses, and motorcycles in dynamic road environments, enabling functionalities like obstacle

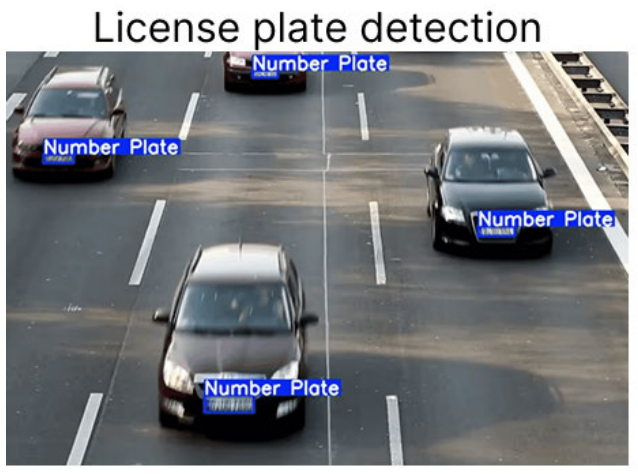

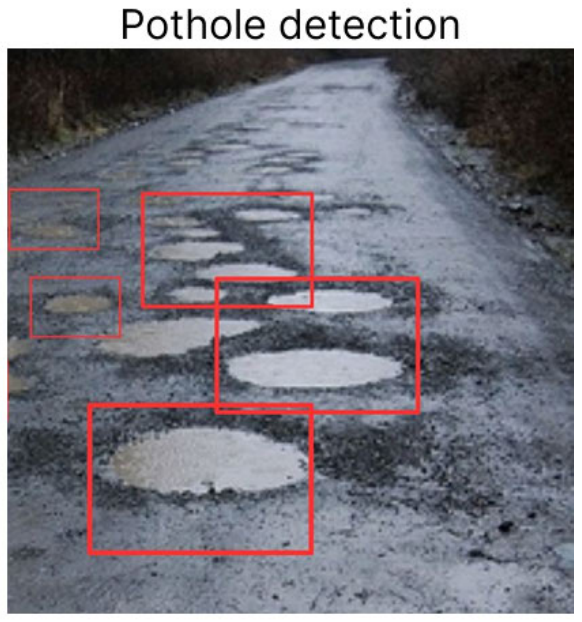

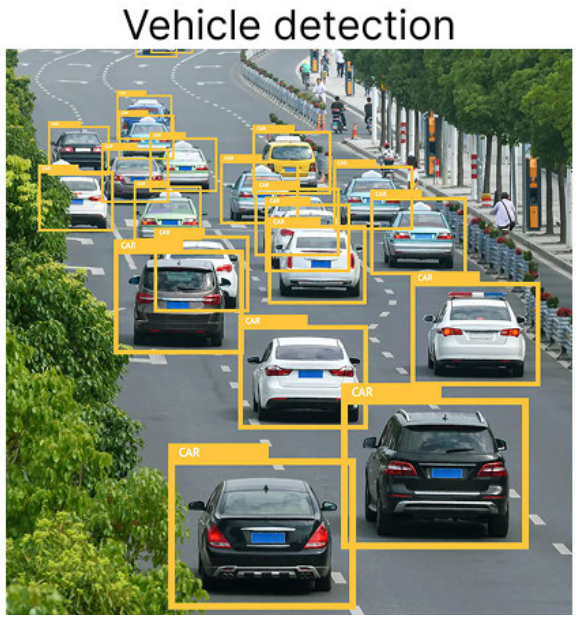

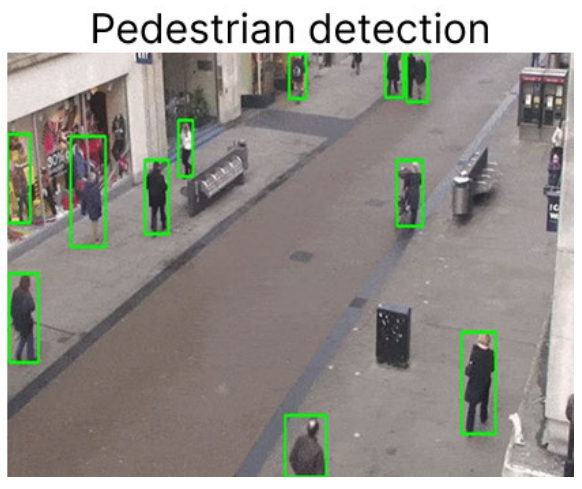

**FIGURE 35.** Examples of YOLO-based object detection tasks in the domain of autonomous driving, where it contributes to road analysis, traffic monitoring, and situational awareness for intelligent transportation systems.

detection, speed estimation, and path planning essential for autonomous navigation.

Beyond analyzing traffic dynamics and surrounding vehicles, autonomous systems must also interpret the condition of the driving surface. A notable application in this area is pothole detection, where YOLO has been used in several studies [106], [107], [108], [109], [110], [111] to identify road surface defects that may pose safety risks or cause vehicle damage. These systems typically rely on onboard cameras to process visual input in real time, enabling autonomous vehicles to adjust their trajectory or trigger maintenance alerts.

Building on vehicle-centric perception, license plate recognition has also become a frequent application of YOLO in traffic-related systems [112], [113], [114], [115], [116], [117], [118]. YOLO is typically used as a first stage to detect the license plate region, which is then passed to an Optical Character Recognition (OCR) module. This pipeline enables real-time identification of vehicles for purposes such as automated tolling, access control, and traffic law enforcement. Traffic sign detection is another closely related application in which YOLO has been employed [119], [120], [121], [122], [123], [124], [125]. The model is trained to detect and classify regulatory, warning, and informational signs under diverse conditions, including motion blur, adverse weather, and partial occlusion. Accurate recognition of these elements is critical for interpreting road regulations and improving situational awareness in dynamic driving environments.

Another traffic-related application involves the detection of parking slots and available parking spaces [126], [127], [128], [129]. YOLO has been integrated into systems that monitor parking availability in real time, typically using overhead cameras or vehicle-mounted sensors. These approaches aim to optimize parking management and reduce search time in congested urban areas.

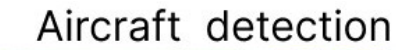

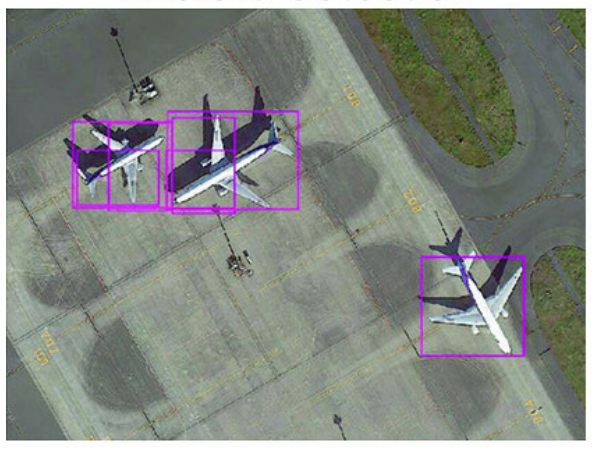

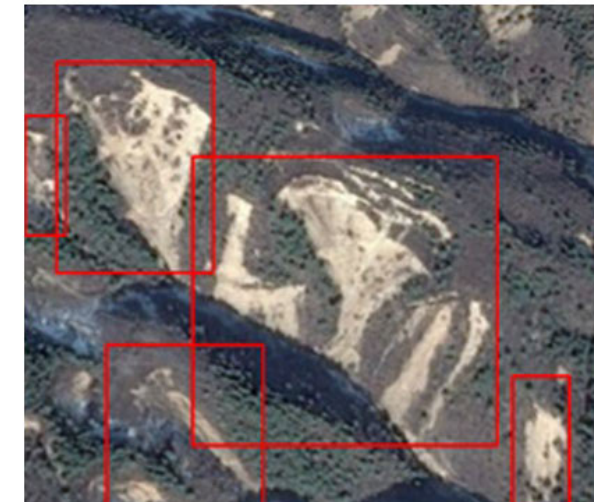

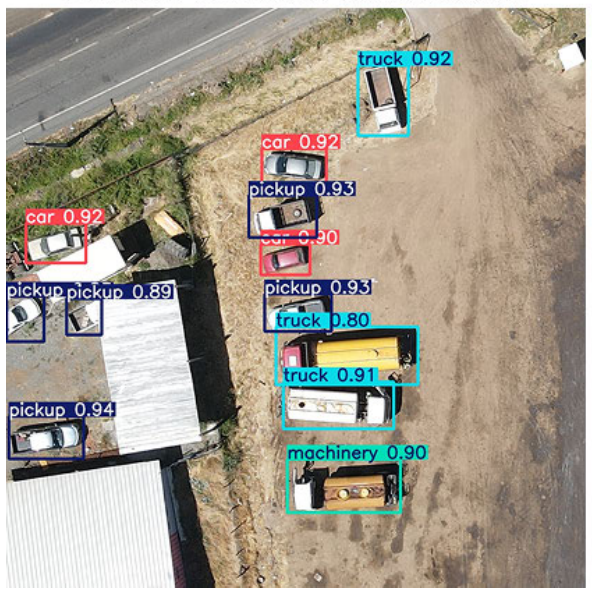

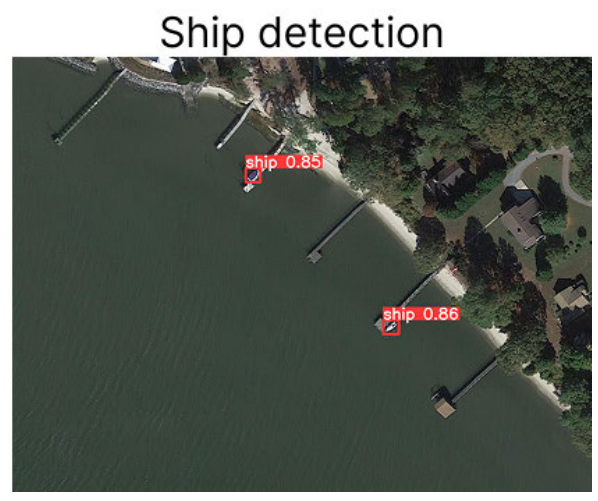

**FIGURE 36. Examples of object detection in remote sensing. Here, targets are typically captured from aerial or satellite perspectives.**

## *B. REMOTE SENSING*

Remote sensing is another domain in which YOLO has gained significant adoption, particularly for object detection tasks in satellite and aerial imagery, as shown in Fig. 36. These images, captured from high altitudes, offer broad spatial coverage but also pose unique challenges, including low object resolution, dense scene complexity, and substantial background clutter. Likewise, other imaging modalities are often used, different from the typical RGB, such as multispectral [130], [131], [132], SAR [133], [134], or LiDAR images [135], [136], [137], which allow a wider range of information to be captured. As a result, object detection models in this domain must be robust to scale variations and capable of localizing different targets embedded in heterogeneous environments.

Many studies [138], [139], [140], [141], [142], [143], [144] in this area focus on general-purpose object detection tasks, where YOLO is trained to identify and localize multiple object categories within remote sensing images. In these settings, models must handle diverse scene layouts and complex backgrounds characteristic of aerial and satellite views. A subset of studies [145], [146], [147], [148], [149], [150], [151] addresses more specific detection tasks involving small objects, where limited spatial resolution and dense object distributions introduce additional challenges. These scenarios require models capable of maintaining high sensitivity despite minimal pixel information per instance.

Other studies focus on specific object categories within remote sensing imagery. For instance, ship detection [152], [153], [154], [155], [156], [157], [158] is often applied in maritime surveillance tasks, where models must identify vessels across large oceanic scenes with low object density. Aircraft detection [159], [160], [161], [162], [163], [164] is typically used for monitoring airfields or restricted zones, requiring precise localization under varying viewpoints. Similarly, landslide detection [165], [166], [167], [168], [169] has been explored using YOLO to identify regions affected by terrain displacement, supporting early response in disaster-prone areas. Another task is building detection [170], [171], where YOLO is employed to delineate structures in urban or semi-urban environments, often under high density and occlusion conditions.

## *C. SMART AGRICULTURE*

Smart agriculture is a domain where YOLO-based models have been widely adopted, driven by the variety of visual tasks involved in monitoring plant health, managing crops, and automating farming processes. The diversity of objects, environments, and conditions in agricultural settings has led to numerous use cases in which object detection plays a central role. Fig. 37 shows detection examples in the domain of smart agriculture.

A fundamental application of YOLO in agriculture is the detection of crops and plant species across diverse environments, forming the basis for more specialized analyses and interventions. This task involves identifying various types of plants in open fields, orchards, and controlled settings such as greenhouses. Models are trained to recognize crops like orange [172], pomegranate [173], [174], mango [175], cherry [176], and apple [177], enabling automated field monitoring. YOLO has also been applied to detect other categories such as trees [178], [179] and flowers [180], [181]. This line of work extends to weed detection [182], [183],

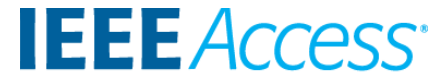

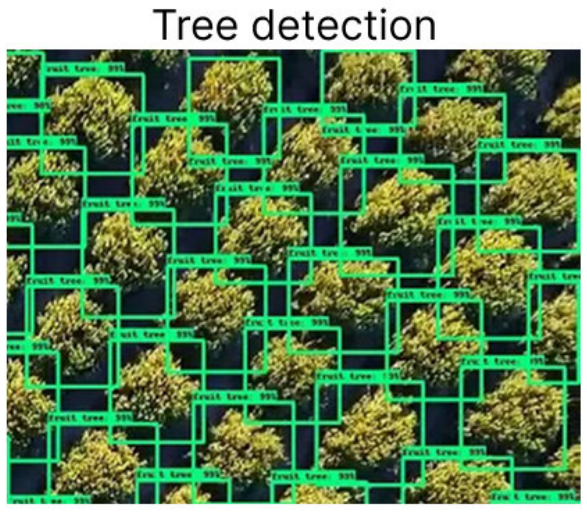

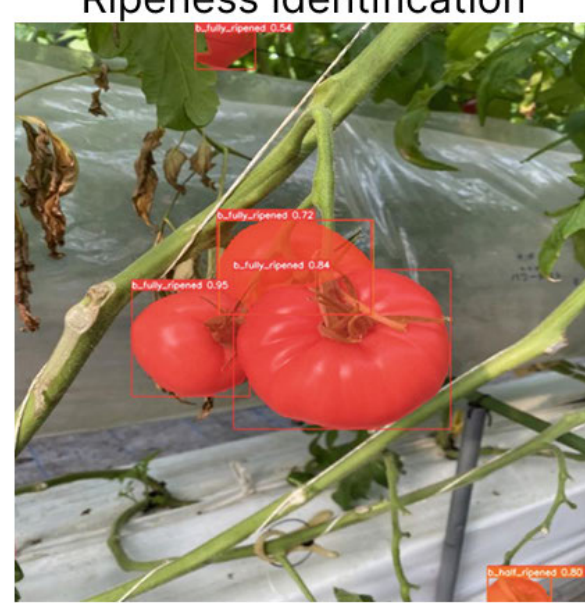

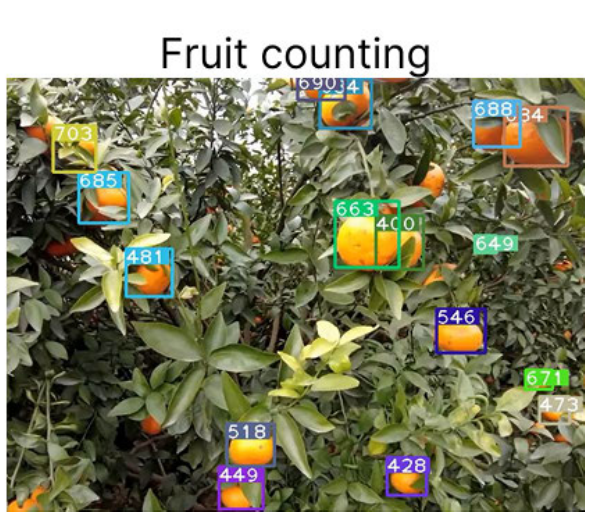

**FIGURE 37. Examples of object detection in smart agriculture, including fruit ripeness assessment, disease identification, and crop counting in orchard and greenhouse environments.**

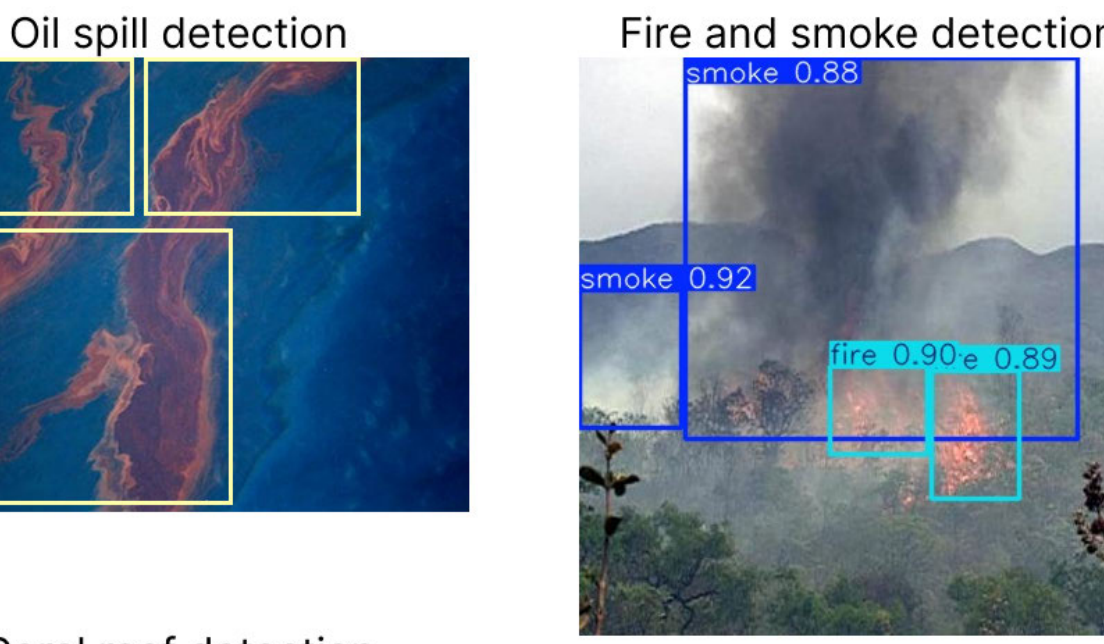

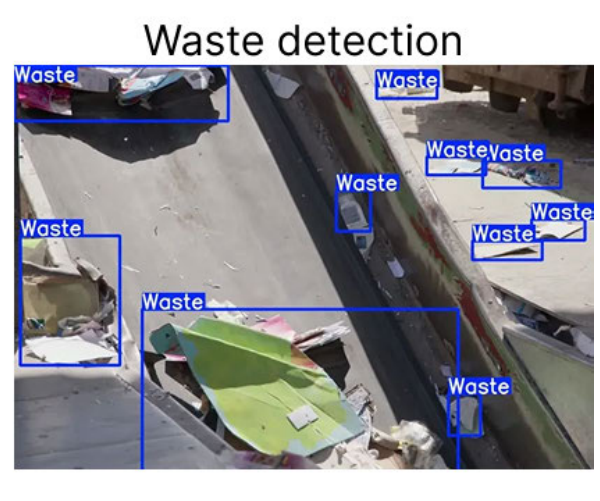

**FIGURE 38. Illustrative examples of detection-based environmental monitoring, showcasing how YOLO models assist in identifying ecological threats, tracking habitat conditions, and supporting sustainability efforts.**

[184], [185], [186], [187], where YOLO is used to distinguish invasive species from crops, supporting targeted removal and reducing herbicide usage.

Another widely explored application of YOLO in agriculture is the detection of leaf diseases. This task typically involves identifying visible symptoms such as spots, discoloration, or texture anomalies on plant leaves. YOLO-based models have been trained to detect and classify various diseases across different crops, including tea [188], [189], maize [190], bell pepper [191], rice [192], [193], and potato [194], [195], [196]. The ability to perform real-time detection in the field makes these models suitable for early diagnosis and large-scale monitoring, supporting timely intervention and minimizing yield loss.

Another relevant application is the assessment of fruit ripeness, where YOLO is used to detect fruits and classify them according to predefined maturity stages. This capability has been applied to crops such as strawberry [197], tomato [198], [199], [200], jujube [201], apple [202], and Camellia oleifera [203], supporting harvest planning and quality control. Similarly, YOLO has been adopted for fruit counting [204], [205], [206], [207], [208], where the goal is to estimate the number of visible fruits in orchards or greenhouse environments. This contributes to yield estimation and resource allocation, especially in contexts where manual counting is impractical or unreliable.

Finally, YOLO has also been integrated into harvesting robots to support various perception tasks in the field. Examples include the localization of picking points for strawberries and grapes [209], [210], the identification and tracking of pumpkins and asparagus spears [211], [212], and the coordination of object detection with robotic arms for automated fruit collection [213], [214]. This integration of object detection and robotics contributes to labor reduction and improves operational efficiency in large-scale agricultural settings.

### *D. ENVIRONMENTAL MONITORING AND WILDLIFE CONSERVATION*

Environmental monitoring and wildlife conservation have become increasingly relevant in the context of global sustainability efforts. These fields involve the observation, analysis, and protection of natural ecosystems, often requiring timely and accurate information acquisition. Such tasks are essential for preserving habitats and biodiversity, ensuring the resilience of vulnerable ecosystems. Owing to its capacity for fast and accurate visual processing, YOLO has become a valuable tool in this domain. Fig. 38 presents representative visualizations from the domain of environmental monitoring and wildlife conservation.

Within the broader scope of environmental monitoring, a substantial body of research focuses on detecting threats and mitigating risks that compromise ecosystem integrity. One prominent application is wildfire and smoke detection [215], [216], [217], [218], [219], where models are trained to identify fire outbreaks or smoke plumes in real time. These systems support early warning mechanisms and enable rapid response, helping to prevent large-scale forest loss and limit fire propagation in vulnerable regions. In aquatic environments, oil spill detection [220], [221], [222], [223] has emerged as another critical use case. YOLO models are employed

Blood cell detection

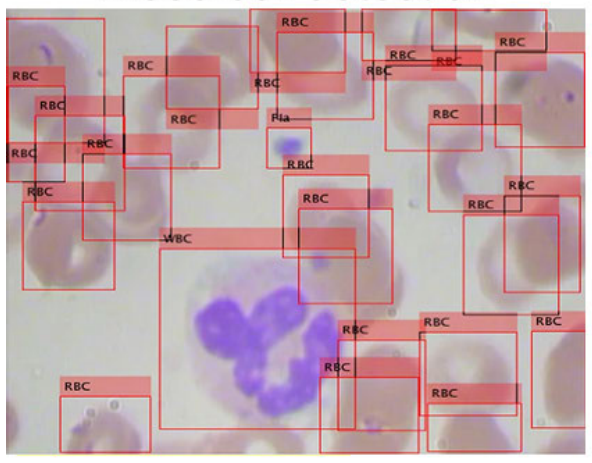

Cancer detection

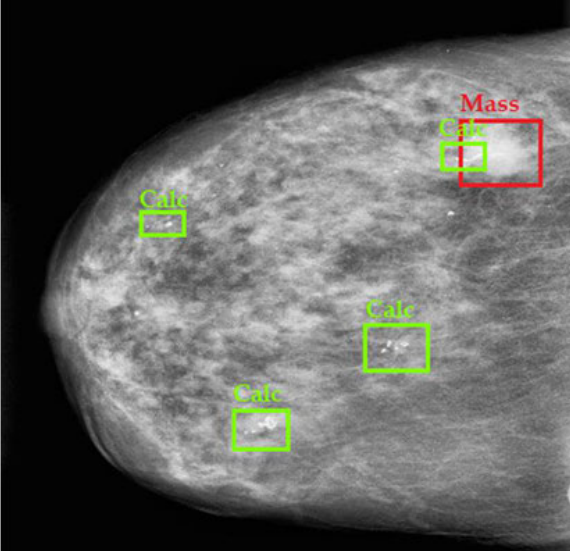

Bone fracture detection

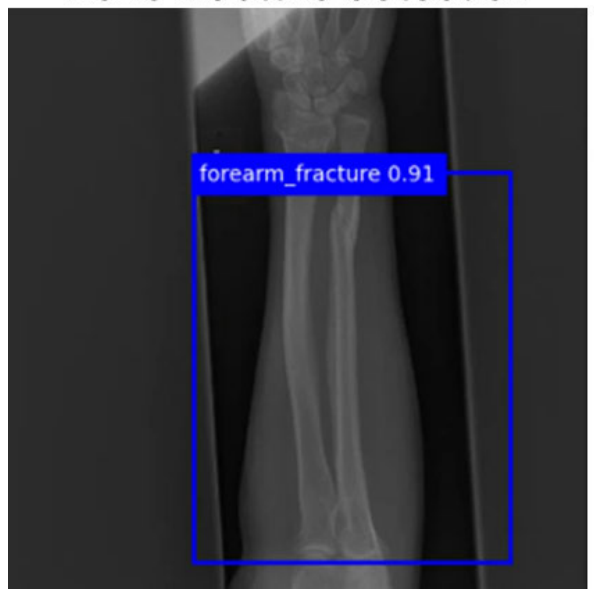

Tooth detection

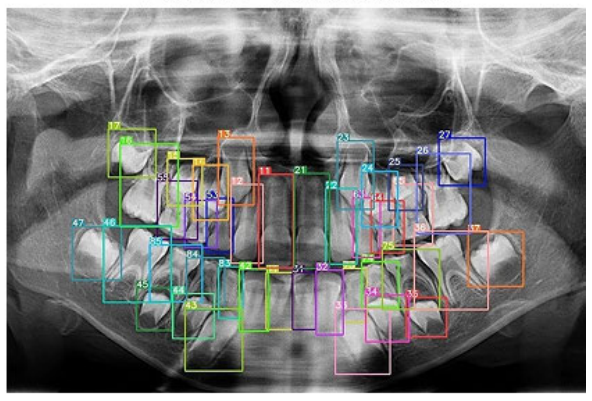

**FIGURE 39.** Examples of YOLO-based detection in medical imaging, encompassing tasks such as lesion identification, anatomical structure localization, and diagnostic support across different imaging modalities.

to detect the presence and extent of oil contamination on water surfaces, facilitating timely monitoring of industrial pollution and accelerating containment and cleanup efforts. Additionally, in the atmospheric domain, YOLO has been applied to the detection of air pollutants and hazardous gases, such as methane and aerosols [224], [225], which pose serious threats to air quality and environmental health.

Another relevant application is waste detection and management. Several studies [226], [227], [228], [229], [230], [231] have employed YOLO-based models to identify and classify different types of waste in real time, supporting cleanup operations, enabling automated recycling systems, and contributing to pollution reduction across diverse environments. YOLO has also been applied in flood-related tasks, both for real-time monitoring and post-disaster assessment. In the first case, models are used to detect early signs of flooding, such as rising water levels or overflowing rivers [232], [233], [234], [235], enabling timely alerts. In the second, post-event analyses focus on identifying flood extent, locating survivors, and assessing infrastructure damage [236], [237], [238]. This dual functionality enhances the capacity of response teams to act proactively and effectively during flood emergencies.

A shift in focus from environmental phenomena to fauna monitoring highlights another important aspect of environmental research. In this context, YOLO has been employed in multiple studies [239], [240], [241], [242], [243], [244], [245], [246] for animal tracking and endangered species detection. These tasks support the identification and monitoring of wildlife in natural habitats, contributing to population control, behavioural analysis, and biodiversity research. Moreover, YOLO models are often integrated into camera trap systems [247], [248], which enable non-intrusive observation of animal activity in the wild. These devices, typically triggered by motion or heat, allow continuous monitoring of species presence and movement without direct human intervention.

Another important group of applications in wildlife monitoring involves animal counting [249], [250], which supports population estimation, migration analysis, and ecological balance assessment. In these cases, YOLO-based models are used to detect and tally animals, offering scalable alternatives to manual annotation and field surveys. Along similar lines, several studies have leveraged YOLO for poaching prevention and anti-trafficking efforts [251], [252], [253], [254], enabling the surveillance of protected areas and assisting in the detection of illegal activities that threaten endangered species.

Beyond fauna, YOLO has also been applied to the monitoring and protection of vulnerable flora, particularly in marine ecosystems [255]. Coral reefs, which play a crucial role in ocean biodiversity and are highly sensitive to environmental changes, have been the focus of several studies [256], [257], [258], [259], [260]. In this context, YOLO-based models are used to detect and localize coral structures in underwater environments, supporting habitat mapping, health assessment, and the early detection of reef degradation.

### *E. MEDICAL IMAGE AND DIAGNOSIS*

Medical imaging represents a critical domain where automated object detection can enhance diagnostic workflows. With the growing availability of digital medical scans and the demand for rapid, accurate interpretation, YOLO has been increasingly adopted to assist in localizing pathological regions across different imaging modalities. These applications are designed to support clinical decision-making by reducing interpretation time and improving consistency in diagnosis. Fig. 39 displays detection outputs from diverse medical imaging contexts, illustrating how YOLO models can assist in clinical diagnosis through automated analysis and interpretation.

Cancer detection is one of the most prominent and extensively studied applications of YOLO in the medical domain. Models are trained to identify irregular patterns, abnormal tissue growth, nodules, and suspicious masses that may indicate the presence of cancer. For instance, YOLO has been applied to breast cancer detection [15], [261], [262], [263], [264], [265] using mammographic images, and to lung cancer identification [266], [267], [268], [269], [270] through Computed Tomography (CT) scans. Other implementations include skin cancer detection [271], [272], [273] with dermoscopic images, brain tumor recognition [274], [275], [276], [277], [278], [279] using Magnetic Resonance Imaging (MRI), and colorectal polyp detection [12], [280], [281], [282], [283] from endoscopic imagery.

Another active area of research involves the detection of bone and dental abnormalities. YOLO-based models have been employed to identify fractures in X-ray images [284], [285], [286], [287], [288], [289], enabling faster and more consistent assessment in emergency and trauma care. These models localize disruptions in bone continuity or structural anomalies indicative of fractures across various anatomical regions, helping reduce diagnostic delays. Similarly, in dental radiography, YOLO has been applied to detect caries, missing teeth, and implants [290], [291], [292], [293], [294], [295], supporting dental diagnostics and treatment planning through automated analysis.

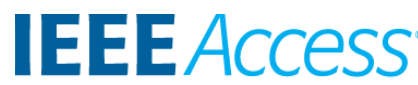

A further area of application is ophthalmology, where YOLO models have been used to support the detection of common eye diseases. For glaucoma, detection focuses on analyzing structural features in the optic disc, such as the cup-to-disc ratio, from retinal fundus images [296], [297], [298]. These indicators are essential for identifying early signs of optic nerve damage and enabling timely intervention. In the case of diabetic retinopathy, YOLO has been applied to detect retinal anomalies including microaneurysms, haemorrhages, and exudates [299], [300], [301], all of which are key indicators of disease progression in diabetic patients.

Finally, YOLO has also been applied to a range of auxiliary tasks within the medical domain. One example is blood cell detection [302], [303], [304], [305], which supports hematological analysis by identifying and classifying different cell types in microscopic images. Another relevant application is the detection and tracking of surgical instruments in operating rooms [306], [307], [308], [309], facilitating workflow analysis, safety monitoring, and integration with robotic surgery systems.

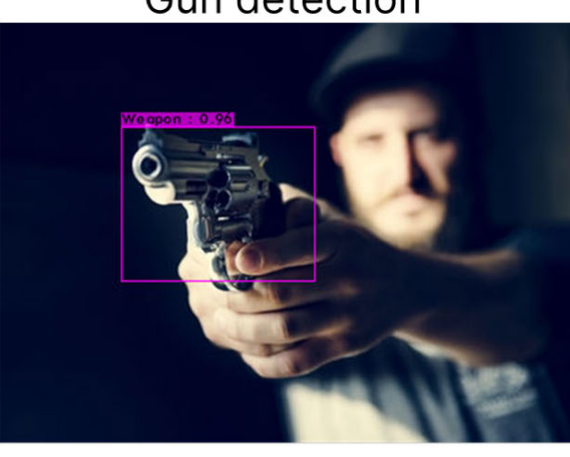

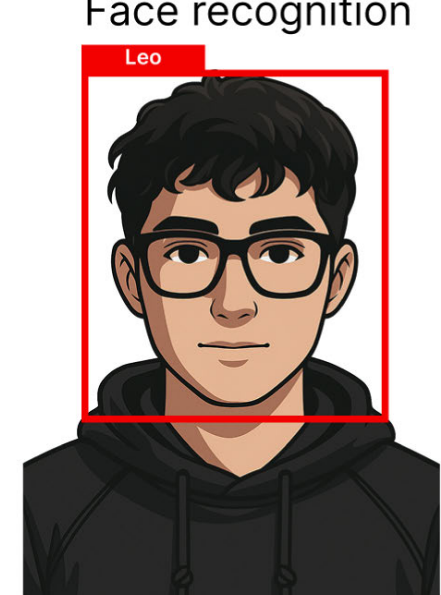

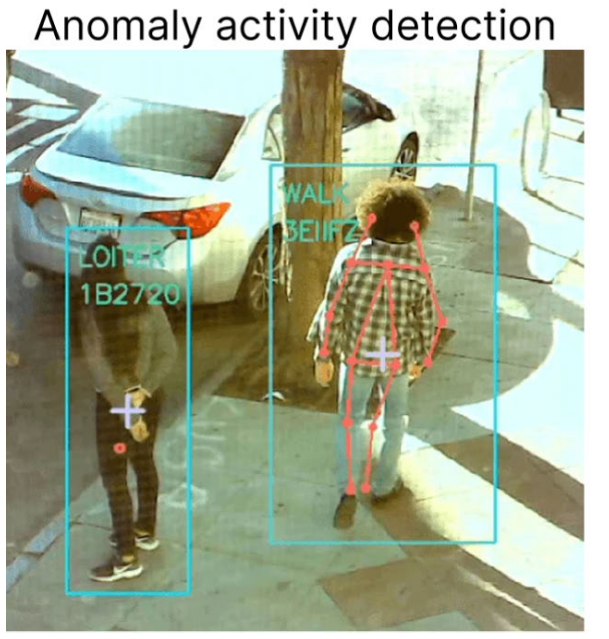

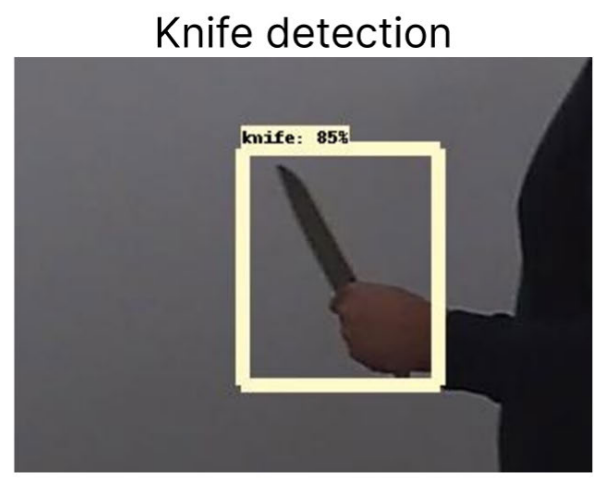

**FIGURE 40. Detection examples in the context of security systems, including weapon identification, suspicious behaviour monitoring, and individual recognition in surveillance settings.**

## *F. SECURITY SYSTEMS*

Security systems are another domain where real-time object detection plays a pivotal role. From monitoring public spaces to enforcing access restrictions, these systems depend on accurate and timely identification of potential threats or anomalous activities. YOLO-based models have gained significant traction in this field due to their ability to deliver fast and reliable detections, making them suitable for integration into both large-scale surveillance infrastructure and compact embedded security solutions. Fig. 40 presents detection results from various security-related scenarios, illustrating how YOLO models contribute to enhancing surveillance and threat detection.

One key application of YOLO in security systems is the rapid identification of potential threats. In particular, YOLO has been widely used for weapon detection, with models trained to recognize objects such as guns [310], [311], [312], [313], knives [314], [315], [316], [317], and other handheld weapons [318], [319], [320]. These systems are designed to operate in real time within crowded or sensitive environments, including airports, schools, and public events. Typically integrated into surveillance infrastructures such as Closed-circuit Television (CCTV) networks, YOLO models continuously analyze video streams to flag the presence of suspicious items and support rapid intervention by security personnel.

Another relevant application in security systems is crowd analysis, where YOLO-based models are used to monitor the movement and density of individuals in public spaces [321], [322]. This facilitates the identification of potentially hazardous situations, such as overcrowding or stampede risk. Building on this capability, several systems have been developed to detect abnormal or suspicious behaviours [323], [324], [325], [326], [327], [328], with models trained to recognize actions that deviate from typical human patterns, such as running in restricted areas or prolonged loitering. YOLO has also been applied to the detection of specific criminal activities, including theft and vandalism [329], [330], [331], [332], contributing to the early identification of threats and enabling timely intervention by surveillance teams.

Additionally, intrusion detection represents a key task in many surveillance systems, particularly in restricted or high-security areas. YOLO has been integrated into monitoring infrastructures to automatically identify unauthorized entries by detecting individuals crossing predefined virtual boundaries or entering without proper clearance [333], [334], [335], [336], [337]. These systems are commonly deployed in facilities such as warehouses, industrial zones, and governmental buildings to prevent security breaches and maintain perimeter integrity. Along similar lines, YOLO has also been applied to access control and identity verification tasks, where models are trained to detect and authenticate authorized personnel based on visual attributes such as facial features [338], [339], fingerprints [340], [341], or iris patterns [342], [343]. These implementations help streamline entry processes, reduce reliance on manual verification, and enhance the security of controlled environments.

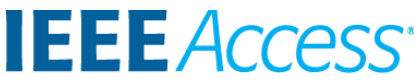

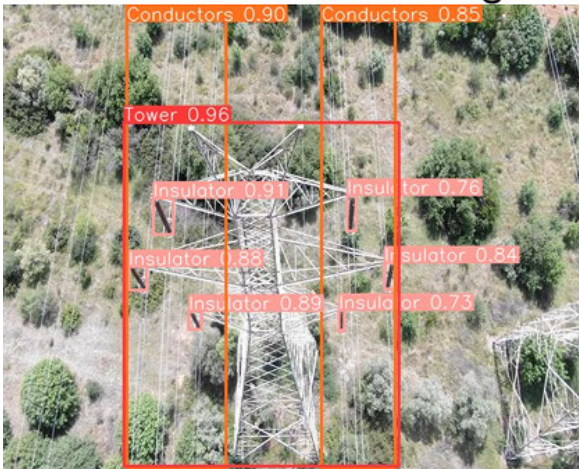

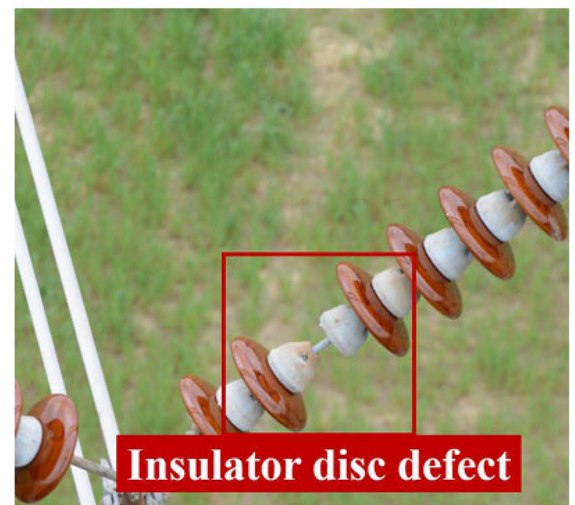

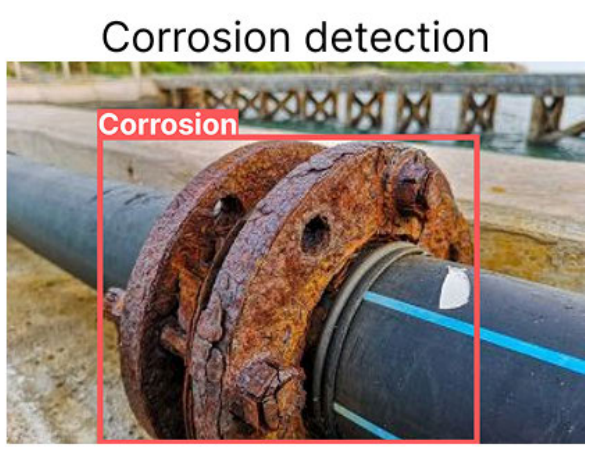

**FIGURE 41. Examples of YOLO-based detection in infrastructure monitoring, where visual analysis supports the identification of structural issues and the planning of timely interventions.**

Finally, YOLO also gained significant relevance during the COVID-19 pandemic through its application in face mask detection. In many public and private spaces, access was restricted for individuals not wearing masks, prompting the deployment of automated surveillance systems capable of verifying mask usage in real time. Several studies [344], [345], [346], [347], [348], [349], [350] leveraged YOLO models for this purpose, demonstrating their effectiveness in real-world conditions. This use case highlights not only the adaptability of YOLO-based detection frameworks, but also their capacity to transition from academic research to operational deployment.

### G. INFRASTRUCTURE MONITORING

Infrastructure monitoring encompasses a broad range of applications aimed at ensuring the safety, functionality, and longevity of physical assets. From large-scale civil structures to industrial facilities and energy networks, maintaining operational integrity is essential to prevent failures and reduce maintenance costs. YOLO has emerged as a valuable tool in this context due to its capacity to detect visual anomalies efficiently and accurately across diverse types of infrastructure. Fig. 41 shows representative detection examples in infrastructure monitoring, illustrating how YOLO can support the identification of structural damage and maintenance needs.

Defect detection is a central application of YOLO in infrastructure monitoring. Models are trained to identify visual anomalies such as deformations or surface irregularities that may compromise structural integrity. The applications span multiple domains, ranging from general metallic structures [351], [352], [353], [354] to more specific cases such as weld quality assessment [355], [356], [357], [358], bridge inspection [359], [360], [361], [362], [363], and defect detection in solar cells [364], [365], [366]. In these contexts, early detection is crucial, as it enables timely maintenance interventions that can prevent system degradation, avoid costly repairs, and reduce the risk of catastrophic failures.

Another relevant application of YOLO in infrastructure monitoring involves the inspection of electrical systems. These include high-voltage transmission lines [367], [368], [369], [370], [371], [372], towers [373], [374], and insulators [375], [376], [377], [378], [379], [380], where faults such as broken components, vegetation encroachment, or thermal anomalies can disrupt energy distribution and pose serious safety risks. YOLO-based models are integrated into aerial or ground-based inspection systems to detect such issues in real time. This automation reduces the need for hazardous manual inspections and enables more frequent, large-scale monitoring, helping utilities maintain grid stability and respond promptly to potential failures.

The oil and gas industry has also integrated YOLO-based methods for infrastructure surveillance, particularly in pipeline inspection. These systems are typically designed to detect surface cracks [381], [382], [383], corrosion [384], [385], leakage points [386], [387], and other structural degradations that may compromise pipeline integrity. Such tasks are essential for maintaining operational safety, enabling timely maintenance interventions, and minimizing the risk of service disruptions.

As a complementary application, YOLO has also been used to enhance industrial safety by monitoring compliance with Personal Protective Equipment (PPE) requirements. Models are trained to detect whether workers are properly equipped with items such as helmets, vests, gloves, and safety glasses [388], [389], [390], [391], [392], [393], [394]. This visual monitoring capability enables automated enforcement of safety protocols, contributing to injury prevention and improved risk management in hazardous work environments.

## VII. ANALYTICAL OUTLOOK ON YOLO

### A. ARCHITECTURAL EVOLUTION

#### 1) BACKBONE PROGRESSION

The progression of backbones across YOLO variants reflects a clear shift in design philosophy, evolving from straightforward depth-based expansion toward more deliberate strategies centered on modular reuse, gradient flow regulation, and structural compactness. Each iteration has contributed not only to gains in accuracy, but also to increasingly efficient mechanisms for extracting, reusing, and stabilizing representational power through more principled architectural designs.

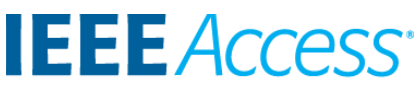

In its early stages (YOLOv1 to YOLOv3), YOLO adhered to a conventional convolutional paradigm, relying on increased depth and filter complexity to enhance feature extraction. This approach culminated in Darknet-53, a deeper and more capable network that became a standard for several years. However, it soon became clear that increasing depth alone was not a sustainable strategy, particularly in light of the growing demand for real-time inference on resource-constrained devices.

YOLOv4 marked the first significant structural departure by introducing CSP connections, aimed at reducing redundancy and stabilizing training. This marked the beginning of a broader shift toward efficient gradient routing and selective feature reuse, principles that have since become central to the evolution of YOLO's backbone.

From YOLOv5 onward, the architectural design shifted not only in backbone type but also in framework, moving from Darknet to PyTorch. From this point, the focus moved away from radical changes in backbone structure toward incremental refinements and internal optimizations. These included adjustments in feature map aggregation and internal block design. Starting with YOLOv8, for example, rather than adopting a new external architecture, the focus turned to rethinking internal modules to improve expressiveness while preserving efficiency. This reflects a consistent trend in later versions: maintaining overall architectural continuity while upgrading internal mechanisms to enhance information flow.

YOLOv9 and subsequent versions further illustrate this trend by introducing sophisticated aggregation mechanisms, not by stacking additional layers, but by reorganizing the interactions among existing ones. These changes aim to maximize parameter efficiency and gradient stability without compromising inference speed. Notably, rather than adopting external backbone families such as EfficientNet or ResNet, as seen in other detection models, YOLO continues to develop purpose-built backbones specifically tailored for detection, with a strong emphasis on tight integration between the backbone, neck, and head.

Another emerging pattern in recent versions, particularly YOLOv11, YOLOv12, and YOLOv13 is the increasing use of lightweight, attention-inspired modules. These are designed to improve spatial and channel interactions, but are applied selectively to avoid introducing latency overhead. This careful balance between expressiveness and speed reflects a more mature design philosophy, as YOLO backbones evolve from being mere feature extractors to becoming specialized enablers of real-time perception.

#### 2) EVOLUTION OF NECK MODULES

The neck component in YOLO architectures, which bridges the backbone and the detection head, has undergone significant evolution, transitioning from absence to becoming a key focus of architectural innovation. In YOLOv1 and YOLOv2, there was no formal neck; these models forwarded features directly from the backbone to the head. While this minimalistic design offered speed, it limited the ability to incorporate fine-grained features from early layers or to handle multi-scale objects effectively. A formal neck was first introduced in YOLOv3 through the integration of a FPN, laying the groundwork for multi-scale feature aggregation.

Early neck designs like FPN and PANet emphasized multi-scale aggregation, but offered limited adaptability. As versions progressed, modular enhancements like CSP integration, reparameterization (Rep-PAN), quantization-aware reparameterized blocks (QA-RepVGG), and refined pooling strategies (SPPF) enabled the network to preserve multi-scale features while accelerating inference. This reflects a broader move toward agile fusion mechanisms that adapt to varied object sizes and scene complexity without compromising speed.

A second trend is the growing interplay between the neck and training dynamics. While early designs focused primarily on inference-time performance, later iterations, particularly from YOLOv9 onward, began to treat the neck as a training-aware module. Mechanisms such as PGI introduced auxiliary gradient pathways, signaling a paradigm shift: the neck is no longer limited to feature aggregation but also contributes to training stability, convergence optimization, and information flow regulation across the model's depth. This functional duality aligns with the increasing architectural complexity of modern YOLO variants.

Third, there is a growing integration of attention mechanisms, consistently constrained by the need for real-time performance. From spatial attention modules such as C2PSA, to area-based approximations like A2, and memory-efficient strategies like FlashAttention, recent versions reveal a clear pattern: attention is integrated only in forms that scale with the practical demands of high-speed inference. In this context, YOLOv13 introduces FullPAD, a full-pipeline aggregation-and-distribution system that channels enriched features throughout the backbone, neck, and head via dedicated pathways, enhancing both information flow and gradient propagation. This balance between selective focus and computational feasibility reflects a shift in how attention is understood, not as a passing trend but as a calibrated design instrument.

Finally, there is a clear architectural convergence between the neck and the backbone. Beginning with C2f and subsequent modules, a unified design logic emerges across components, positioning the neck as a structural extension rather than an isolated block. This modular alignment helps reduce training inconsistencies, simplifies model scaling, and reinforces the coherence of the network's hierarchical representation. Recent additions such as DS-C3k, DS-C3k2, and C3AH extend this trend by combining lightweight separable convolutions with adaptive hypergraph semantics, further tightening the integration of backbone and neck while preserving efficiency.

#### 3) SHIFTS IN DETECTION HEADS

Compared to the substantial architectural evolution observed in backbones and necks, the detection head in YOLO

models has undergone a more gradual transformation. Early versions, such as YOLOv1 and YOLOv2, used a monolithic, fully connected head directly attached to the backbone, responsible for predicting class probabilities and bounding box coordinates simultaneously. This coupled design persisted through YOLOv3 to YOLOv5, where the adoption of multi-scale detection via FPN and PANet enabled the head to operate at different spatial resolutions. However, the core principle of joint regression and classification remained largely unchanged.

A notable turning point appears with YOLOX and YOLOv6, where decoupling emerges as a central design priority. These models introduce separate branches within the detection head to independently handle classification, regression, and objectness estimation. This structural separation mitigates the representational conflicts inherent in shared feature spaces, enabling more specialized and stable learning. The decoupled architecture becomes a defining feature of YOLOv8 and later versions, reflecting a broader shift in object detection frameworks toward disentangled task processing to improve precision and convergence speed.

Beyond structural modifications, some versions have introduced training-time mechanisms to refine supervision. YOLOv7 implemented an auxiliary head for deep supervision through soft label generation, while YOLOv10 expanded on this with the Consistent Dual Assignments strategy, which harmonizes one-to-many and one-to-one label assignments, enabling rich supervision during training without incurring inference-time overhead. These innovations aim to narrow the gap between training and deployment dynamics, supporting stable optimization and latency-aware inference.

Recent versions, particularly YOLOv11 and YOLOv12, preserve the decoupled structure while incorporating lightweight attention mechanisms such as C2PSA and Area Attention into the head. Although these additions are more restrained than the extensive changes introduced in the backbone or neck, they reflect an ongoing effort to enhance spatial focus and prediction accuracy, especially for small or densely packed objects. Notably, this also indicates a shift toward adopting attention mechanisms as core components across all stages of the YOLO architecture, marking a departure from earlier versions where such modules were avoided due to concerns over computational overhead.

The architectural evolution of YOLO models over the past decade reveals a consistent shift in priorities. Rather than pursuing aggressive improvements in mAP, recent developments have focused on efficiency, training stability, and deployment feasibility. Each new version has increasingly prioritized practical aspects such as easier optimization, improved gradient flow, multi-scale robustness, and faster inference, all while maintaining competitive accuracy. As shown in Table 16, although variations in mAP exist, models from YOLOv5 to YOLOv13 demonstrate steady performance under the challenging MS-COCO benchmark. Early versions, such as YOLOv1, were evaluated on simpler datasets like Pascal VOC and lacked support for multiple object predictions per grid cell, limiting the relevance of direct comparisons. The consistent results across recent versions suggest a matured architecture focused on reliability and real-world applicability rather than chasing peak benchmark scores, reinforcing YOLO's role as a versatile and efficient solution for object detection in practical scenarios.

**TABLE 16. Summary of representative YOLO models illustrating the evolution of performance, architectural choices, and implementation frameworks across a decade of development. For each architecture, the highest-performing variant has been reported.**

| Model | AP (%) | Anchor | Head | Dataset | Framework | Year |
|---|---|---|---|---|---|---|
| YOLOv1 | 63.4 | × | Coupled | Pascal VOC | Darknet | 2015 |
| YOLOv2 | 78.6 | ✓ | Coupled | Pascal VOC | Darknet | 2016 |
| YOLOv3 | 57.9 | ✓ | Coupled | MS-COCO | Darknet | 2018 |
| YOLOv4 | 44.3 | ✓ | Coupled | MS-COCO | Darknet | 2020 |
| YOLOv5 | 50.7 | ✓ | Coupled | MS-COCO | PyTorch | 2020 |
| PP-YOLO | 45.2 | ✓ | Coupled | MS-COCO | PaddlePaddle | 2020 |
| Scaled-YOLOv4 | 55.8 | ✓ | Coupled | MS-COCO | PyTorch | 2021 |
| YOLOR | 55.4 | ✓ | Coupled | MS-COCO | PyTorch | 2021 |
| YOLOvX | 51.5 | × | Decoupled | MS-COCO | PyTorch | 2021 |
| YOLOv6 | 52.8 | × | Decoupled | MS-COCO | PyTorch | 2022 |
| YOLOv7 | 56.8 | × | Decoupled | MS-COCO | PyTorch | 2022 |
| YOLOv8 | 53.9 | × | Decoupled | MS-COCO | PyTorch | 2023 |
| YOLO-NAS | 52.2 | × | Decoupled | MS-COCO | PyTorch | 2023 |
| YOLOv9 | 55.6 | × | Decoupled | MS-COCO | PyTorch | 2024 |
| YOLOv10 | 54.4 | × | Decoupled | MS-COCO | PyTorch | 2024 |
| YOLOv11 | 54.7 | × | Decoupled | MS-COCO | PyTorch | 2024 |
| YOLOv12 | 55.4 | × | Decoupled | MS-COCO | PyTorch | 2025 |
| YOLOv13 | 54.8 | × | Decoupled | MS-COCO | PyTorch | 2025 |

## B. RELEVANT PARADIGM SHIFTS AND DETECTION STRATEGIES

### 1) MULTI-SCALE PREDICTION AND SPATIAL ROBUSTNESS

Multi-scale prediction marked a fundamental transition in YOLO's evolution. Early versions, such as YOLOv1, performed detection at a single resolution level, which limited their ability to capture objects of varying sizes within an image. This constraint became particularly evident in complex scenes, where small or distant objects were often missed or poorly localized. Multi-scale prediction was introduced in YOLOv3 and retained in all subsequent versions. Instead of relying on a single output resolution, the detection head produces feature maps at multiple spatial scales, enabling the model to identify both small and large objects more effectively.

Throughout YOLO's development, multi-scale detection has remained a central design element. Although the specific implementations vary, with approaches such as FPN, PANet, SPPF, or C2f, the core objective has been consistent, as it aims to enhance spatial robustness and maintain detection quality across object sizes. Its continued presence, despite evolving fusion strategies and architectural adjustments, underscores its role as a foundational principle. Rather than a one-time enhancement, multi-scale prediction has become a persistent and prioritized strategy throughout YOLO's decade-long evolution.

### 2) FROM ANCHOR-BASED TO ANCHOR-FREE

The progression from anchor-based to anchor-free detection paradigms represents one of the most decisive shifts in the development of YOLO. YOLOv1 did not include the concept of anchors; instead, it relied on a rigid grid-based strategy in which each cell predicted bounding boxes directly. YOLOv2 introduced anchor boxes, enabling models to better approximate object shapes and scales using predefined templates. This anchor-based approach remained dominant for several years, characterizing models from YOLOv2 through YOLOvR.

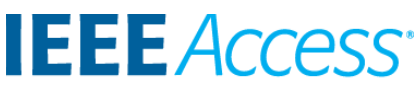

However, the anchor-based paradigm carried inherent limitations, such as the need for careful anchor configuration and sensitivity to object scale distributions. The adoption of anchor-free strategies, beginning with YOLOX, marked a deliberate departure from this long-standing methodology. By removing the dependence on predefined priors, anchor-free designs allow for direct prediction of object centers and sizes. This shift has been consistently maintained in newer versions, becoming a consolidated design principle in the modern evolution of YOLO.

#### 3) NON-MAXIMUM SUPPRESSION AND ITS ALTERNATIVES

NMS has traditionally been a core component in the prediction pipeline of YOLO. While effective, this approach introduces a degree of rigidity by discarding overlapping information and exhibiting sensitivity to threshold selection. For several versions, NMS remained unchanged, reflecting its deeply rooted role in real-time object detection frameworks.

Attempts to move beyond traditional NMS began to emerge gradually. YOLOX introduced an optional end-to-end training mode based on one-to-one label assignment and stop-gradient operations, enabling the model to bypass NMS entirely during inference. However, this configuration involved slight performance trade-offs and did not become the default. Later, YOLOv8 optimized the NMS stage indirectly by reducing the number of candidate boxes per prediction, improving inference speed while retaining standard NMS.

A more decisive step was taken with YOLOv10, which formally removed the need for NMS through the Consistent Dual Assignments strategy. This approach enables direct prediction without redundancy, aligning the training and inference processes more closely. As a result, YOLOv10 operates in a fully end-to-end manner, streamlining the inference pipeline and reducing post-processing complexity. More importantly, it demonstrated for the first time that removing NMS is not only feasible but can be achieved without compromising detection quality.

Despite these advances, subsequent versions such as YOLOv11, YOLOv12, and YOLOv13 have not explicitly continued in this direction, suggesting that while the removal of NMS remains a promising avenue, it has not yet been consolidated across the YOLO family. The persistence of traditional NMS alongside isolated alternatives reflects an ongoing tension between architectural innovation and the reliability of established mechanisms. Although the architectural possibility exists, its generalization and consistent effectiveness across contexts remain open challenges.

### C. TRAINING TECHNIQUES AND OPTIMIZATION

#### 1) DATA AUGMENTATION METHODS

Throughout the development of YOLO, data augmentation has remained a key strategy for mitigating overfitting and improving generalization. In early versions such as YOLOv1, augmentation primarily involved geometric transformations and color perturbations in the HSV space. These conventional techniques laid the groundwork for more advanced strategies introduced in later iterations.

A clear turning point came with the introduction of advanced augmentation techniques such as Mosaic, MixUp, and CutMix. These compositional methods not only diversified the training data but also simulated more challenging visual conditions, improving robustness to occlusion and scale variation. Models such as YOLOv5 and YOLOX retained and expanded this approach by incorporating spatial, color-based, and compositional augmentations, including random affine transformations and Copy-Paste, into a standardized training pipeline.

In recent versions, strategies such as Mosaic, MixUp, and Copy-Paste have become increasingly stable and recurrent, forming a consistent augmentation foundation across the YOLO family. This shift highlights the contrast between early versions, which relied on basic perturbations to prevent overfitting, and later models, which consolidated a robust framework supporting scalability and generalization. The consistency of this pipeline indicates that data augmentation has evolved from a supplementary technique into a core design element, closely tied to training efficiency and convergence behaviour across architectures.

#### 2) LOSS FUNCTION EVOLUTION

The evolution of loss functions in YOLO models reflects a broader shift from simplistic formulations toward more specialized and task-aligned optimization strategies. In early versions, such as YOLOv1, object detection was framed as a pure regression task, with all components including localization, objectness, and classification optimized using sum-squared error. While conceptually simple, this formulation lacked the precision needed for accurate spatial localization and introduced instability when objects varied significantly in scale or shape.

As YOLO matured, cross-entropy gradually replaced squared losses for classification, while localization loss transitioned toward IoU-based formulations. Starting with YOLOv4, the adoption of Complete IoU loss marked a turning point, shifting from pure coordinate differences to metrics that account for geometric overlap and alignment quality. This allowed the models to better penalize inaccurate localization, even when predicted and ground-truth boxes partially overlapped.

Subsequent models explored more granular and distribution-aware variants, such as DFL for bounding box regression and VFL for classification. These functions aimed to enhance the training signal, particularly in scenarios involving class imbalance or ambiguous localization. Recent YOLO versions have not replaced the overall loss structure, but have instead refined individual components to align with architectural changes.

Overall, the progression of loss functions reflects a steady evolution in how YOLO models manage precision and stability during training. These changes did not involve abrupt shifts but rather a gradual refinement of existing ideas, aligning loss design with architectural updates such as anchor-free detection and decoupled heads. While the architecture has evolved rapidly, the loss components have advanced more conservatively, stabilizing methods that promote accuracy, training consistency, and generalization across scales.

#### 3) ALTERNATIVE TRAINING STRATEGIES

Recent research has explored semi-supervised extensions of YOLO as a way to reduce reliance on large annotated datasets [395]. For example, approaches such as SSDA-YOLO [396] adopt a Mean Teacher paradigm, where a teacher network generates pseudo-labels for unlabeled data, and a student network refines its learning based on both labeled and pseudo-labeled inputs. This framework allows YOLO to leverage abundant unlabeled samples while maintaining detection accuracy, and has shown improvements in cross-domain scenarios such as transferring from clear-weather datasets to adverse conditions like fog or rain. Semi-supervised YOLO therefore illustrates a paradigm in which annotation bottlenecks can be mitigated by combining labeled and unlabeled data in the training pipeline.

Knowledge distillation has also been investigated as a strategy to optimize YOLO and related detectors for deployment in resource-constrained environments. In this paradigm, a larger and more accurate teacher model guides the training of a smaller student model [397], transferring knowledge through soft targets or intermediate feature maps. Methods such as Scale-Equivalent Distillation (SED) [398] address scale imbalance and false negatives, enhancing detection robustness in dense or complex scenes. When applied to YOLO, such distillation frameworks enable the creation of lightweight yet competitive variants, reinforcing the broader trend of tailoring detection models to specific hardware and efficiency requirements.

### D. MODEL SCALING AND VARIANT DESIGN

The concept of model scaling represents a decisive shift in the design philosophy of the YOLO family. Early versions, from YOLOv1 through YOLOv4, were released as single-model solutions with no size variants, limiting their adaptability to different computational environments. This fixed configuration constrained deployment, particularly in scenarios requiring lightweight models for edge or embedded systems such as Jetson Nano or Orin NX.

From YOLOv5 onward, model variants became a standard part of release cycles, introducing multiple configurations (e.g., nano, small, medium, large, extra large) to accommodate different trade-offs between speed, size, and accuracy. This practice is not an isolated decision but a consolidated design choice across newer versions, including YOLOv6 through YOLOv13, and extends into derivative projects such as YOLO-NAS.

The rationale behind model scaling goes beyond performance benchmarking, addressing the increasing need for flexible deployment in resource-constrained environments. The ability to select a model variant based on available hardware has made YOLO particularly well suited for edge computing scenarios, where limitations in memory, processing power, and latency impose strict operational constraints.

This model scaling approach not only enables real-time inference in lightweight setups, but also contributes to the democratization of object detection by eliminating the need for high-end GPUs or server-based deployment. Unlike other frameworks, particularly region-based methods with heavier and less configurable designs, YOLO's scalable architecture allows practitioners to tailor model selection to specific use cases without compromising speed or accuracy.

The flexibility demonstrated by later YOLO models is arguably a key factor behind the framework's widespread adoption across diverse domains and applications. It reinforces YOLO's position as a leading solution for real-time object detection in both academic and industrial settings.

### E. FORKING AND ECOSYSTEM FRAGMENTATION

Unlike other object detection frameworks developed under the direction of a single research group or institution, the YOLO family has evolved through a decentralized and fragmented process. Following the original versions by Redmon et al., subsequent models have emerged from a diverse range of institutions, companies, and independent developers, leading to a dynamic and heterogeneous ecosystem. This includes contributions from both academic and industrial actors, such as Ultralytics, Megvii, Baidu, Meituan, Tsinghua University, among others.

This openness has played a key role in accelerating innovation, as reflected in the diverse architectural directions pursued by forks such as YOLOv5, YOLOX, and YOLO-NAS. Developers have contributed custom modules, novel training strategies, and varied optimization objectives, often aligned with specific deployment requirements. This decentralization has enabled YOLO to evolve rapidly and respond to emerging demands in ways that would be difficult to achieve under a centralized development pipeline.

In contrast to other detector families, YOLO has demonstrated a distinctive ability to sustain its relevance over more than a decade. SSD, although efficient, struggles with accuracy particularly on small objects and cluttered scenes, which limits its long-term adoption. Faster R-CNN and other region-based approaches achieve high accuracy but at the cost of heavy computation, making them less practical for real-time or resource-constrained applications. YOLO, by design, combines competitive accuracy with unmatched inference speed while steadily incorporating architectural refinements that preserve efficiency across hardware scales. These qualities have ensured YOLO's survival and continuing evolution, whereas SSD and Faster R-CNN remain important historical milestones but no longer define the state-of-the-art on object detection.

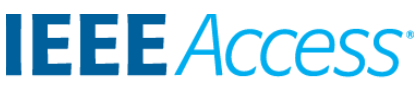

However, an important point to consider is the accelerating pace of releases. While early YOLO models were separated by years, recent iterations have appeared at a much faster rate, sometimes multiple times within a single year. This can be seen as a sign of community vitality, but it also raises concerns about long-term cohesion and the risk of saturating the field with incremental versions that may dilute YOLO's conceptual identity. Whether this trend will lead to meaningful refinement or diminishing returns remains an open question, one that reflects the broader tension between rapid iteration and architectural consolidation in modern deep learning.

### F. APPLICATIONS, BIASES, AND ETHICAL CONCERNS

The widespread adoption of YOLO across domains highlights its status as one of the most influential object detection frameworks to date. As analysed in Section V, its combination of versatility and real-time performance has enabled deployment in both research and production settings, with implementations spanning a broad range of geographic, social, and operational contexts.

However, the widespread use of YOLO also brings to the forefront critical concerns related to algorithmic bias. Since model performance is inherently shaped by the data it is trained on, limited demographic or contextual coverage in training datasets can lead to skewed representations and systematic underperformance in underrepresented scenarios. In practice, this may result in disproportionate detection errors affecting specific social groups, increasing the risk of unfair outcomes when models are deployed in sensitive applications. Mitigating these risks requires promoting dataset diversity, incorporating fairness-aware optimization strategies [399], [400], and applying diagnostic tools capable of auditing prediction behaviour across subpopulations. Such measures are essential to ensure that YOLO-based systems remain reliable and equitable in real-world deployments.

Beyond bias, the use of YOLO in sensitive domains raises important ethical concerns. Its deployment in areas such as healthcare, surveillance, and autonomous systems can involve significant risks if not properly regulated. For instance, in surveillance contexts, YOLO may be used to extract facial features, track movements, or infer personal behaviour without consent, potentially infringing on individual privacy. In medical applications, misidentifications could result in incorrect diagnoses or safety hazards, while in autonomous vehicles, detection failures could contribute to accidents. There is also growing concern over its integration into military systems, where real-time detection capabilities may be exploited for harmful purposes; concerns serious enough to prompt the original YOLO author to step away from further work on the project.

Given these risks, it is essential to implement safeguards in any system built on object detection technologies. Practices such as anonymizing data, securing informed consent, and blurring sensitive regions in datasets can help mitigate privacy violations. Furthermore, systems should be designed with inclusivity and transparency in mind, and their deployment guided by ethical oversight. Ultimately, researchers and developers must remain aware of the broader ethical implications of their work and actively guide its application toward socially beneficial outcomes. By addressing these concerns proactively, the community can help ensure that YOLO remains a powerful and responsible tool with meaningful real-world impact.

## VIII. FUTURE DIRECTIONS FOR YOLO

### A. SUSTAINED FOCUS ON EFFICIENT AND ADAPTABLE DESIGN

The evolution of YOLO has consistently prioritized a balance between accuracy, efficiency, and deployability, favoring architectures that enable real-time inference and hardware adaptability over marginal improvements in benchmark scores. As recent versions have maintained this equilibrium without compromising detection performance, it is likely that future developments will continue to refine this design philosophy. This consolidation of principles suggests that upcoming iterations will place greater emphasis on architectural stability, lightweight modules, and deployment-aware choices as guiding priorities.

### B. EMERGING ROLE OF ATTENTION

The integration of attention mechanisms into the YOLO architecture has gained momentum in recent iterations. While research efforts have explored combining YOLO with Transformer components [401], [402], [403], [404], these approaches have remained peripheral to the main development line. This changed with YOLOv11 and especially YOLOv12, where attention modules are incorporated natively, signaling a growing interest in more expressive representations. YOLOv13 advances this trajectory by introducing HyperACE, which extends beyond local or pairwise attention to capture higher-order semantic correlations across multiple regions of an image, thus reinforcing the trend toward richer representational capacity. Whether this will lead to a broader architectural shift remains uncertain, but it opens the door to future YOLO variants adopting hybrid designs. As demonstrated by the rise of Transformer-based models in Natural Language Processing and other domains, such a shift could eventually redefine the boundaries of what is achievable within the YOLO framework.

### C. REEVALUATING BENCHMARKS AND DATASET PRACTICES

A significant portion of YOLO's development has been benchmarked primarily on MS-COCO, which has long served as a standard reference for evaluation and comparison. However, the operational landscape for detection models has changed substantially since the creation of this dataset. As models are increasingly integrated into real-world systems, the need for benchmarks that reflect contemporary challenges, such as fairness, domain diversity, and deployment-specific constraints, becomes increasingly evident.

Future directions may involve adopting or designing alternative benchmarks that align more closely with current priorities and incorporate reduced bias and stronger ethical safeguards. In this context, the use of synthetic datasets is also gaining relevance, offering controlled environments to test specific behaviours, augment underrepresented classes, or evaluate performance under extreme conditions. While synthetic benchmarks may play an important role in expanding evaluation strategies, concerns remain regarding domain gaps and the risk of overestimating model robustness when relying heavily on artificial data. Careful validation and complementary real-world testing will be essential to ensure that these resources contribute meaningfully to the evaluation of YOLO models.

### D. ECOLOGICAL FOOTPRINT AND SUSTAINABLE AI

Training YOLO models involves significant ecological considerations. YOLO training contributes to a carbon footprint because it is a computationally intensive process that consumes large amounts of electricity, which in many regions still comes from fossil-fuel-based power grids. This impact grows with the use of large datasets that require many hours, days, or even weeks of training, as well as the repeated runs needed for hyperparameter tuning and model selection. Inference also adds to the overall footprint, particularly in video-based applications or when models are deployed at scale. Beyond the computation itself, modern GPUs demand advanced cooling systems and continuous power delivery, further increasing energy use.

Given the importance of environmental responsibility in the current global context, this issue must be considered when developing and deploying YOLO. While the rise of AI has delivered impressive technological progress, it has also contributed to increasing energy consumption and associated emissions. Practical measures to mitigate this include model optimization techniques such as quantization and pruning, the careful selection of smaller variants when suitable, and strategies like reusing pretrained models or fine-tuning instead of training from scratch. These approaches help reduce computational demand without sacrificing accuracy in most scenarios.

A promising path forward is aligned with the broader movement of Sustainable AI (or Green AI), which emphasizes the practice of developing and deploying AI models in an environmentally responsible way. This includes reporting energy consumption and hardware details as part of experimental results, creating greater awareness of ecological costs, and promoting efficiency-aware benchmarks alongside accuracy and latency. Integrating these principles into the YOLO ecosystem would ensure that its continued evolution not only advances detection performance but also reduces its environmental footprint in line with global sustainability goals.

### E. INTEGRATION WITH MULTIMODAL SYSTEMS

The growing influence of generative models and Large Language Models has accelerated interest in multimodal architectures capable of interpreting both visual and textual information. In this context, the integration of vision-language mechanisms is emerging as a natural direction for object detection frameworks. Within the YOLO ecosystem, this shift is exemplified by YOLO-World, which enables detection based on textual prompts rather than fixed class vocabularies, marking the first substantial effort to embed language-guided inference into the YOLO family.

As the field progresses, future directions may explore how prompt-driven mechanisms and language-conditioned detection can become more deeply integrated into core YOLO releases, particularly for applications that require flexible adaptation to dynamic object categories. This convergence of vision and language points to a possible expansion of YOLO's scope beyond fixed-taxonomy recognition, enabling tighter integration with modalities such as video, language, and audio in more complex perceptual systems.

### F. APPLICATION GROWTH AND DOMAIN EXPANSION

As the YOLO framework enters its second decade, its role has evolved far beyond that of a single detection model. What began as a streamlined architecture for real-time object detection has matured into a diverse and adaptable ecosystem, capable of supporting a wide range of deployment scenarios. Its modularity, efficiency, and community-driven development have enabled integration into increasingly specialized domains, including agriculture, medicine, environmental monitoring, and industrial automation. This cross-domain expansion is not merely a consequence of popularity, but a reflection of YOLO's ability to balance performance and usability across heterogeneous contexts.

Rather than approaching saturation, YOLO appears to continue evolving in response to increasingly specific operational demands. Its role has shifted from that of an object detector to a foundational component within task-oriented visual systems. Looking ahead, YOLO is expected to expand into sectors where object detection has traditionally played a minor role, and to adapt further to support more complex pipelines that combine detection with reasoning, tracking, or interaction. This trajectory suggests that YOLO will likely remain a central element in the development of real-time, domain-adaptive visual intelligence.

## IX. CONCLUSION

The year 2025 marks a milestone in the field of object detection, as it coincides with the tenth anniversary of YOLO, one of the most influential detection frameworks to date. Motivated by this, the present review examines the trajectory of YOLO from its original formulation to its most recent iterations. The analysis is structured around four main goals: providing a technical overview of the principal YOLO architectures developed over the past decade; identifying key architectural trends; surveying the main domains of application; and reflecting on future directions in the ongoing evolution of the framework.

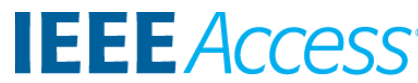

The findings of this review reveal that, over the past decade, YOLO has followed a trajectory defined by consolidation around efficient design, sustained architectural refinement, and increasing adaptability to deployment scenarios. While major milestones have included the shift to anchor-free mechanisms and the standardization of scalable variants, the most notable trend has been the consistent prioritization of real-time inference and hardware-aware optimization. This balance has allowed YOLO to remain relevant in both academic research and industrial applications, sustaining a level of adoption that few object detection frameworks have achieved.

Looking ahead, future developments are expected to build upon this foundation by exploring new forms of architectural integration and functional expansion. Key directions include the incorporation of attention-based components, the transition toward multimodal systems that combine visual and linguistic input, and the gradual diversification of benchmark practices. Together, these directions point to a continued evolution of YOLO beyond conventional object detection, extending its role toward more adaptive, semantically enriched, and context-aware visual understanding.

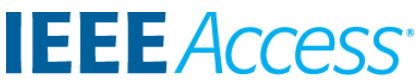

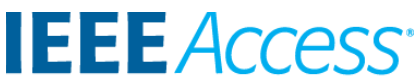

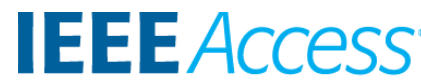

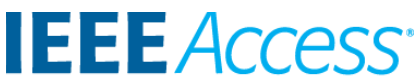

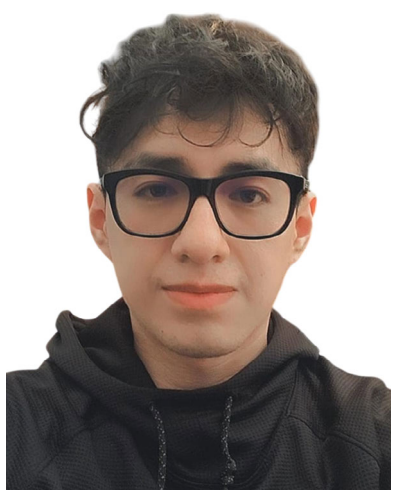

**LEO THOMAS RAMOS** (Graduate Student Member, IEEE) received the degree (magna cum laude) in information technology engineering with a focus on artificial intelligence from Yachay Tech University, Ecuador, in 2023. He is currently pursuing the Ph.D. degree in computer science with the Computer Vision Center, Universitat Autónoma de Barcelona, Spain. He has also held positions as a Research Fellow with The University of Texas at Arlington, USA, where he supervised undergraduate and master's students on scientific research; and a Research Intern with the King Abdullah University of Science and Technology, Saudi Arabia. He is also a Research Engineer with Kauel Inc., Silicon Valley, USA, contributing to the research and development of computer vision systems for the energy industry. He has collaborated with multidisciplinary teams across Europe, the Middle East, and the USA, including renowned researchers from Dubai Police, the Environment Agency Abu Dhabi, Halliburton, Saudi Aramco, Honeywell, and the NASA Jet Propulsion Laboratory. His research interests include deep learning, computer vision, multispectral imagery, medical image analysis, remote sensing, and broader applications of artificial intelligence.

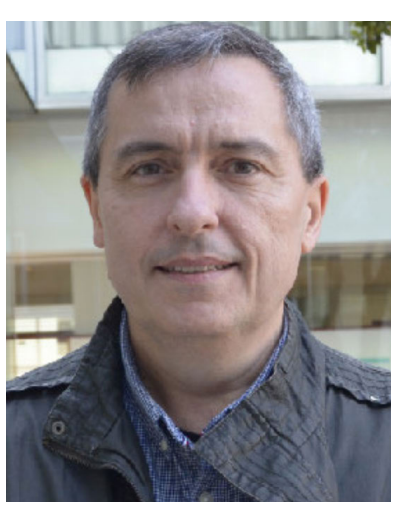

**ANGEL D. SAPPA** (Senior Member, IEEE) received the degree in electromechanical engineering from the National University of La Pampa, Argentina, in 1995, and the Ph.D. degree in industrial engineering from the Polytechnic University of Catalonia, Spain, in 1999. He is currently a Senior Scientist with the Computer Vision Center, Universitat Autónoma de Barcelona, Spain; and a full-time Professor with ESPOL Polytechnic University, Ecuador, where he leads the Computer Vision Team, CIDIS Research Center. He has over 25 years of experience in computer vision and artificial intelligence, with more than 300 publications in top-ranked journals and leading international conferences. From 2023 to 2024, he was included in Stanford University's top 2% most-cited researchers in the subfield of artificial intelligence (career-long ranking). He serves as an Associate Editor for *Pattern Recognition* and a Guest Editor for *Robotics and Autonomous Systems* and *Journal of Computational Science*.